\documentclass[preprint,nofootinbib,letterpaper]{article}

\makeatletter
\let\c@lofdepth\relax
\let\c@lotdepth\relax
\makeatother

\usepackage[english]{babel}
\usepackage{amsmath}
\usepackage{amsfonts}
\usepackage{amsthm}
\usepackage[T1]{fontenc}
\usepackage[utf8]{inputenc}
\usepackage{XCharter}
\usepackage[dvipsnames]{xcolor}
\usepackage{geometry}
\usepackage{sectsty}
\usepackage{titling}
\usepackage{lettrine}
\usepackage{fix-cm}
\usepackage{multicol}
\usepackage{subfigure}
\usepackage{placeins}
\usepackage[ruled,vlined,longend]{algorithm2e}
\usepackage{boldline}
\usepackage{float}
\usepackage{graphicx}
\usepackage{hyperref}

\newcommand{\fancyItalic}[1]{{\footnotesize\usefont{OT1}{phv}{m}{sl}\color{Black}#1}}
\allsectionsfont{\usefont{OT1}{phv}{b}{n}}

\pretitle{\vspace{-2cm}\color{Maroon}\rule{\linewidth}{1pt}\vspace{0.5cm}\fontsize{24}{30}\usefont{OT1}{phv}{b}{n}\selectfont\color{Black}\newline}
\posttitle{\par\vspace{0.5cm}}
\preauthor{\noindent\Large}
\postauthor{\par\color{Maroon}\rule{\linewidth}{1pt}\vspace{-1cm}}

\usepackage[backend=bibtex,style=numeric]{biblatex}

\title{Multi-point Dimensionality Reduction to Improve Projection Layout Reliability}
\author{Farshad Barahimi\,\textsuperscript{*}
	\newline
	\textsuperscript{*}\fancyItalic{PhD student at Dalhousie University, Canada, farshad.barahimi@dal.ca}
}

\date{}

\begin{document}

\maketitle

{\LARGE\noindent Abstract\vspace{0.1 cm}}
\lettrine[lines=2]{\color{Maroon}I}{}{
n ordinary Dimensionality Reduction (DR), each data instance in a high dimensional space (original space), or on a distance matrix denoting original space distances, is mapped to (projected onto) one point in a low dimensional space (visual space), building a layout of projected points trying to preserve as much as possible some property of data such as distances, neighbourhood relationships, and/or topology structures, with the ultimate goal of approximating semantic properties of data with preserved geometric properties or topology structures in visual space. In this paper, the concept of Multi-point Dimensionality Reduction is elaborated on where each data instance can be mapped to (projected onto) possibly more than one point in visual space by providing the first general solution (algorithm) for it as a move in the direction of improving reliablity, usability and interpretability of dimensionality reduction. Furthermore by allowing the points in visual space to be split into two layers while maintaining the possibility of having more than one projection (mapping) per data instance , the benefit of separating more reliable points from less reliable points is dicussed notwithstanding the effort to improve less reliable points. The proposed solution (algorithm) in this paper, named Layered Vertex Splitting Data Embedding (LVSDE), is built upon and extends a combination of ordinary DR and graph drawing techniques. Based on the experiments of this paper on some data sets, the particular proposed algorithm (LVSDE) practically outperforms popular ordinary DR methods visually (semantics, group separation, subgroup detection or combinational group detection) in a way that is easily explainable.}

\begin{center}
\color{Maroon}\rule{0.7\linewidth}{1pt}
\end{center}

If you are just looking for an implementation of LVSDE, visit \url{https://web.cs.dal.ca/~barahimi/chocolate-lvsde/}

\begin{center}
\color{Maroon}\rule{0.7\linewidth}{1pt}
\end{center}

\begin{multicols}{2}

\section{Introduction}
Dimensionality Reduction (DR) or data embedding in a low dimensional space has gained widespread attention in different applications. Typically, a DR technique maps data from a high dimensional space (original space), or on a distance matrix denoting original space distances, onto a 2D visual space, preserving some property of the data~\cite{a32_nonato, a33_rafael}. While some DR techniques such as t-SNE~\cite{a4} and UMAP~\cite{a30,a31} have become essential tools in many application domains such as genomics~\cite{a13,a18}, machine learning~\cite{a20,a19}, NLP~\cite{a87,a88}, cancer research~\cite{a13}, and protein folding~\cite{a89}, the meaning of each dimension of the visual space is not as well-defined as some other DR techniques called linear DR techniques such as Principal Component Analysis (PCA)~\cite{a8} in which each dimension of visual space is a linear combination of dimensions of original space. Regardless, the possibility of errors in the resulting visual representations invites more attention as the presence of groups that do not exist or fake neighbourhoods or mismatching topology structures, or mismatching distances is inevitable in many scenarios.

\begin{figure*}[!ht]
\centering
 \subfigure[LVSDE on 1000 genomes distances]{\fbox{\includegraphics[width=.41\linewidth, trim=-1pt -1pt 0 -1pt]{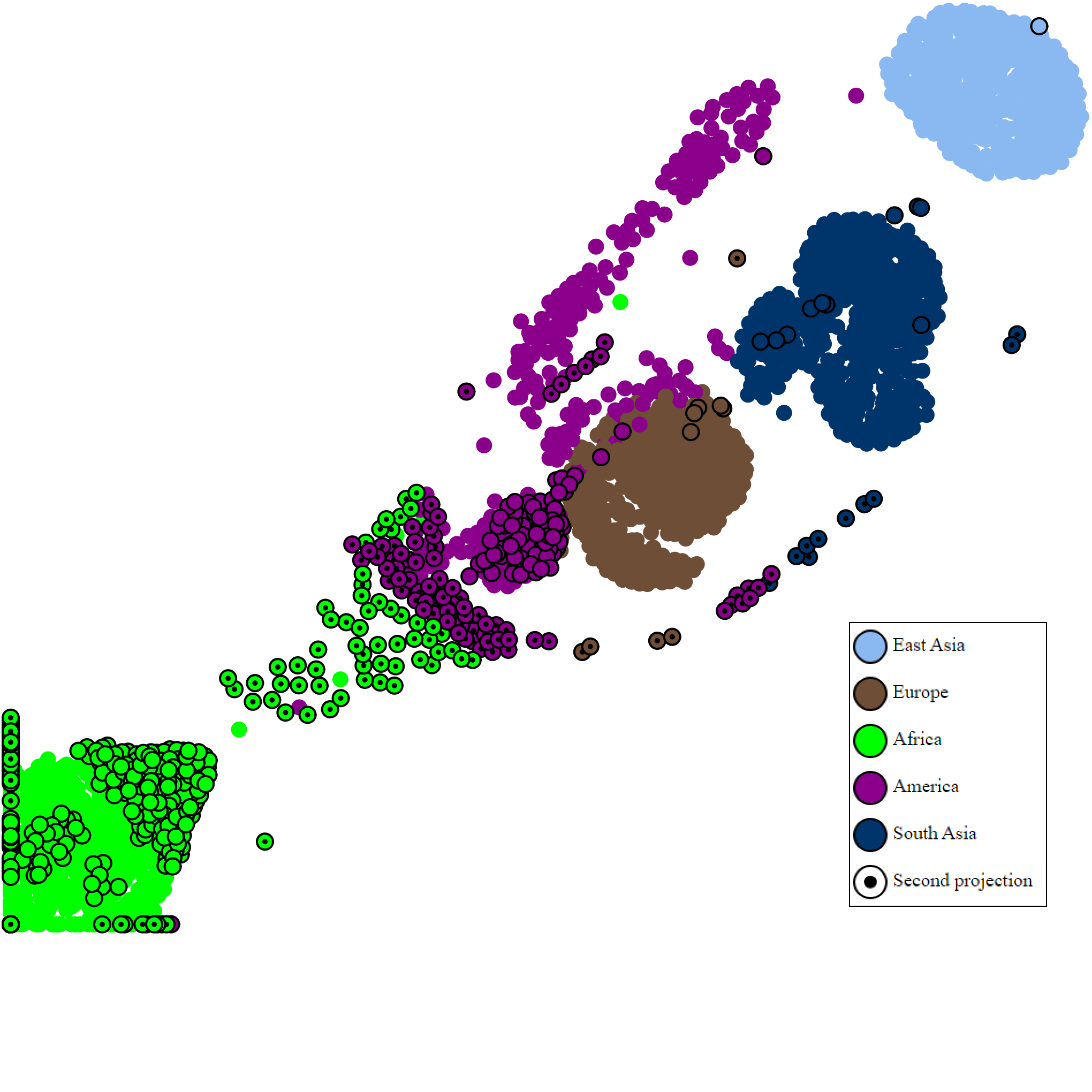}}\quad}\quad
 \subfigure[Duplicates flagged for user selection]{\fbox{\includegraphics[width=.48\linewidth, trim=-2pt -2pt -2pt -2pt]{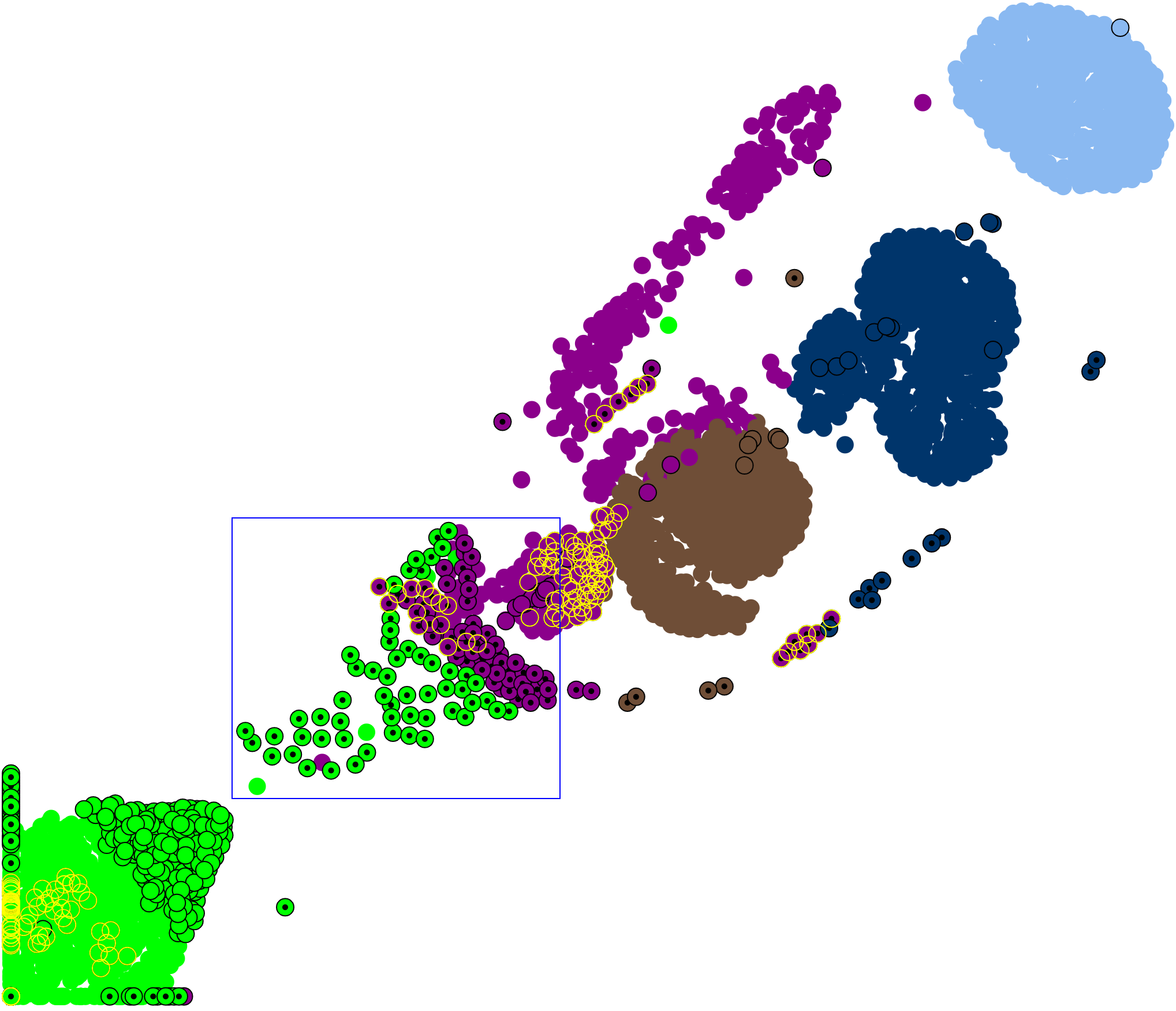}}\quad\quad}\quad
\caption{(a) An LVSDE embedding of the 1000 genomes project data set~\cite{a51_1000_genomes} distances. Points with a black circle around them are in gray layer. Points with a black dot inside them are second projection of an input data instance. Points without a black circle around them are in red layer. Duplication of some of points of America from some area close to Europe to an area close to duplicates of some of points of Africa in an area of visual space between Europe and Africa, matches immigration from Europe and Africa to America. LVSDE has successfully selected points to duplicate and successfully guided them in a meaningful way. For LVSDE configuration 1 is used.
(b) Yellow circles specify points that have a corresponding duplicate point in the blue rectangle meaning that for each point with yellow circle around it there is a point in the blue rectangle which is another projection of the same data instance of original space. The blue rectangle is specified by the user.
}
\label{fig:1000_genomes_distances_embedding_LVSDE_}
\end{figure*}

\begin{figure*}[!ht]
\centering
 \subfigure[LVSDE on a subset of MNIST]{\fbox{\includegraphics[width=.42\linewidth]{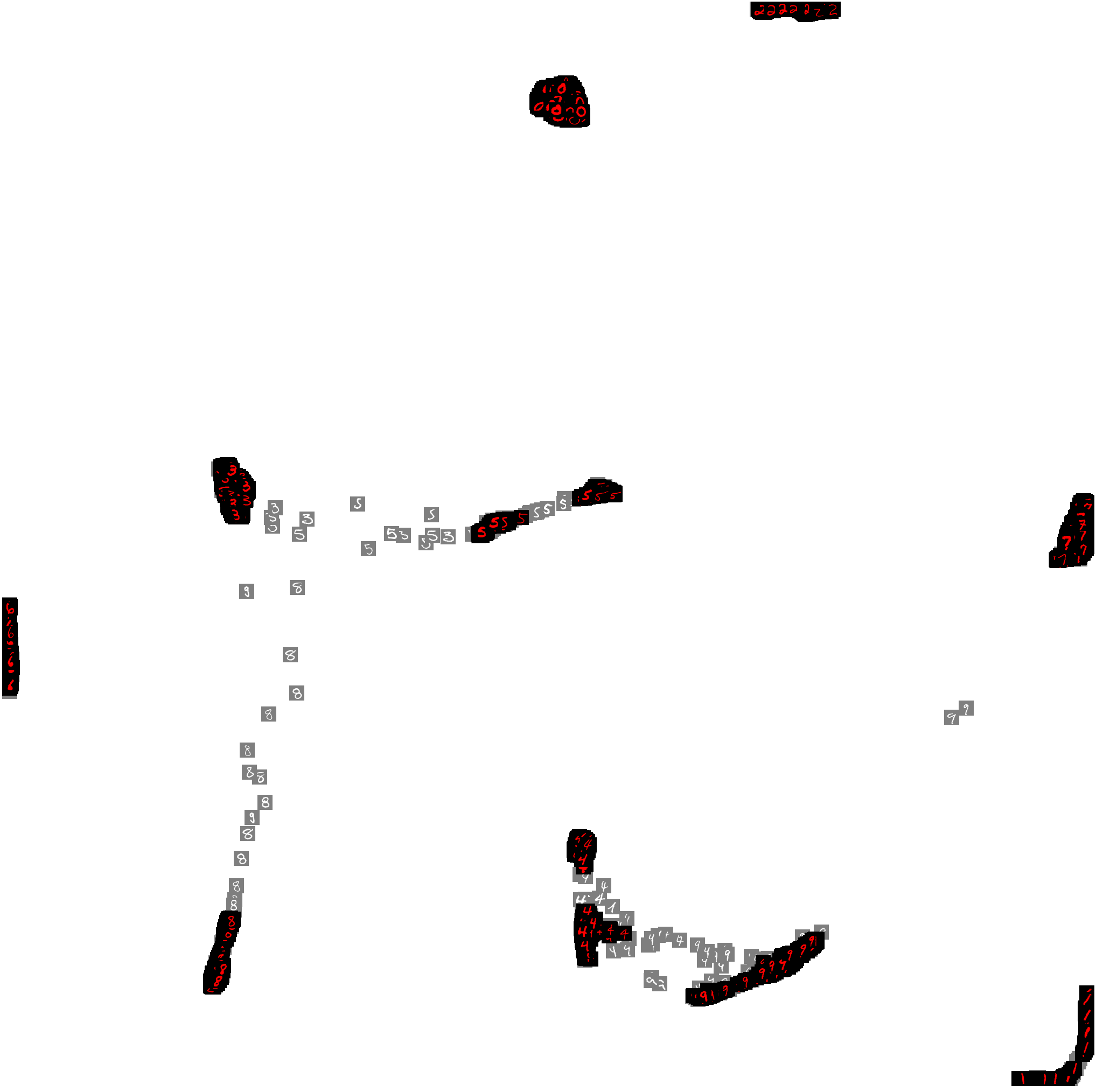}}\quad\quad}\quad
 \subfigure[A cropped section of LVSDE on a subset of MNIST]{\fbox{\includegraphics[width=.45\linewidth]{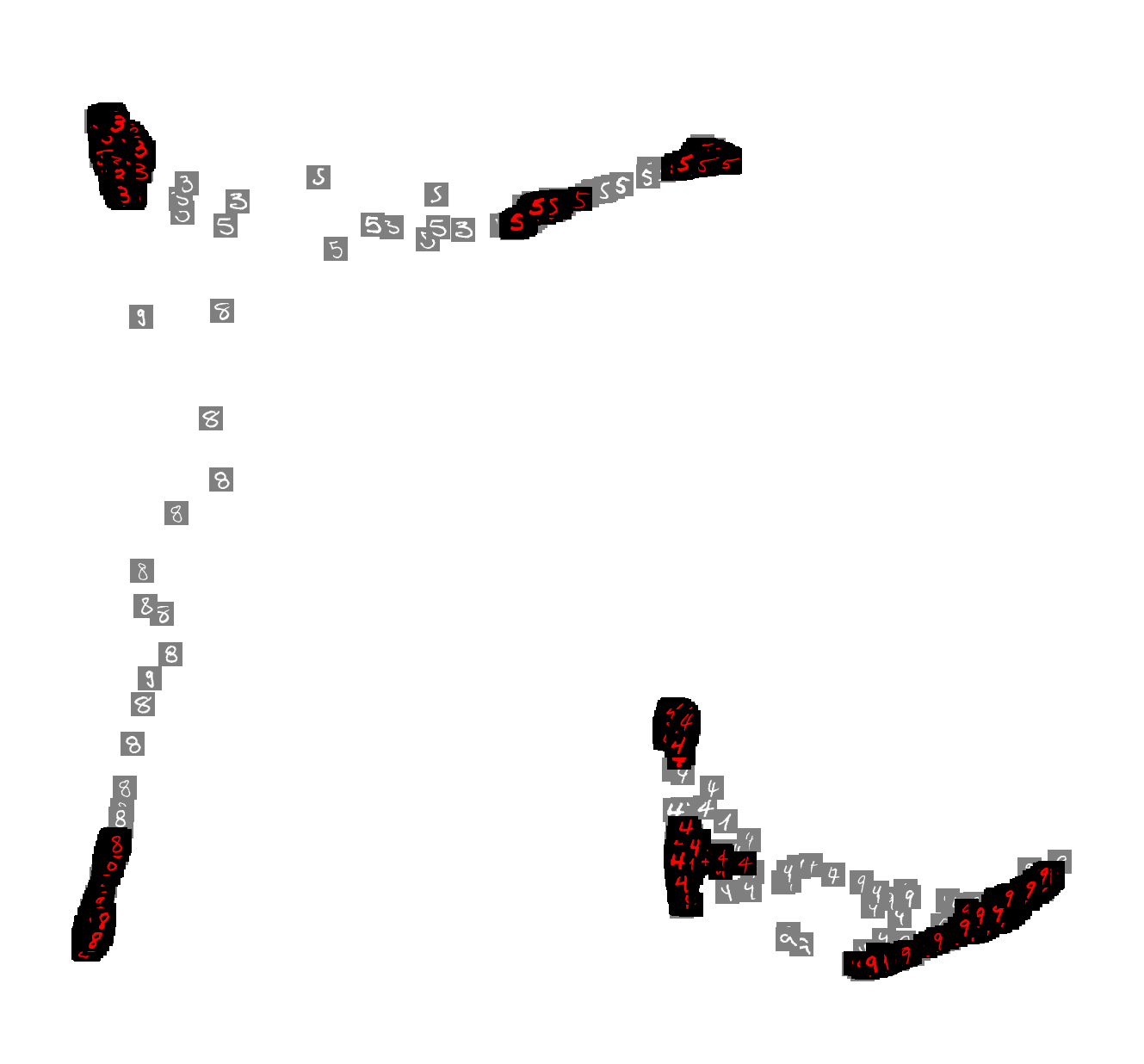}}\quad\quad}\\
 \subfigure[A cropped section of LVSDE on a subset of MNIST]{\fbox{\includegraphics[width=.50\linewidth]{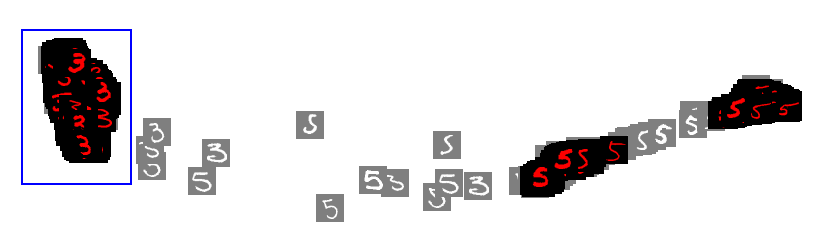}}\quad\quad}\quad
 \subfigure[Overlap reduced for blue rectangle in part c]{\fbox{\includegraphics[width=.17\linewidth]{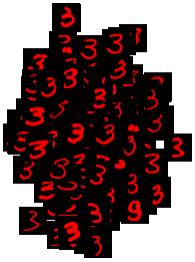}}\quad\quad\quad}\\
 \subfigure[A cropped section of LVSDE on a subset of MNIST]{\fbox{\includegraphics[width=.50\linewidth]{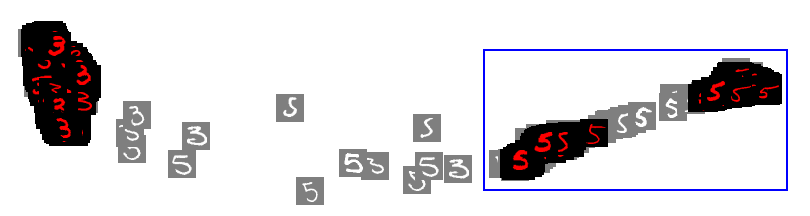}}\quad\quad}\quad
 \subfigure[Overlap reduced for blue rectangle in part e]{\fbox{\includegraphics[width=.2\linewidth]{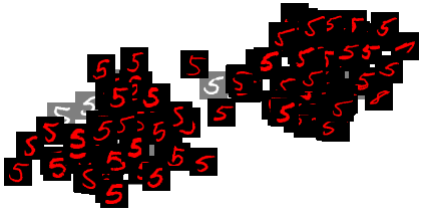}}\quad\quad}\quad
\caption{(a) An LVSDE embedding of a subset of MNIST data set using the information from the whole data set training data.
(b) A cropped section of the LVSDE embedding of a subset of MNIST data set. Separation of digits $3$,$5$ and $8$ on the red layer are better than popular techniques. Separation of digits $4$ and $9$ on the red layer are better than popular techniques. By using layers, the identification of groups is much easier when only considering red layer while gray layer provides some background information.
(c,e) A cropped section of the LVSDE embedding of a subset of MNIST data set.
(d,f) A variant of overlap reduction algorithm of \cite{a62} by Nachmanson et al. is used on the points in blue rectangle in part c or e of this figure.
}
\label{fig:digits_1_}
\end{figure*}

While on the theoretical side, a clear picture of the philosophy of (Multi-layered) Multi-point Dimensionality Reduction as a general DR formulation is drawn in this paper by showing how and which problems it can solve either theoretically or practically from multiple perspectives building on top of previous efforts by others on related problems and prior desires by others for some remedy conceptually the same as Multi-point Dimensionality Reduction or different from it, on the empirical side it is shown in this paper that the particular proposed algorithm (LVSDE) practically outperforms popular ordinary DR methods visually (semantics, group separation, subgroup detection or combinational group detection) in a way that is easily explainable (see Figs.~\ref{fig:1000_genomes_distances_embedding_LVSDE_},~\ref{fig:two_configs},~\ref{fig:digits_1_},~\ref{fig:digits_2},~\ref{fig:IRIS},~\ref{fig:MeeefTCD_1},~\ref{fig:MeeefTCD_2} and~\ref{fig:images}) and performs in close proximity to top on most of the data sets studied in the paper in a quantitative analysis based on KNN classification accuracy.

\end{multicols}

\begin{multicols}{2}
On the motivation, in the proceeding Section~\ref{section:motivation_and_philosophy} the relevance of visualization of fuzzy sets~\cite{a81} in contrast to fuzzy clustering~\cite{a80} is discussed in addition to the relevance of visualizing non-metric distances previously highlighted by Van der Maaten and Hinton~\cite{a5} and Cook et al. \cite{a6}, and then the discussion of the inherent issue came to light with non-metric distances is expanded to high dimensional spaces by looking at neighbourhood interpretation in original space farther than the superficial Euclidean distance neighbourhood definition borrowing some ideas from Principal Component Analysis (PCA)~\cite{a8}, and then the relevance to different semantic contexts for word embeddings previously highlighted by Huang et al.~\cite{a25} is discussed and finally the current inability but the important potential relevance of preservation of well-defined finite point set topology structures is discussed.
\end{multicols}

\twocolumn

\FloatBarrier
\section{Motivation And Philosophy}
\label{section:motivation_and_philosophy}

While dimensionality reduction has been relying on preserving geometric properties and topology structures of data, the ultimate goal on the application front has been to visualize the semantic properties of data with high reliability trying to visualize groups of data based on a meaningful grouping of data for that application. Such groupings are sometimes approximated with geometric properties and topology structures of data and then preserved into visual space delivering the ultimate goal to some extent. An important shortcoming of dimensionality reduction with exactly one projection per data instance point is that it cannot reliably and to the full extent show independent belonging to multiple fuzzy sets. It is important to distinguish between the definition of fuzzy sets~\cite{a81} and fuzzy clustering~\cite{a80} as the definition of fuzzy clustering has an additional restriction that the fuzzy sets~\cite{a81} definition does not have which is the sum of intensities of belonging to different clusters should be 1. In the definition of fuzzy clustering~\cite{a80} which has taken a more univariate approach, the intensity of belonging to each cluster depends on the intensity of belonging to the other clusters but in the model of belonging to multiple fuzzy sets by Zadeh~\cite{a81} which is a more multivariate approach there is no such restriction and therefore the intensity of belonging to one cluster can change independent of the intensity of belonging to another cluster and is more in line with multivariate classification~\cite{a82} research emerging as a hot topic in many applications~\cite{a83,a84,a85}. While ordinary dimensionality reduction has been used in the past for exploring fuzzy clusterings by Ying Zhao, et al.~\cite{a61} in a user study, it can be only accurate or reliable for univariate fuzziness in fuzzy clustering~\cite{a80} rather than the multivariate model of belonging to multiple fuzzy sets~\cite{a81}, and also, while their paper focuses on comparing user's responses within a set of ordinary dimensionality reduction techniques on a particular limited set of data sets already gone through a fuzzy clustering algorithm, the method in this paper not only is examined on more realistic data that has not already gone through a fuzzy clustering algorithm, it goes beyond univariate fuzzy semantics and pushes toward visualizing multivariate fuzzy semantics more accurately and more reliably. While a point shown between two groups of points in visual space can indicate some form of fuzzy relationship to the two groups, the fact that if a point is between two groups of points, moving toward one group will be moving away from the other group restricts the possibility independently specifying the intensity of belonging to the first group and intensity of belong to the second group using any position-based visual clue in visual space. Having possibly more than one projection per data instance can be a better indicator of unrestricted belonging to multiple fuzzy sets by allowing different indicators for each fuzzy set. Existence of such indicators improves the potential of preserving geometric and topology structures of data for approximating multivariate fuzzy semantics.

While in visual space, Euclidean distance from an area of visual space or some continuous transformation of it, is generally considered as a visual clue for how related a point is from that area of visual space, the fact that Euclidean distance has to adhere to triangular inequality as it is a metric distance, restricts the closeness it can imply to different areas of visual space creating a mismatching model if the closeness relationship in the data to be visualized does not adhere to triangular inequality. The inability of ordinary dimensionality reduction to properly visualize non-metric distances is well highlighted by Van der Maaten and Hinton~\cite{a5} and Cook et al. \cite{a6}, however they fell short of bringing a proper solution to the problem they raised correctly as their idea of embedding into to multiple maps, each with exactly one projection per data instance is not adequately resolving the problem and causes more issues than it fixes, in part due to the effect of change blindness \cite{a58,a59} on multiple embeddings and in part due to the observation that the aggregate sum of the total number of projected points in all of the multiple embedding maps as they suggest is unjustifiably high and also due to the fact that the total number of projected points needed to explain the same set of closeness relationships is much higher for embedding into multiple maps each with exactly one projection per data instance, rather than embedding into a single map with possibly more than one projection per data instance. Indeed, embedding into a single map with possibly more than one projection per data instance is a more natural way.

\begin{figure*}[!ht]
 \centering
\includegraphics[width=7cm]{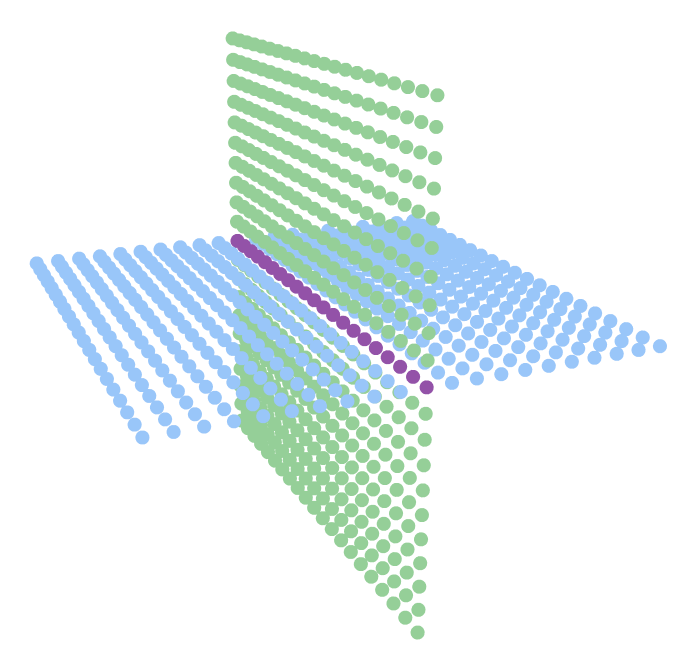}
 \caption{Points on the surface of two planes regularly distributed. The points on the purple line have two distinct neighbourhoods based on two different planes.}
 \label{fig:two_planes}
\end{figure*}

\begin{figure*}[!ht]
\centering
 \subfigure[One visual neighbourhood]{\fbox{\includegraphics[width=.45\linewidth, trim=-1pt -1pt 0 -1pt]{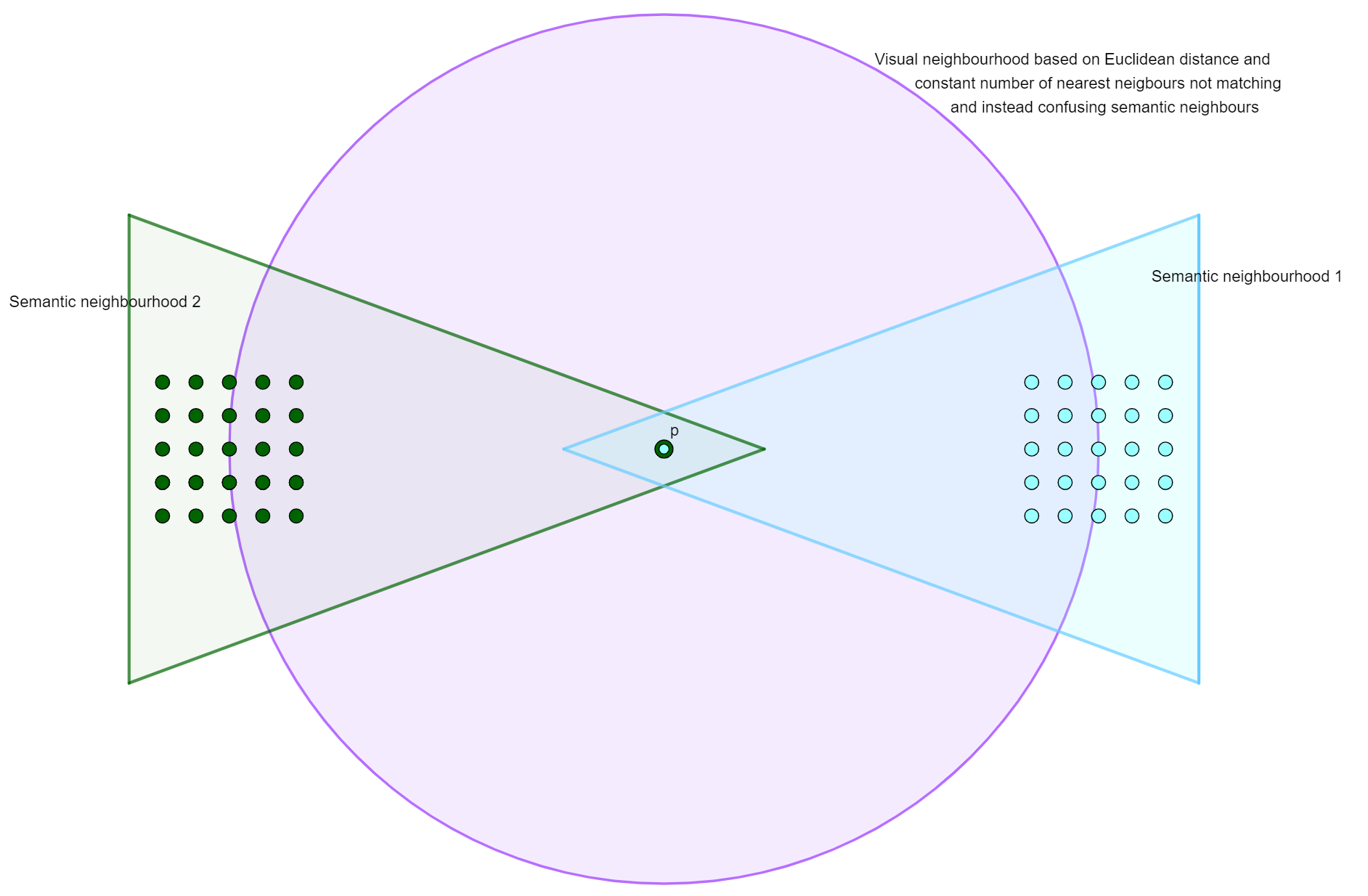}}\label{fig:}}\quad
 \subfigure[Two visual neighbourhoods]{\fbox{\includegraphics[width=.45\linewidth, trim=-2pt -2pt -2pt -2pt]{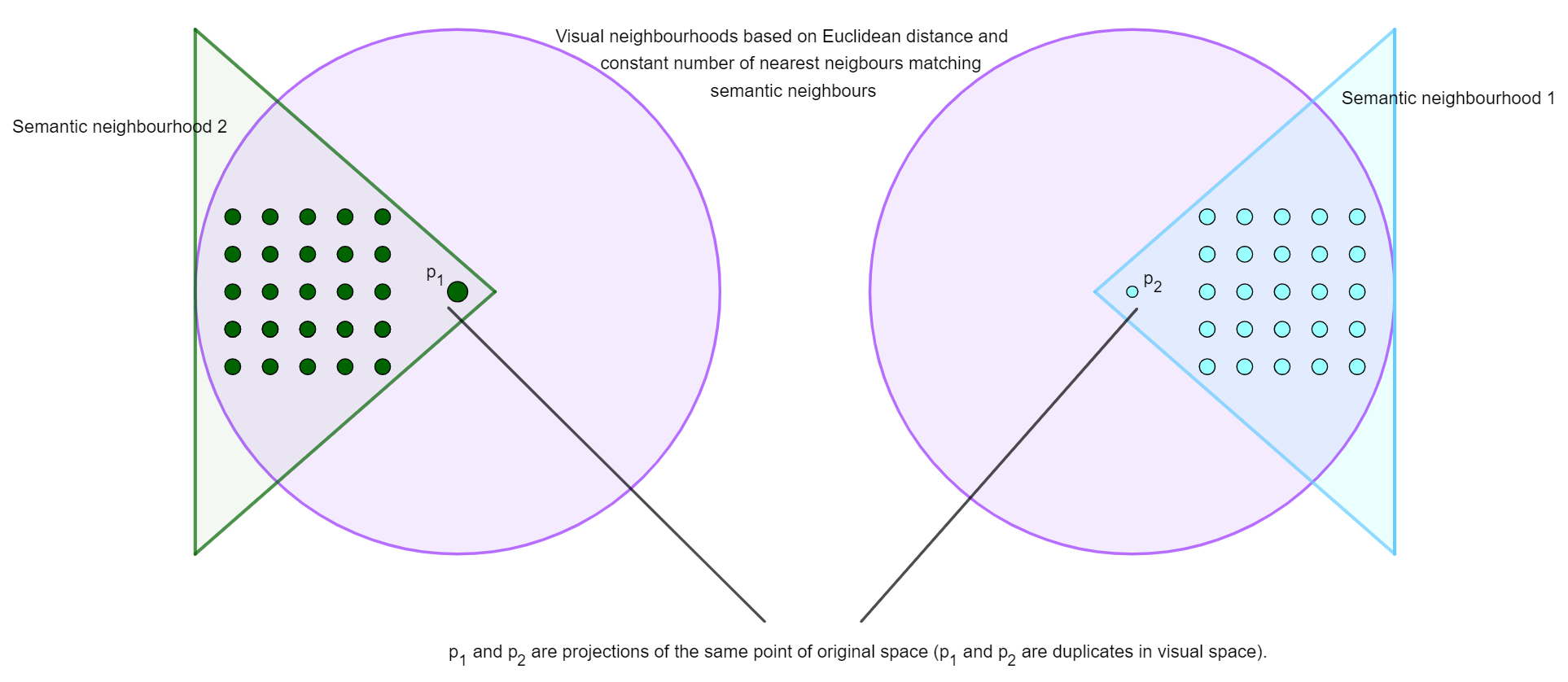}}\label{fig:}\quad\quad}\quad
\caption{(a) A data instance projected into visual space and two semantic neighbourhoods around it. One visual neighbourhood based Euclidean distance and constant number of nearest neighbours not matching and instead confusing semantic neighbourhoods.
(b) Two projections of a data instance into visual space. Two visual neighbourhood based on Euclidean distance and constant number of nearest neighbours matching semantic neighbourhoods.
}
\label{fig:multiple_neighbourhoods}
\end{figure*}

Interesting argument pops up when looking at the fact that for many applications there exists data already in a non-metric distance form not adhering to triangular inequality, and then looking at how other data (i.e. high dimensional data) are dealt with. If the semantic distance in data can be non-metric, then why Euclidean distance is usually used to model the semantics of neighbourhoods in original space when the original space data is in high dimensional form, knowing that Euclidean distance is a metric distance and has to adhere to triangular inequality. Is this just looking at the problem superficially? One might argue that Cosine distance has been used too and Cosine distance is not a metric distance but Cosine distance still is far from an unrestricted distance model as it equates to half of squared Euclidean distance of normalized vectors. Borrowing some ideas from Principal Component Analysis (PCA)~\cite{a8}, this paper contributes to the philosophy that semantic neighbourhoods should be modelled on the set of perpendicular geometric projections of points of original space on each hyper-line of a specific set of hyper-lines or each manifold of a specific set of manifolds of the high dimensional original space where the specific hyper-lines or manifolds are meant to represent different semantic aspects of data as assuming just one aspect for data is an oversimplification of an important problem. Then a point can have several semantic neighbourhoods defined around it based on different hyper-lines or manifolds. Then the challenge would be how to bring those neighbourhoods to a visual space usually perceived through Euclidean distance which has to adhere to triangular inequality. Multi-point Dimensionality Reduction can better show multiple neighbourhoods around a single point of original space by having multiple projections of it in visual space giving a data instance more freedom to express itself in different neighbourhoods of visual space using multiple projected points of it instead of one limited by triangular inequality. Also layers can improve the interpretation of those neighbourhoods of original space in visual space by providing a concept for separation of the set of visualized points allowing a dimensionality reduction method to visually restrict a neighbourhood to a particular layer. Therefore Multi-layered Multi-point Dimensionality reduction can better visually model multiple semantic neighbourhoods around a point of original space. In this paragraph the meaning of perpendicular geometric projection on a manifold, is based on the normal vector of the tangent space on any point on the manifold creating a perpendicular ray from any point on the manifold which means a point can have multiple perpendicular geometric projections on a manifold. A simplification of multiple neighbourhoods around a point of original space can be understood from Fig.~\ref{fig:two_planes} example. A further complicated example in Fig.~\ref{fig:multiple_neighbourhoods} shows how multiple projections per data instance can elevate the challenge of visualizing multiple semantic neighbourhoods around a point of original space in visual space. 

\setlength{\fboxsep}{0pt}

The relevance of visualizing multiple neighbourhoods for dimensionality reduction has also recently been focus of interesting work of Huroyan, et al.~\cite{a96} where they use dimensionality reduction to three dimensions as a remedy to have simultaneous neighbourhoods, however it has been long disputed whether the fact that 3D visualization are rendered into a 2D visualization before perception limits the ability for perceiving all information in a 3D visualization in a single view. Nevertheless, in ~\cite{a97} Barahimi and Wismath argued that the virtual reality technologies and stereotype 3D might provide some improvement over plain 3D.

The relationship between hubness~\cite{a90,a91} and dimensionality reduction is studied quite well, and the challenge that hub data instances (data instances with high number of occurrences in other points' $k$ nearest neighbours in the directed nearest neighbours graph of original high dimensional space) pose to dimensionality reduction, clustering and machine learning is well expressed~\cite{a90,a91,a92}. When building nearest neighbours graphs of original space and then interpreting it in visual space, if a data instance can have more than one projected point, each edge of the nearest neighbours graph of original space can be considered only incident on just one of the projected points of that data instance in visual space, allowing visual space to have lower vertex degrees and therefor reduce hubness of some vertices. Therefor Multi-point Dimensionality Reduction can mitigate the challenge of data instances with high hubness to some extent. Apart from the problem formulation, the particular algorithm (LVSDE) that is proposed in this paper, in the forth phase of it, can change the neighbourhood graph in a way that reduces the degree (number of incident edges) of the projected points that are duplicated and their neighbours.

For word embeddings, there has been a specific research by Huang et al.~\cite{a25} about how different contexts should affect them as they opted to produce different word embeddings for each word based on the context. While their technique as a natural language processing technique is specific to word embeddings and is not generalizable to other applications, it also requires additional context data, distancing it to some extent from dimensionality reduction paradigm. In this paper, not only a general solution (LVSDE) is proposed applicable to different application domains, it does not require any context data and tries to find the semantic contexts implicit in the data when duplicating data.

One big but still challenging and not resolved potential of Multi-layered Multi-point dimensionality reduction is preservation of a well-defined topology structure where the meaning of topology is interpreted in the context of finite point set topology~\cite{a63,a64} rather than infinite geometric topology~\cite{a65} or infinite shape theory~\cite{a66} although being relevant. If the challenge of visually connecting multiple projections of a point is adequately addressed this can become a significant potential for Multi-layered Multi-point Dimensionality Reduction. The preservation is in a sense that a well defined procedure finds a topology structure in original space and a corresponding well defined visual procedure finds a close or exactly the same topology structure in visual space. Although preservation of a well-defined topology structure is not the objective of this paper, an attempt to partially and insufficiently address the challenge of visually connecting multiple projections of a point is made in this paper. A relevant prior work in the direction of preserving topology structures, is the interesting work of Doraiswamy et al.\cite{a70} in which they preserve 0-dimensional persistence diagram of the Rips filtration~\cite{a71,a72} of high dimensional data.

In part of an interesting research by Aupetit\cite{a69}, the question of whether as a result of dimensionality reduction a manifold in original space that data instances were laying on is torn in visual space is studied. While the study stops short of giving any theoretical or empirical remedy to preserve neighbourhoods when tearing happens, it is definitely related to the theoretical argument that is made in Section~\ref{section:problem_formulation_formalization} that Multi-layered Multi-point Dimensionality Reduction as a problem formulation, in a theoretical fashion can maintain neighbourhoods when manifold tearing happens. For the particular example of points on a 3D cylinder that is talked about in Section~\ref{section:problem_formulation_formalization}, no empirical solution is provided in this paper and rather the discussion is only done in theoretical fashion.

\FloatBarrier
\section{Problem Formulation Formalization And Visual Metaphors}
\label{section:problem_formulation_formalization}

To start formalization of the problem formulation, let's recall the definition of ordinary dimensionality reduction or data embedding. Given a set of data instances $Q=\{q_1,q_2,...,q_n\}$ either specified by points in an $m$-dimensional space called original space where each $q_i \in \mathbb{R}^m$, or specified on rows and columns of a distance matrix $\bar{\delta}$ denoting original space distances, in an ordinary dimensionality reduction the goal is to project (map) each $q_i$ onto a point $p_i$ in a 2-dimensional space called visual space so that some of geometric properties or topology structures of the original space or original space distances, such as pairwise distances or local neighbourhoods are preserved as much as possible \cite{a33_rafael,a60}.

One approach to dimensionality reduction, is distance preservation strategy in dimensionality reduction where, if distances in the original space between $q_i$ and $q_j$ are denoted by $\bar{\delta}(q_i,q_j)$ and distances in the visual space between $p_i$ and $p_j$ are denoted by $D_v(p_i,p_j)$, a dimensionality reduction technique with distance preservation strategy tries to minimize the difference between $D_v(p_i,p_j)$ and $\bar{\delta}(q_i,q_j)$ for all $1 \le i < j \le n$ or some variant of that \cite{a33_rafael,a60,a2}.

\begin{figure*}[!t]
 \centering
\includegraphics[width=12cm]{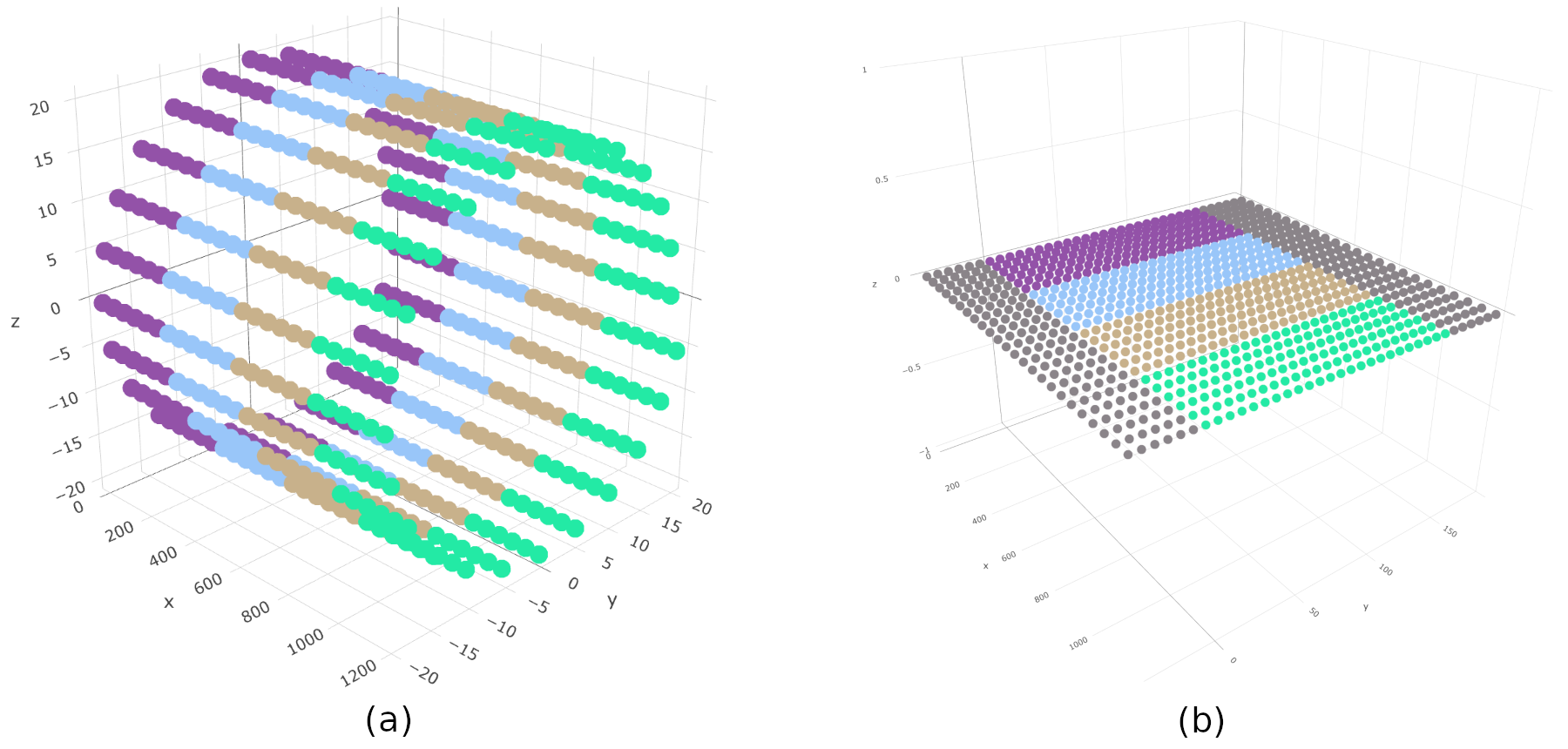}
 \caption{(a) Points on the surface of a cylinder regularly distributed. (b) The torn and unfolded cylinder with duplicate points added to the sides where the points coloured gray on each side are duplicates of the points with non-gray colour on the other side (the image is not result of any particular empirical dimensionality reduction method and only highlights the potential of Multi-layered Multi-point Dimensionality Reduction as a problem formulation in a theoretical fashion).}
 \label{fig:cylinder}
\end{figure*}

Although well-accepted, if distances are Euclidean distance, this minimization usually fails to perfectly reach zero difference for all $1 \le i < j \le n$, even for simple 3D data sets, such as points on a cylinder's surface distributed regularly like the one in Figure~\ref{fig:cylinder}(a). The failure expands to preserving local neighbourhoods too which is better addressed with duplicate points in Figure~\ref{fig:cylinder}(b). Duplicate points that are not allowed in ordinary dimensionality reduction, but are allowed in the Multi-point Dimensionality Reduction defined bellow, enable a data instance to be projected onto two far neighbourhoods in visual space by allowing two separate projections for that data instance as if the first one is duplicated so that the second one becomes another mapping of the same data instance.

\newtheorem{definition}{Definition}[section]
\begin{definition}{Multi-point Dimensionality Reduction.}

Given a set of data instances $Q=\{q_1,q_2,...,q_n\}$ either specified by points in an $m$-dimensional space called original space where each $q_i \in \mathbb{R}^m$, or specified on rows and columns of a distance matrix denoting original space distances, a Multi-point Dimensionality Reduction maps each data instance $q_i$ to a non-empty set of points $S_i=\{s_{i_1}, s_{i_2}, ..., s_{i_{\lambda_i}}\}$ where $\forall s_{i_t} \in \mathbb{R}^2$.
\end{definition}

With this definition, a variant of the classical distance preservation problem for ordinary dimensionality reduction can be transformed to \textit{Multi-point Dimensionality Reduction} as follows:

\begin{definition}{Multi-point Dimensionality Reduction Distance Preservation Problem.}

Given a set of data instances $Q=\{q_1,q_2,...,q_n\}$ either specified by points in an $m$-dimensional space called original space where each $q_i \in \mathbb{R}^m$, or specified on rows and columns of a distance matrix denoting original space distances, the Multi-point Dimensionality Reduction Distance Preservation Problem seeks to map each data instance $q_i$ to a non-empty set of points $S_i=\{s_{i_1}, s_{i_2}, ..., s_{i_{\lambda_i}}\}$ where $\forall s_{i_t} \in \mathbb{R}^2$, and minimize
 
\begin{equation}
\label{eq:dbmpp}
\frac{\sum\limits_{i=1}^n\sum\limits_{j=1}^n( {\min\limits_{(s_{i_t}, s_{j_r})\in S_i \times S_j} ( [\bar{\delta}(q_i,q_j) - D_v(s_{i_t},s_{j_r})]^2 )} )}{n^2}
\end{equation}
 
 subject to $\sum_{i=1}^{n}|S_i| \le \varphi$, where $\varphi$ is a constant for limiting the number of points in visual space.
\end{definition}

While in Section~\ref{section:related_work_and_expanded_motivation}, the benefit of splitting the set of projected points into different layers for distinction between different neighbourhoods around a projected point was discussed, by combining the idea of layers with the idea of more than one projection per data instance, Multi-layered Multi-point Dimensionality Reduction is defined as follows:

\begin{definition}{Multi-layered Multi-point Dimensionality Reduction.}

Given a set of data instances $Q=\{q_1,q_2,...,q_n\}$ either specified by points in an $m$-dimensional space called original space where each $q_i \in \mathbb{R}^m$, or specified on rows and columns of a distance matrix denoting original space distances, a Multi-layered Multi-point Dimensionality Reduction maps each data instance $q_i$ to a non-empty set of points $S_i=\{s_{i_1}, s_{i_2}, ..., s_{i_{\lambda_i}}\}$ where $\forall s_{i_t}\in \mathbb{R}^2$, and also maps each $s_{i_t} \in S_i$ to a layer number $\eta(s_{i_t})$. For convenience $\gamma_{i_\theta} \subseteq S_i$ is defined as the subset of points in $S_i$ such as $s_{i_\Omega}$ for which the condition $\eta(s_{i_\Omega})=\theta$ holds, meaning that $\gamma_{i_\theta}$ is the set of projections of $q_i$ in the layer $\theta$ of visual space. 
\end{definition}

One of the challenges introduced by Multi-layered Multi-point Dimensionality Reduction is that a point might not have any projection in one of the layers in visual space. If only that layer is displayed, this can hide some relevant information. One approach to address this challenge is to display all layers but with a different visual indicator for each layer such as colour. While the interpretation of the meaning of the distinction between different layers is open in the definition of Multi-layered Multi-point Dimensionality Reduction, this paper wants to emphasize on a particular type of Multi-layered Multi-point Dimensionality Reduction where there are only two layers named red layer and gray layer. The red layer is intended for more reliable points and the gray layer is intended for less reliable points. As a result, a Red Gray Embedding is defined below:

\begin{definition}{Red Gray Embedding.}

A Red Gray Embedding, is a \textit{Multi-layered Multi-point Dimensionality Reduction} with just two layers named red layer and gray layer. The red layer is intended for more reliable points and the gray layer is intended for less reliable points.
\end{definition}

In the above definition a point can have several projections in the red layer and several projections in the gray layer at the same time. The following stricter definition is more pragmatic as it is explained after.

\begin{definition}{Strict Red Gray Embedding.}

A Strict Red Gray Embedding, is a Multi-layered Multi-point Dimensionality Reduction with just two layers named red layer and gray layer where points that have more than one projection are mapped to gray layer. (points that have exactly one projection can be mapped to either red layer or gray layer). The red layer is intended for more reliable points and the gray layer is intended for less reliable points.
\end{definition}

Such a strict definition is designed based on the theory supported by the experimental results of this paper that if a point is projected unreliably, most likely it is because it is related to multiple groups of points as independent separate fuzzy sets in a way that is not consistent with triangular inequality and therefore having more than one projection might help mitigate the unreliability.

To visualize a Strict Red Gray Embedding, several simple visual metaphors are used in this paper to fill the gap of necessity rather than to form the ideal visual metaphors, as a user-study paper may be better suited for that purpose. For example, in the simplest form, the points in red layer are rendered with red colour and the points in gray layer are rendered with gray colour. Another example is that giving priority to one layer when overlap happens, meaning that first the points of one of the layers are rendered and then the points of the other layer are rendered on top of them. To be able display class colours in addition to separation of layers, a visual metaphor is used where class colours are used as the rendering colour for the points of visual space but a black circle is drawn around points of only one of the layers, or points of only one of the layers are drawn smaller. Another variant that is considered for grayscale images especially for the readability of MNIST digits is to draw images for the points in red layer by a linear transformation of grayscale values $[0-255]$ to the range of colours starting with full black and ending to full red. For the points in the gray layer however, the linear transformation that transforms $[0-255]$ grayscale values to the range of colours starting from mid gray colour to full white is considered.

\section{Method}
\label{sec:method}

In this section, the proposed method (algorithm) of this paper named Layered Vertex Splitting Data Embedding (LVSDE)\footnote{LVSDE was named Red Gray Plus in the first arXiv publication but has been renamed.} is discussed which produces a Strict Red Gray Embedding. Although this section starts with giving summaries as a mean to build the big picture description of the algorithm, the details and reasoning follows it. The notation that is used in this section and in this paper with respect to distances in original space or visual space is $D_v$ for distances in visual space, $\bar{\delta}$ for distances in original space before any transformation, $\delta$ for original space distances after transformation using the neighbour-normalized distance transformation that is introduced later in this paper, $D_{v_{max}}$ for maximum distance in visual space and $\delta_{max}$ for maximum of transformed distance of original space.

\subsection{Overview}
\begin{figure*}[ht!]
 \centering
\includegraphics[width=\linewidth]{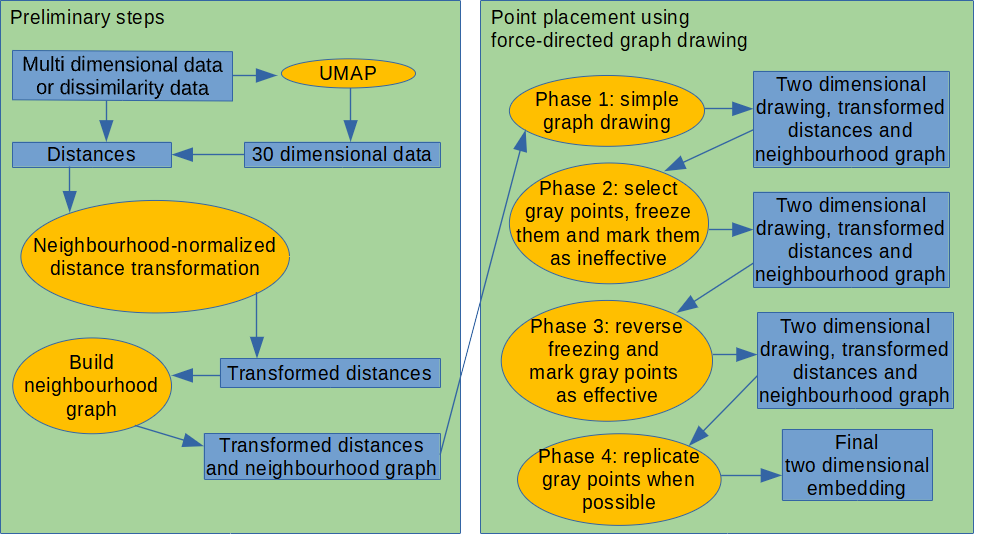}
 \caption{Summary of the proposed dimensionality reduction algorithm of this paper (LVSDE) showing the flow of data through different components of the algorithm}
 \label{fig:chart}
\end{figure*}

An overall picture of the proposed algorithm (LVSDE) is summarized in Fig.~\ref{fig:chart} with some of details missing in the figure but discussed in Procedures~\ref{procedure:applyRepulsiveForces} and~\ref{procedure:applyAttractiveForces}, and/or the text of this paper. The algorithm builds upon, extends and modifies a combination of possibly UMAP \cite{a30,a31}, Fruchterman and Reingold \cite{a10} force-directed graph drawing algorithm and force scheme \cite{a2} force-directed data embedding technique. It is mentionable that the Fruchterman and Reingold \cite{a10} force-directed graph drawing algorithm, that is modified in this paper to form the foundation of the proposed algorithm of this paper, although being relatively old, has been actively used in recent years in a variety of research areas and credible publication venues such as the Nature, Nature communications and Nature medicine ~\cite{a73,a74,a75}, and has more recent GPU-based faster variants~\cite{a76,a77} although the GPU based variants of it are avoided in this paper.

The proposed algorithm in this paper has three preliminary steps and four phases but the first preliminary step which is UMAP \cite{a30,a31} to 30 dimensions, is optional as it is discussed later. The second preliminary step is neighbourhood-normalized distance transformation. The third preliminary step is to build a neighbourhood graph $G$ based on transformed distances from the previous preliminary step. Once a neighbourhood graph is ready a four phase force-directed graph drawing process is started but it deviates from general force-directed graph drawing technique of \cite{a10} from multiple perspectives as it is summarized below and described later in Section \ref{subsection:force_directed_graph_drawing}. The first perspective is using transformed distances in addition to neighbourhood graph and visual distances in calculating the forces. The second perspective is that not all vertices necessarily apply forces or get forces applied to them at all phases of the algorithm. The third perspective is the objective of the formula used for computing forces and therefore the formula used to compute forces. The forth perspective is that the neighbourhood graph is modified in the forth phase of the algorithm. The fifth perspective is that while the Fruchterman and Reingold \cite{a10} algorithm assumes unit mass for each vertex practically equating aggregate force on a vertex to its acceleration in the Newton's second law of motion (subject to some barrier limitations), in the proposed method of this paper the vertex mass is changed from its initial unit mass when duplicating a vertex (projected point) to account for lower number of edges on the duplicated vertices (projected points). The sixth perspective is that while the concept of temperature is kept which is a variable limitation on how far a vertex can move in a single iteration, the temperature is not decreased monotonically from the beginning to end, and instead temperature monotonically is decreased inside each phase and increase temperature just before next phase. The seventh perspective is that a frame border is not used at the first phase but starting at the beginning of second phase, a frame border is used.

While the proceeding Section~\ref{preliminary_steps} elaborates on the three preliminary steps of the proposed algorithm of this paper, the Section~\ref{subsection:force_directed_graph_drawing} which follows it, elaborates the four phases of the proposed algorithm through a force-directed graph drawing.

\subsection{Preliminary Steps}
\label{preliminary_steps}

\subsubsection{UMAP to 30 dimensions}
The first preliminary step which is UMAP to 30 dimensions, is optional. The reason for optional UMAP to 30 dimensions is to be able to benefit from its manifold learning capabilities when the data or sufficient portion of the data is uniformly distributed on a Riemannian manifold and to be able to skip UMAP when the data or sufficient portion of the data is not uniformly distributed on a Riemannian manifold. In some scenarios, another benefit of UMAP to 30 dimensions as an optional preliminary step for the proposed algorithm, is that UMAP is much faster than the rest of the proposed algorithm and therefore such a preliminary step allows partially learning the embedding from a larger data size and then for the rest of the algorithm using a smaller data size. If the user opts to use UMAP to 30 dimensions as a preliminary step, then after UMAP to 30 dimensions is finished a choice needs to be made on how to compute distances on the resulting 30 dimensional data. While the typical choice here is Euclidean distance, another possibility especially for text data sets is Cosine distance and those distances will be considered as original space distances ($\bar{\delta}$) for the rest of the algorithm.

\subsubsection{Neighbourhood-normalized distance transformation}
The second preliminary step is neighbourhood-normalized distance transformation. The proposed algorithm relies on drawing a neighbourhood graph initially being a $\widehat{p}$ nearest neighbour graph where $\widehat{p}$ is a constant parameter indicating neighbourhood size, but one of the challenges is that although nearest neighbours graphs with constant number of neighbours for each vertex are directed graphs but the graph drawing algorithm that is chosen to be modified and extended to achieve the goal of this paper is designed for undirected graphs. A closely related discussion is the discussion of symmetry in the t-SNE paper~\cite{a4} when they describe the addition of symmetry as one of the major differences of t-SNE from its predecessor SNE~\cite{a78}. The distance transformation named \textit{neighbourhood-normalized distance transformation} that is introduced below at the very least is designed to make neighbourhood graphs more balanced where more balanced means less number of edges without an edge in opposite direction, and in practice shows to have a very effective effect.

Given a neighbourhood size $z$ (default value $z=20$), for each point $q_i$ in $Q$ a constant value $m_i$ is computed such that: $$tan^{-1}(\bar{\delta}_{i_z} \cdot m_i)=1$$ where $\bar{\delta}_{i_z}$ is the Euclidean distance or Cosine distance or a custom precomputed distance from the $z^{th}$ nearest neighbour of $q_i$ in the original space to $q_i$. Now the \textit{neighbourhood-normalized distance} is defined as:
\begin{equation}
\delta(q_i,q_j)=\frac{tan^{-1}(\bar{\delta}(q_i,q_j) \cdot m_i)+tan^{-1}(\bar{\delta}(q_i,q_j) \cdot m_j)}{2}
\label{eq:neighbourhood_normalized_distance_transformation}
\end{equation}
where $\delta(q_i,q_j)$ becomes the transformed distance between $q_i$ and $q_j$ and $\bar{\delta}(q_i,q_j)$ is the Euclidean distance or Cosine distance or a custom precomputed distance between $q_i$ and $q_j$. The repeating of $\bar{\delta}(q_i,q_j)$ in Equation~\ref{eq:neighbourhood_normalized_distance_transformation} is not a typo as it is described later. The custom precomputed distance mentioned above corresponds to the case where the input data is in distance matrix form.

\begin{figure}[h!]
 \centering
\includegraphics[width=6cm]{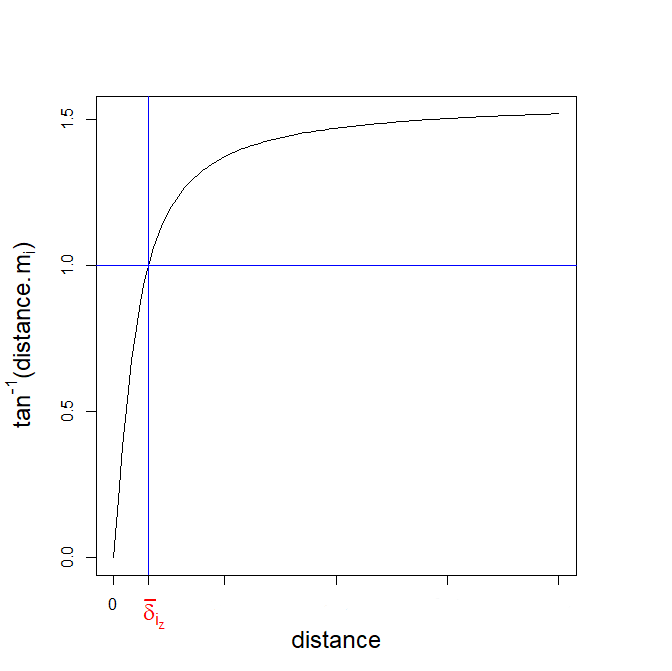}
 \caption{Computing $m_i$ for neighbourhood-normalized distance transformation}
 \label{fig:plot}
\end{figure}

While Fig.~\ref{fig:plot} is helpful in understanding the computation of $m_i$ for neighbourhood-normalized distance transformation, attention needs to be made on how density of distances around each data instance of original space is brought into play by $\bar{\delta}_{i_z}$ in a similar way but in a different paradigm that t-SNE computes and uses a per data instance standard deviation $\sigma_i$ of Gaussian distribution in order to bring density of distances around each data instance of original space into play. The role that the value $z$ plays is in analogy with perplexity for t-SNE. Needless to say, t-SNE being a statistical and probabilistic method, uses a different paradigm than the geometric neighbourhood-based graph drawing paradigm that is used in this paper but the requirements of a good embedding remain the same. One important observation to make is that in a nearest neighbours graph with constant number of neighbours for each vertex based on original space distances, when there is an edge without an edge in the opposite direction between the same pair of vertices, it is an indication of density change in original space, so in a more balanced neighbourhood graph where there are less number of edges without an edge in opposite direction, then in some regions, either density changes become smoother or a big separation replaces them depending on how the conflict of an edge without edge in opposite direction is resolved for some edges to either two edges in opposite direction or no edge at all.

While in most cases the original space distances before any transformation are symmetric, and the distinction between $\bar{\delta}(q_i,q_j)$ and $\bar{\delta}(q_j,q_i)$ wouldn't be relevant to Equation~\ref{eq:neighbourhood_normalized_distance_transformation}, in rare cases where the original space distances before transformation might be non-symmetric, if $\bar{\delta}(q_j,q_i) \cdot m_j$ was used instead of $\bar{\delta}(q_i,q_j) \cdot m_j$ in Equation~\ref{eq:neighbourhood_normalized_distance_transformation}, it would transform a non-symmetric distance matrix to a symmetric distance matrix, removing an important property of input data. However the distance transformation still brings non-symmetric distances closer to symmetric but not to completely symmetric.

The reason that the proposed method does not synchronize $z$ (the neighbourhood size of neighbourhood-normalized distance transformation) with the neighbourhood size for building the neighbourhood graph ( $\widehat{p}$ ), is that a change to $\widehat{p}$ would have brought a more dramatic change to the final embedding but the intention was to reduce the volatility of parameters.

\FloatBarrier

\subsubsection{Building the neighbourhood graph}
The third preliminary step is to build a neighbourhood graph $G$ based on transformed distances from the previous preliminary step choosing the $\widehat{p}$ nearest points as the neighbours of a point where $\widehat{p}$ is a constant parameter specifying the number of neighbours for each point for building the neighbourhood graph. The lower that $\widehat{p}$ is chosen, the more local the method becomes.

\subsection{Point Placement Using Force-Directed Graph Drawing}
\label{subsection:force_directed_graph_drawing}

In the four phases of the proposed algorithm, the~\cite{a10} force-directed graph drawing algorithm is used but several modifications are made to it before using it on the neighbourhood graph built in the preliminary step 3 and the transformed distances built in the second preliminary step.

\subsubsection{Brief review of the Fruchterman And Reingold \cite{a10} algorithm}
The algorithm in~\cite{a10} for drawing a graph in two dimensions, uses two different sets of forces based on the geometric distance in the drawing and adjacency of vertices of graph. The first set of forces is called repulsive forces and is applied to every pair of graph vertices. The second set of forces is called attractive forces and is applied only to adjacent pairs of vertices. Going through different iterations, assuming a unit time (in physics) between two iterations and unit mass for each vertex, the algorithm of \cite{a10} moves vertices based on forces but limited to a gradually decreasing temperature (temperature can be seen as the radius of a circle around the location of a vertex in previous iteration which the vertex is stopped on the edge of the circle in the current iteration if it tries to move outside the circle, limiting the amount of movement in each iteration for each vertex). The definitions of repulsive force ($\vec{f_r}$) imposed by any unordered pair of vertices on the each of the two vertices and attractive force ($\vec{f_a}$) imposed by each edge on each of its endpoints as they were defined in \cite{a10} but when brought to the conceptual context and notation of this paper, are in the form of:

$$\vec{f_r}=-\frac{\Gamma^2}{D_v(s_{i_t},s_{j_r})} \cdot \vec{\alpha}$$
and
$$\vec{f_a}=\frac{(D_v(s_{i_t},s_{j_r}))^2}{\Gamma} \cdot \vec{\alpha}$$
where
\begin{equation}
\vec{\alpha}=
\begin{cases}
\frac{s_{j_r}-s_{i_t}}{|s_{j_r}-s_{i_t}|} & \text{if the force is applied on }s_{i_t} \\   \\
\frac{s_{i_t}-s_{j_r}}{|s_{i_t}-s_{j_r}|} & \text{if the force is applied on }s_{j_r}
\end{cases}
\end{equation}
In the notation used, $s_{i_t}$ is a projection of a data instance in visual space and therefore a vertex of the neighbourhood graph, $s_{j_r}$ is a projection of a data instance in visual space and therefore a vertex of the neighbourhood graph, and $\Gamma$ is a constant optimal distance defined as:
$$\Gamma=\sqrt{\frac{width \times height}{NumberOfPoints}}$$
However while when bringing the concept of optimal distance from \cite{a10} to this paper, it would be natural to assume the $NumberOfPoints$ in definition of $\Gamma$ refers to number of projected points in visual space, that wouldn't be constant for the proposed algorithm of this paper, so instead a definition of $\Gamma$ is opted for where $NumberOfPoints$ refers to number of data instances in original space ($|Q|$) which is a constant, so the definition of $\Gamma$ becomes: 
\begin{equation}
\Gamma=\sqrt{\frac{width \times height}{|Q|}}
\end{equation}
Moreover the verdict of the controversy of whether edges of neighbourhood graph should be considered in a directed fashion or not is chosen by considering edges as directed and if there are two edges in opposite direction between two vertices, considering them separate edges and therefore applying separate forces for them. To decrease temperature the simplest form as mentioned by~\cite{a10} is to start with an initial temperature such as one tenth of the width of drawing and linearly decrease temperature so that temperature becomes zero at last iteration. We denote the maximum temperature by $\bar{u}$ and used the default value of $\bar{u}=100$ in the implementation.

\subsubsection{Modifications to the computation of forces and usage of masses in acceleration}
Although the graph layout algorithm of \cite{a10} is generally conceived as a good graph layout algorithm, it is not designed for the purpose of dimensionality reduction, so some modifications are proposed in this paper for that purpose.

The first modification that is proposed in this paper is to change the attractive forces for any edge $(s_{i_t}$,$s_{j_r})$ of the neighbourhood graph $G$ on $s_{i_t}$ or $s_{j_r}$ to the form:
\begin{equation}
\vec{f_a}=\psi \cdot \vec{\alpha}
\end{equation}
where
\begin{equation}
\psi=\left(\frac{D_v(s_{i_t},s_{j_r})}{\Gamma}\right)^{(1-b)}
\label{eq:psi}
\end{equation}
where the parameter $b$ is a \textit{visual density adjustment parameter} (default $b=0.9$ but can be adjusted in different scenarios). This modification which can increase attractive forces, changes the balance between attractive and repulsive forces in a way that is against two characteristics of the \cite{a10} algorithm, distributing vertices evenly and uniform edge lengths. Although distributing vertices evenly and uniform edge lengths are considered as good properties in general graph drawing literature, they are not good properties for dimensionality reduction, mainly because they can hide the inherent structure in the original space, one of the main goals of dimensionality reduction. While without this change the forces would have tried to achieve a uniform density connected visual space, this change can split the visual space into different sections as attractive forces can become more dense in one part of visual space than another part of visual space. In particular the parameter $b$, controls how much the density of edges in the induced subgraph of any subset of vertices of the neighbourhood graph should be reflected in the density of projected points (vertices) of that induced subgraph in the visual space, something that in the original attractive forces defined in~\cite{a10} correctly was assumed to be irrelevant to the objective of their paper for graph drawing literature. While the induced subgraph can be any induced subgraph of $G$, if the vertices of the induced subgraph correspond to a semantic group for an application of dimensionality reduction, then if the density for the part of visual space corresponding to those vertices is different from the density of surrounding area, a separate visual group is perceived, the ultimate goal in many applications of dimensionality reduction. To find a good value for $b$ quantitative metrics, number of projected points on the frame border, type of data set, size of data set, or visual quality may be used but quantitative metrics that do not rely on classes such as trustworthiness, or number of projected points on the frame border, may be more practical when dealing with data that the classes are not known or the type of data set is not familiar.

The second modification that is proposed in this algorithm is to some degree allow the exact value of distances in the original space (after transformation) affect the forces in the visual space already defined based on the neighbourhood graph, the distances in the visual space and $\Gamma$. The reason for that is that although the neighbourhood graph is built based on the transformed distances it does not retain the exact value of transformed distances and therefore is blind to relative importance of neighbours. So a multi-objective approach is pursued here were the embedding of neighbourhood graph has a higher priority but relative distance preservation can fine tune but not dramatically change the embedding. Therefore in the second modification, attractive forces are modified again to the form
$$\vec{f_a}= \widehat{\psi} \cdot \vec{\alpha}$$
where
$$
\widehat{\psi}= \psi +
\begin{cases}
\min\left(\frac{\left|\psi\right|}{2},h\right) & \text{if } h > 0 \\   \\ 
\max\left(\frac{\left|\psi\right|}{-2},h\right) & \text{if } h \le 0
\end{cases}
$$
where $h$ is a force defined based on distances in the original space (after transformation) and $h$ is in the form $h=\frac{\delta(q_i,q_j)}{\delta_{max}}-\frac{D_v(s_{i_t},s_{j_r})}{D_{v_{max}}}$. Definition of $h$ somehow borrows ideas from force scheme algorithm of \cite{a2} where distances of original space are the core of force computations. In the implementation of this algorithm, $D_{v_{max}}$ is only computed once after initial random layout and used it for subsequent iterations too. It is important to note that in the second modification, the effect of $h$ on the attractive force is limited to only half of the attractive force obtained in the previous modification, in order to stop it from dramatically changing the embedding. While neighbourhood graph creates the substance of the embedding, exact value of transformed distances help fine tune the embedding with a limited scope designed to keep significance of effect of neighbourhood graph on the embedding intact.

A third relevant modification but not to the forces is that while the algorithm of \cite{a10} assumes unit mass for each vertex for computing acceleration of a vertex from aggregate forces on a vertex, when calculating acceleration by attractive forces, the proposed algorithm opts to start from unit mass for each vertex (projected point) but changing it when duplicating a vertex(projected point) to account for lower number of edges on a vertex in the process of duplicating that is discussed later in this paper. The mass for each projected point $s_{i_t}$ is denoted as $\omega_{i_t}$ initially set to 1 for each projected point. While in \cite{a10} there was no need of explicit mention of mass in computing acceleration from a force using the Newton's second law of motion, this paper needs to bring mass explicitly to the calculation of acceleration from forces and therefore the acceleration of a vertex (projected point) $s_{i_t}$ by an attractive force $\vec{f_a}$ is calculated in the proposed algorithm in this paper as $\vec{\Delta}=\frac{\vec{f_a}}{\omega_{i_t}}$. For acceleration of a vertex (projected point) by a repulsive force $\vec{f_r}$, a unit mass is assumed for each vertex because the repulsive forces are not based on the edges of neighbourhood graph and therefore the changes made to the neighbourhood graph is not relevant to them.

While Procedure~\ref{procedure:applyRepulsiveForces} summarizes how repulsive forces are computed and applied to projected points on each iteration, Procedure~\ref{procedure:applyAttractiveForces} summarizes how attractive forces are computed and applied to projected points on each iteration.

\floatstyle{boxed}
\newfloat{procedureFloat}{htbp}{prc}
\floatname{procedureFloat}{Procedure} 

\begin{procedureFloat*}
\raggedright
\SetAlgoLined
    
    \For{$i \in \{1,2,...,|Q|\}$}
    {
        \For{$j \in \{1,2,...,|Q|\}$}
        {
            \For{$s_{i_t} \in S_i$}
            {
            	$T_{i_t}=(0,0)$ \:\:\: Comment: $T_{i_t}$ is a 2D vector.
            	
                \For{$s_{j_r} \in S_j$ and $s_{j_r} \neq s_{i_t}$}
                {
					\If{$s_{i_t}$ is not frozen and $s_{j_r}$ is not ineffective}
					{
						$T_{i_t}=T_{i_t} - \frac{\Gamma^2 \cdot (s_{j_r} - s_{i_t})}{|s_{j_r} - s_{i_t}|^2}$
					}
                }
                
                \eIf{$|T_{i_t}| \le temperature$}
                {
                	$s_{i_t} = s_{i_t} + T_{i_t}$
                }
                {
                	$s_{i_t} = s_{i_t} + \frac{T_{i_t}}{|T_{i_t}|} \cdot temperature$
                }
                
                \If{there is a border frame and $s_{i_t}$ is outside the frame}
                {
                	Bring $s_{i_t}$ on the frame.
                }
            }
        }
    }

\caption{Summarizes how repulsive forces are computed and applied to projected points on each iteration.}
\label{procedure:applyRepulsiveForces}
\end{procedureFloat*}

\begin{procedureFloat*}
\scriptsize
\raggedright
\SetAlgoLined
    \For{$i \in \{1,2,...,|Q|\}$}
    {
        \For{$s_{i_t} \in S_i$}
        {
        	$T_{i_t}=(0,0)$ \:\:\: Comment: $T_{i_t}$ is a 2D vector.
        }
    }
    
    \For{$i \in \{1,2,...,|Q|\}$}
    {
        \For{$s_{i_t} \in S_i$}
        {
            \For{$s_{j_r}$ in neighbors of $s_{i_t}$ in $G$}
            {
                \If{$s_{i_t}$ is not frozen and $s_{j_r}$ is not ineffective}
                {
                    \BlankLine
					
					$\psi= \left(\frac{D_v(s_{i_t},s_{j_r})}{\Gamma}\right)^{(1-b)}$
					
					\BlankLine
					
					$h=\frac{\delta(q_i,q_j)}{\delta_{max}}-\frac{D_v(s_{i_t},s_{j_r})}{D_{v_{max}}}$
					
					\BlankLine
					
                    \eIf{$h>0$}
                    {
                        $\widehat{\psi} = \psi + \min(\frac{|\psi|}{2},h)$
                    }
                    {
                        $\widehat{\psi} = \psi + \max(\frac{-|\psi|}{2},h)$
                    }
                    
                    $T_{i_t} = T_{i_t} + \frac{\widehat{\psi} \cdot (s_{j_r}-s_{i_t})}{\omega_{i_t} \cdot |s_{j_r} - s_{i_t}|}$

                    \BlankLine
                    
                    $T_{j_r} = T_{j_r} + \frac{\widehat{\psi} \cdot (s_{i_t}-s_{j_r})}{\omega_{j_r} \cdot |s_{i_t}-s_{j_r}| }$
                }
            }
        }   
    }
    
    \For{$i \in \{1,2,...,|Q|\}$}
    {
        \For{$s_{i_t} \in S_i$}
        {
        	\eIf{$|T_{i_t}| \le temperature$}
            {
            	$s_{i_t} = s_{i_t} + T_{i_t}$
            }
            {
            	$s_{i_t} = s_{i_t} + \frac{T_{i_t}}{|T_{i_t}|} \cdot temperature$
            }
            
			\If{there is a border frame and $s_{i_t}$ is outside the frame}
			{
				Bring $s_{i_t}$ on the frame.
			}
        }
    }

\caption{Summarizes how attractive forces are computed and applied to projected points on each iteration.}
\label{procedure:applyAttractiveForces}
\end{procedureFloat*}

\subsubsection{Computing replication pressure in phase 2 and modifying the neighbourhood graph in phase 4}
\label{subsubsection:replication_pressure_and_modifying_neighbourhood_graph}
\begin{figure}[h!]
 \centering
\includegraphics[width=8cm]{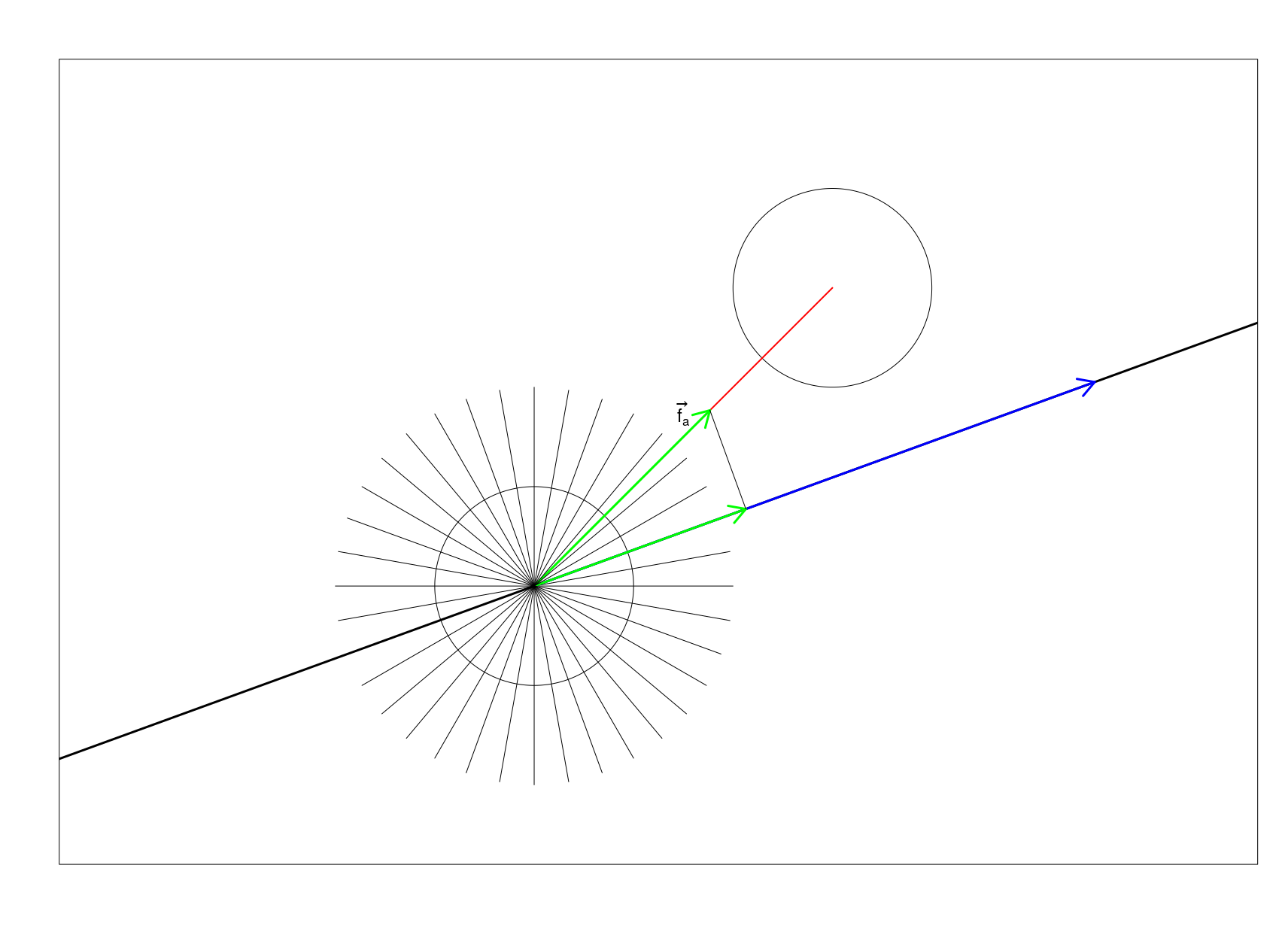}
 \caption{The 36 radial axes around a vertex (point) in G. Perpendicular projection of $\vec{f_a}$ on the blue axis. (Perpendicular projection of $\vec{f_r}$ is similar)}
 \label{fig:drawing}
\end{figure}

In phase 2 to be described in section~\ref{section:phase_two} and phase 4 to be described in section \ref{section:phase_four}, in each iteration a replication pressure per projected point is computed based on the attractive and repulsive forces on that projected point and 36 different possibly non-orthogonal axes on the projected point selected based on a circle around the projected point (see Fig. \ref{fig:drawing}) regularly organized $10^{\circ}$ apart radially from the previous one and the origin of all set on on the projected point. For each axis, all force vectors on that projected point are first perpendicularly projected onto that axis and sum of magnitude of projected vectors that are on the positive side of the axis, is considered as the positive pressure on that axis and also the sum of magnitude of projected vectors that are on the negative side of the axis is considered as the negative pressure on that axis. The replication pressure of the axis on the projected point is defined as the aggregate sum of the positive pressure on the axis and the negative pressure on that axis. The replication pressure of the projected point is then computed as the maximum of replication pressures among all of the 36 axis on the projected point. In phase 4 to be described in section~\ref{section:phase_four}, when trying to duplicate a projected point of the gray layer, the neighbourhood graph is modified based on the axis which produced the maximum replication pressure. When a projected point is duplicated, the projected point that was duplicated is denoted as the first point of duplication and the projected point that was added is denoted as the second point of duplication meaning that the first one is duplicated so that the second one becomes another mapping of the same data instance initially located at the same location in visual space. The neighbours of first point of duplication that are in one of the half planes resulted from cutting the plane with a line perpendicular to the axis that produced the maximum pressure and incident on the first point of duplication, are now considered as the neighbours of the second point of duplication instead. While in the process of duplication the location of the first point of duplication remains unchanged, the location of the second point of duplication is moved to the centroid of its neighbours and after that any neighbour of the first point of duplication which is closer to the second point of duplication is also considered as the neighbour of the second point of duplication instead. If the point to be duplicated has $c_1$ neighbours in the neighbourhood graph before duplication and first point of duplication has $c_2$ neighbours in the neighbourhood graph after duplication and second point of duplication has $c_3$ neighbours in the neighbourhood graph after duplication, then the value of the mass $\omega$ for the first point of duplication for applying attractive forces is multiplied by $\frac{c_2}{c_1}$ compared to before duplication and the value of the mass $\omega$ for the second point of duplication for applying attractive forces is multiplied by $\frac{c_3}{c_1}$ compared to before duplication. If modifying the neighbourhood graph results in no neighbours either for the first point of duplication or the second point of duplication, the duplication is considered as a failed duplication and neighbourhood graph is not changed by a failed duplication as if no attempt was made to duplicate the point.

\subsubsection{The four phases}
After the three preliminary steps, the LVSDE algorithm starts the modified force-directed graph drawing through four phases spanning over a number of iterations (default 1830) starting with a random embedding with exactly one projection per data instance in which all projected points are in the red layer and no frozen projected point exist where a frozen projected point means a projected point designated to have no movement and no effect on the movement of other projected points unless the designation is removed. While the first phase is meant to build the general structure of the embedding, the second phase tries to identify the projected points to be moved to the gray layer, freezes them and marks them as ineffective, phase 3 marks the projected point in gray layer as effective and reverses the freezing where reversing freezing means the projected points in the red layer become frozen and the projected points in the gray layer are unfrozen, and the forth phase is were the magic of duplication happens on some or all of the projected points of the gray layer. Once a projected point is designated as ineffective not only it does not move but also it does not apply force on any other projected point while the term freezing a projected point means the projected point does not move but does not necessarily mean that it does not apply forces to other projected points. An ineffective projected point is a frozen projected point but a frozen projected point is not necessarily a an ineffective projected point.

\subsubsection{Phase one: simple graph drawing}
Phase one is where exactly one projection per data instance is used and runs for a number of iterations (default 500) but temperature is reduced in the modified force-directed algorithm as if there are more iterations in this phase to allow more freedom toward the end (By default temperature in phase one is $\frac{(1000-\mu)\bar{u}}{1000}$ where $\mu$ is the iteration number in current phase). First iteration starts with a random location of projected points in the visual space and all in the red layer none of them frozen or ineffective, and ends in a drawing which ideally should have balanced forces. The next three phases try to improve the drawing resulted in the phase 1. Therefore in the first phase a border frame is not used in visual space, in order to allow the general structure of the embedding to be shaped in phase one and then at the very beginning of the second phase a frame slightly bigger than the current embedding is put around the embedding which will stay until the end of the algorithm so that the extra freedom given in the phases 2 to 4 do not enlarge the size of the embedding so high that it requires the embedding to be scaled significantly affecting the visual perception of the embedding negatively.

\subsubsection{Phase two: selecting gray points and freezing them}
\label{section:phase_two}
The second phase starts after the first phase is finished and at the beginning of the second phase after a frame slightly bigger than the current embedding is put around the embedding. In the beginning of each iteration of second phase, a replication pressure is computed for each projected point as was discussed in Section~\ref{subsubsection:replication_pressure_and_modifying_neighbourhood_graph}. At the first iteration of of second phase, after replication pressures for the first iteration are computed, the maximum number of points in gray layer is chosen. While a good measure for computing replication pressure is achieved in this paper, the discussion on the ideal maximum number of data instances in gray layer is a more open question and just to fill the gap of necessity rather than forming the ideal, the maximum number of data instances in gray layer is computed by the minimum of two threshold. The first threshold is calculated by counting the number of projected points whose replication pressure is outside the range centred at the mean of replication pressures of all projected points and expanded in each direction 1.2 times standard deviation of replication pressure of all projected points where projected points are considered from the embedding at the very beginning of the second phase. While bidirectional style of this threshold might seem unnatural, it is meant to address the cases where distribution of the replications pressures is twisted toward one direction in a non-symmetric way from the mean. The second threshold which is the most important one is the number of data instances divided by 4, in order to ensure that the number of data instances in the red layer is not less than 75\% of the total number of data instances embedded. In the second phase at each iteration a projected point with highest replication pressure is chosen and marked as ineffective until as many as the maximum number of data instances in gray layer are marked as ineffective. Once a projected point is designated as ineffective not only it does not move but also it does not apply force on any other projected point while the term freezing a projected point means the projected point does not move but does not necessarily mean that it does not apply forces to other projected points. An ineffective projected point is a frozen projected point but a frozen projected point is not necessarily a an ineffective projected point. In second phase when a projected point is marked as ineffective, it is also moved it to gray layer. A projected point in gray layer remains in gray layer until the end of the algorithm but whether it is marked as ineffective or not can change in next phases. The second phase continues for a number of iterations (default 450) and temperature is adjusted as if the iterations interval is almost second half of a higher number of iterations (By default temperature in phase two is $\frac{(1000-(\mu+500))\bar{u}}{1000}$ where $\mu$ is the iteration number in current phase).

\subsubsection{Phase 3: reversing freezing and marking gray points as effective}
In the third phase, at the very beginning of the third phase, every projected point that was marked as ineffective in the second phase is marked as effective again but the rest of projected points are frozen. This also implies reversing the freezing as the projected points in red layer become frozen but effective and the projected points in gray layer are unfrozen. The third phase continues for a number of iterations (default 390) and temperature is adjusted as if the iterations interval is almost second half of a higher number of iterations (By default temperature in phase three is $\frac{(1000-(\mu+510))\bar{u}}{1000}$ where $\mu$ is the iteration number in current phase).

\subsubsection{Phase 4: replicating gray points when possible}
\label{section:phase_four}
At the beginning of forth phase, in the first iteration of phase 4, an attempt is made to duplicate each of the projected points of the gray layer. When duplicating a projected point, the new added point is also set to be in gray layer. When duplicating, the neighbourhood graph is also modified as was discussed in Section~\ref{subsubsection:replication_pressure_and_modifying_neighbourhood_graph}. Not all duplication are successful and a failed duplication may happen when trying to modify the neighbourhood graph. Failed duplication does not change the embedding or neighbourhood graph. In the proposed algorithm a maximum of 2 projections per data instance limit is put on the number of projections per data instance but Strict Red Gray Embeddings in general are not required to adhere to that restriction. The force-directed graph drawing continues in the forth phase through a number of iterations (default 490) and temperature is adjusted as if the iterations interval is almost second half a higher number of iterations (By default temperature in phase two is $\frac{(1000-(\mu+510))\bar{u}}{1000}$ where $\mu$ is the iteration number in current phase).

\subsection{Choosing One Of The Iterations As The Result}
While in absence of any indication for choosing a different iteration, the last iteration should be perceived as the ultimate result of the algorithm, nothing stops choosing a different iteration of the algorithm including for example the iteration that performs best with respect to a specific evaluation measure.

\subsection{Parallel speedup}
A parallel version of the proposed algorithm is possible and is implemented which scales well with increase in the number of CPU cores (Finding how to make the algorithm parallel should be easy for a computer science graduate student as it just requires making two for loops parallel).

\subsection{Parameters}
The three parameters visual density adjustment parameter ( $b$ ), the number of neighbours for building neighbourhood graph ( $\widehat{p}$ ), and whether to use UMAP to 30 dimensions as a preliminary step can significantly change the outcome of the algorithm. While visual density adjustment parameter can be freely set to any value, at least one of the values from the set $\{-0.9, -0.5, -0.1, 0.1, 0.5, 0.9\}$ is usually good for this parameter unless there is no good value, so practically the search for a good value usually does not have to test a lot of values. While the number of neighbours for building neighbourhood graph can be freely set to any value, at least one of the values from the set $\{10, 20, \frac{|Q|}{3}, \frac{|Q|}{4}, \frac{|Q|}{5}\}$ where $|Q|$ is the number of data instances in original space is usually good for this parameter unless there is no good value, so practically the search for a good value usually does not have to test a lot of values. This way by testing the outcome of a total of 60 configuration for the three parameters optimizing for a quantitative criteria or a qualitative criteria which may or may not use class labels, usually a good embedding is found unless LVSDE is incapable of producing a good embedding. For text related high dimensional data sets, whether to use Euclidean distance or Cosine distance is a consideration that needs to be added to the above discussion as Cosine distance in general is known to usually perform better on text related high dimensional data.

\setlength{\fboxsep}{0pt}

\begin{figure*}[!ht]
\centering
 \subfigure[End of phase 1]{\fbox{\includegraphics[width=.21\linewidth, trim=-1pt -1pt 0 -1pt]{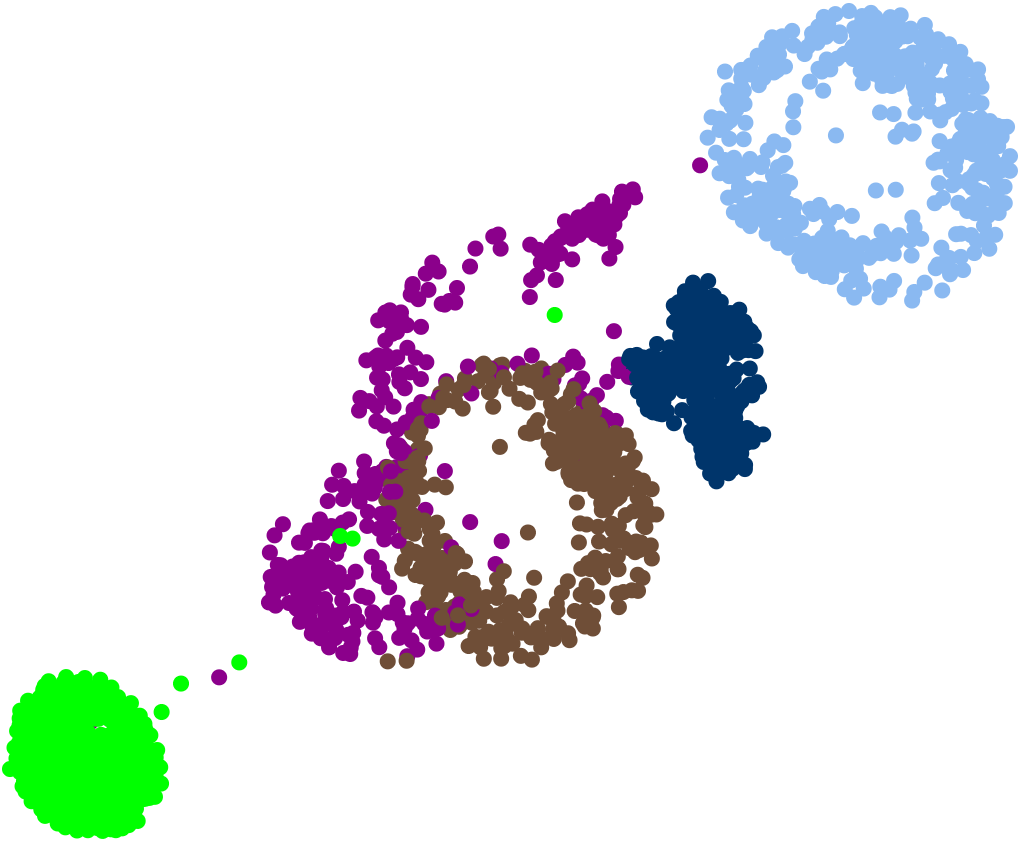}}\label{fig:}}\quad
 \subfigure[End of phase 2]{\fbox{\includegraphics[width=.21\linewidth, trim=-2pt -2pt -2pt -2pt]{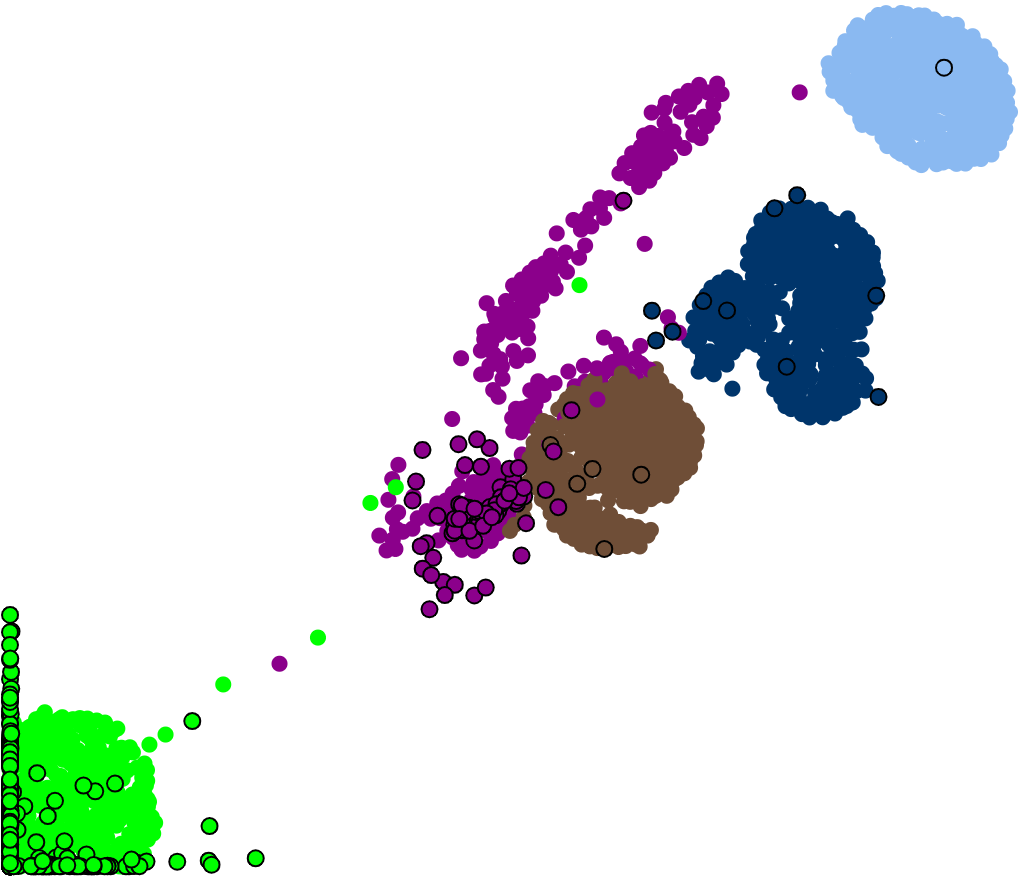}}\label{fig:}\quad}
 \subfigure[End of phase 3]{\fbox{\includegraphics[width=.21\linewidth, trim=-2pt -2pt -2pt -2pt]{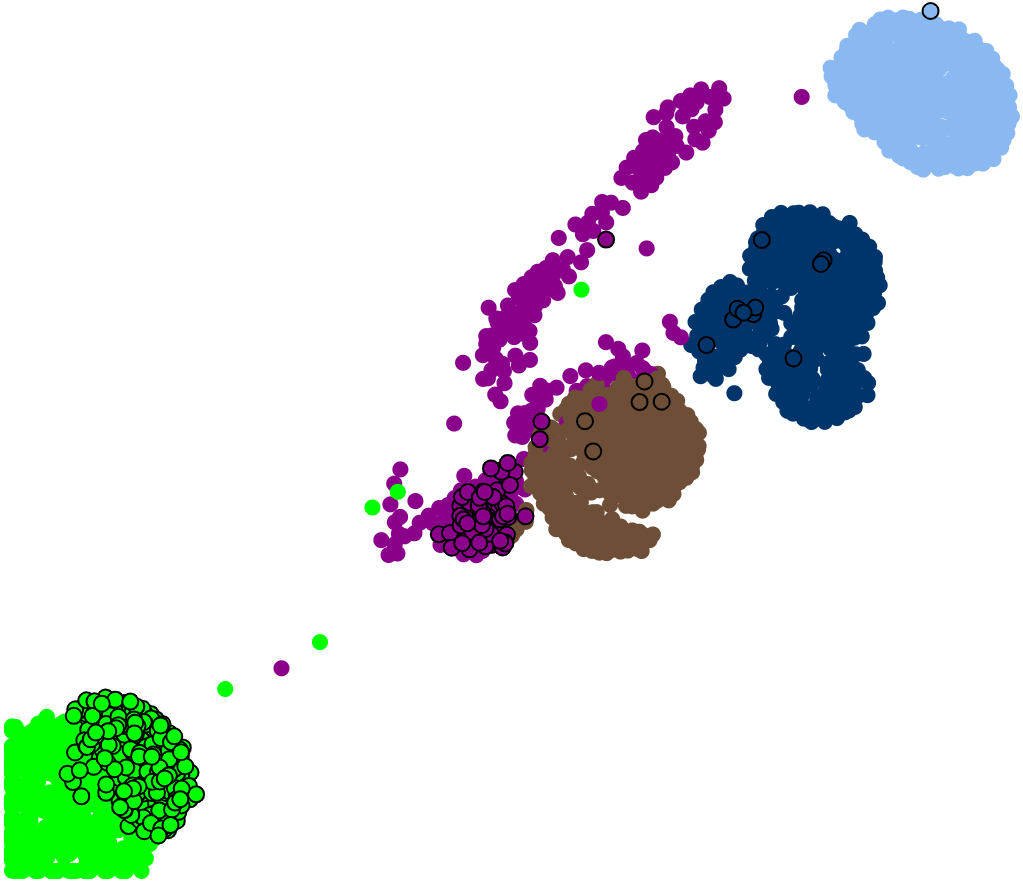}}\label{fig:}\quad}
 \subfigure[End of phase 4]{\fbox{\includegraphics[width=.21\linewidth, trim=-2pt -2pt -2pt -2pt]{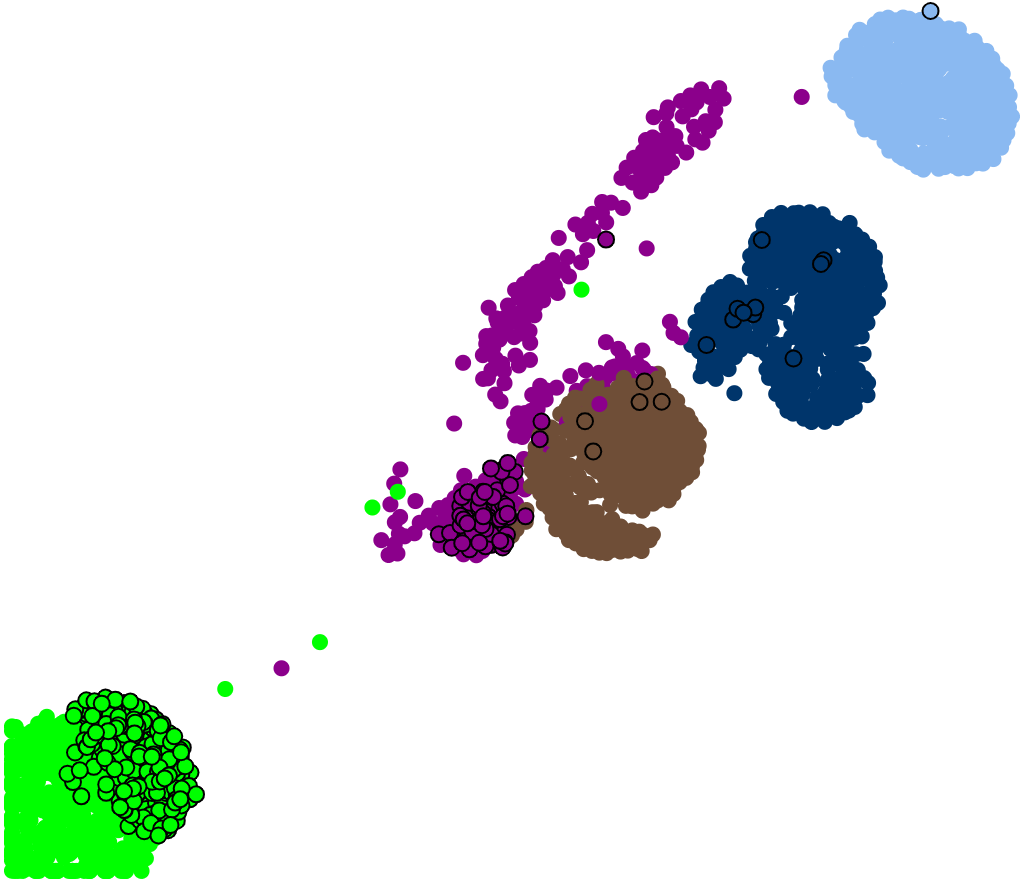}}\label{fig:}\quad}
\caption{An LVSDE embedding of the 1000 genomes project data set~\cite{a51_1000_genomes} distances at the end of each phase of LVSDE. Points with a black circle around them are in gray layer. Points with a black dot inside them are second projection of an input data instance. Points without a black circle around them are in red layer.}
\label{fig:phases}
\end{figure*}

\subsection{Example of phases outcome on a real data set}
Fig.~\ref{fig:phases} shows an outcome of the phases of LVSDE on the 1000 genomes project data set~\cite{a51_1000_genomes} distances at the end of each phase of LVSDE.

\FloatBarrier

\section{Experimental Setup}
\label{sec:experimental_setup}

\subsection{Data Sets}
In this paper four different data sets are considered. First is the MeefTCD \cite{a93} enhanced email features data set and has 2400 points and 48557 dimensions and is labelled with 8 classes described in \cite{a93}. Second is the MNIST data set of handwritten digits but its test data is not used and not necessarily all rows of the training data is used in all stages of techniques, evaluation or display (details in subsection \ref{subsec:experimental_setup_details}). The third data set is the 1000 genomes project~\cite{a51_1000_genomes} data set consisting DNA samples from 2504 individuals in different locations in the world (The data from the third phase of the 1000 genomes project is used and converted to a distance matrix using the version 2.0 of the PLINK~\cite{a52_plink,a53_plink} bioinformatics toolkit which has an implementation of \cite{a94} method for producing a similarity relationship matrix followed by a simple linear transformation to produce non-negative dissimilarity distances -- maximum similarity minus the similarity as distance). The forth data set is the IRIS data set classifying 3 different classes of the Iris plant.

\subsection{Experimental Setup Details}
\label{subsec:experimental_setup_details}
The proposed method (\textit{LVSDE}) is evaluated in this paper in comparison to UMAP \cite{a30,a31} and t-SNE \cite{a4} Barnes-Hut variant \cite{a95} on the four data sets. For UMAP \cite{a30,a31}, the umap-learn 0.5.3 python implementation of UMAP \cite{a30,a31} is used. For all of the experiments of this paper whenever t-SNE \cite{a4} is mentioned, the Barnes-Hut variant \cite{a95} of it is used.

For the 1000 genomes project data set, for \textit{LVSDE} method, two different configurations are reported on and the difference between them is discussed. For the first configuration, the parameters that are used are visual density adjustment parameter $b=0.1$, one third of number of data instances as the number of neighbours in the initial neighbourhood graph, and opting not to use UMAP to 30 dimensions. For the second configuration, the parameters that are used are visual density adjustment parameter $b=0.5$, $20$ as the number of neighbours in the initial neighbourhood graph, and opting not to use UMAP to 30 dimensions. For 1000 genomes project~\cite{a51_1000_genomes}, the data from the third phase of the 1000 genomes project is used and converted to a distance matrix using the version 2.0 of the PLINK~\cite{a52_plink,a53_plink} whole genome analysis toolkit which has an implementation of \cite{a94} method for producing a similarity relationship matrix followed by a simple linear transformation to produce non-negative dissimilarity distances ( maximum similarity minus the similarity as distance), and the distance matrix is used as the input to all methods.

For the MNIST data set, for \textit{LVSDE} method the parameters that are used are visual density adjustment parameter $b=0.1$, and $20$ as the number of neighbours in the initial neighbourhood graph, opting to use UMAP to 30 dimensions and using Euclidean distance for UMAP to 30 dimensions and Euclidean distance to convert 30 dimensional data of UMAP output to distances. For the MNIST data set, for \textit{LVSDE}, all MNIST training data are embedded into 30 dimensions but only the first 2000 rows are picked for the rest of algorithm, evaluation and display. Similar is done for UMAP where UMAP is applied to all training data but then only the first 2000 rows are picked for evaluation and display. Similar is done for t-SNE where t-SNE is applied to all of training data but then only the first 2000 rows are picked for evaluation and display. For all methods, Euclidean distance is used as the distance measure.

For the IRIS data set the parameters that are used for \textit{LVSDE} are visual density adjustment parameter $b=-0.1$, and $20$ as the number of neighbours in the initial neighbourhood graph, and opting not to use UMAP to 30 dimensions. Euclidean distance is used to convert the input 4-dimensional data into distances. For all methods, Euclidean distance is used as the distance measure.

For the MeefTCD email features data set, for \textit{LVSDE} method, the parameters that are used were visual density adjustment parameter $b=0.5$, $40$ as the number of neighbours in the initial neighbourhood graph, opting to use UMAP to 30 dimensions and using Cosine distance as the distance measure for UMAP to 30 dimensions preliminary step but after UMAP to 30 dimensions, Euclidean distance is used. For t-SNE and UMAP Cosine distance is used as the distance measure.

\section{Empirical Results And Comparison}
\label{sec:results}

\begin{figure*}[!ht]
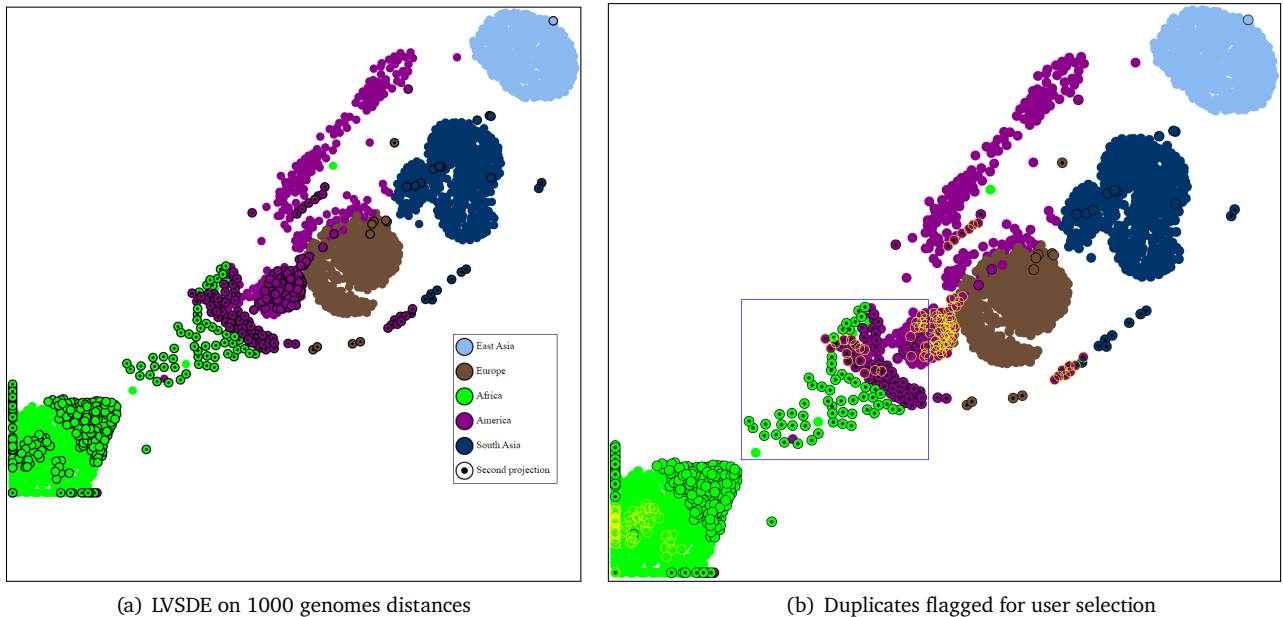

\centering
 \subfigure[LVSDE on 1000 genomes distances]{\fbox{\includegraphics[width=.41\linewidth, trim=-1pt -1pt 0 -1pt]{overall/images/polished/006.png}}}\quad
 \subfigure[Duplicates flagged for user selection]{\fbox{\includegraphics[width=.48\linewidth, trim=-2pt -2pt -2pt -2pt]{overall/images/polished/007.png}}\quad\quad}\quad
\caption{(a) An LVSDE embedding of the 1000 genomes project data set~\cite{a51_1000_genomes} distances. Points with a black circle around them are in gray layer. Points with a black dot inside them are second projection of an input data instance. Points without a black circle around them are in red layer. Duplication of some of points of America from some area close to Europe to an area close to duplicates of some of points of Africa in an area of visual space between Europe and Africa, matches immigration from Europe and Africa to America. LVSDE has successfully selected points to duplicate and successfully guided them in a meaningful way. For LVSDE configuration 1 is used.
(b) Yellow circles specify points that have a corresponding duplicate point in the blue rectangle meaning that for each point with yellow circle around it there is a point in the blue rectangle which is another projection of the same data instance of original space. The blue rectangle is specified by the user.
}
\label{fig:1000_genomes_distances_embedding_LVSDE}
\end{figure*}

\begin{figure*}[!ht]
\centering
 \subfigure[LVSDE 1 on 1000 genomes distances]{\fbox{\includegraphics[width=.34\linewidth, trim=-1pt -1pt 0 -1pt]{overall/images/polished/006.png}}\quad\quad}\quad
 \subfigure[LVSDE 2 on 1000 genomes distances]{\fbox{\includegraphics[width=.34\linewidth, trim=-1pt -1pt 0 -1pt]{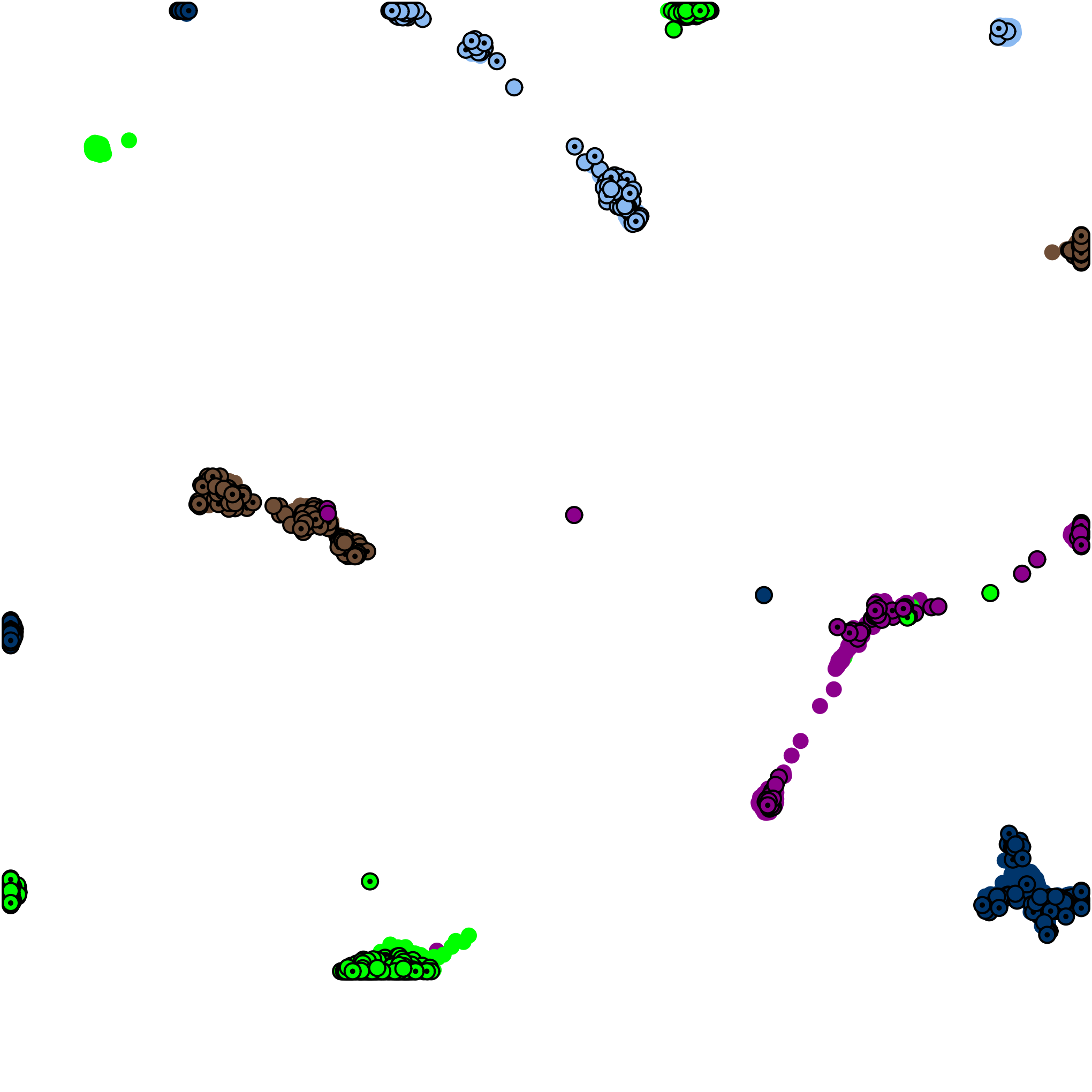}}\quad\quad}\quad
 \subfigure[LVSDE 2 on 1000 genomes distances]{\fbox{\includegraphics[width=.34\linewidth, trim=-1pt -1pt 0 -1pt]{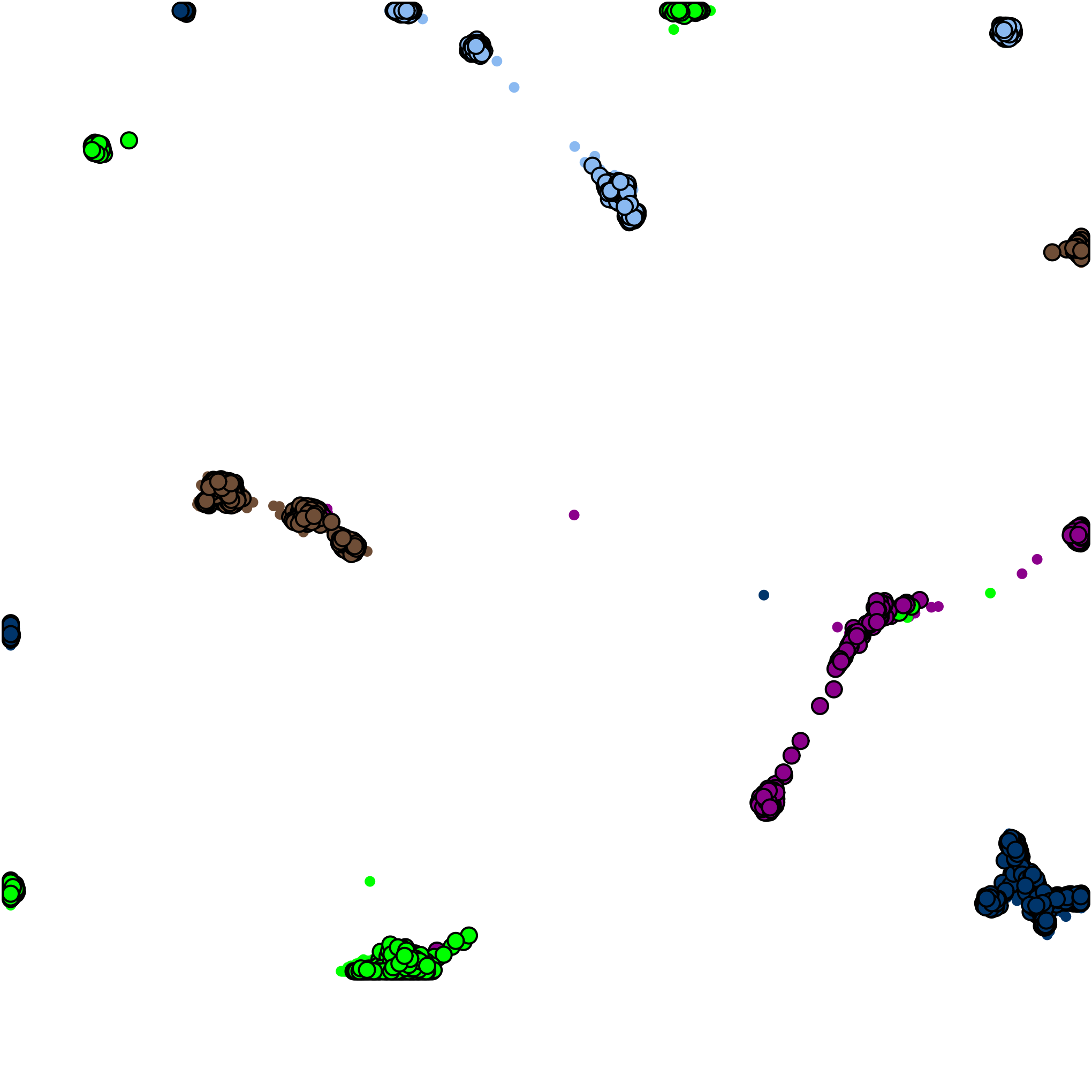}}\quad\quad}\quad
\caption{Two configurations of LVSDE used for embedding the 1000 genomes project data set~\cite{a51_1000_genomes} distances. While configuration 1 performs better semantically, configuration 2, like UMAP and t-SNE artificially performs better in groups separation, however knowing the phylogenetic connection of human genomes that separation has less semantic harmony with the reality of significant connection between human genomes. Configuration 1 although being better does not show the ideal representation of connection human genomes either and for future work it is an interesting research area to improve that in order to have an embedding method that shows the connection of human genomes better and more inclusive. The shortcoming in representation of that connection is an important flaw in the current state dimensionality reduction methods improved by LVSDE but not to the full extent. In part a and b, points with a black circle around them are in gray layer and points with a black dot inside them are second projection of an input data instance and points without a black circle around them are in red layer. In part c, points with a black circle around them are in the red layer and the points in the gray layer are drawn smaller.}
\label{fig:two_configs}
\end{figure*}

\begin{figure*}
\tiny
 \centering
 \subfigure[{\scriptsize LVSDE on MeefTCD}]{\fbox{\includegraphics[width=.32\linewidth,height=4.5cm]{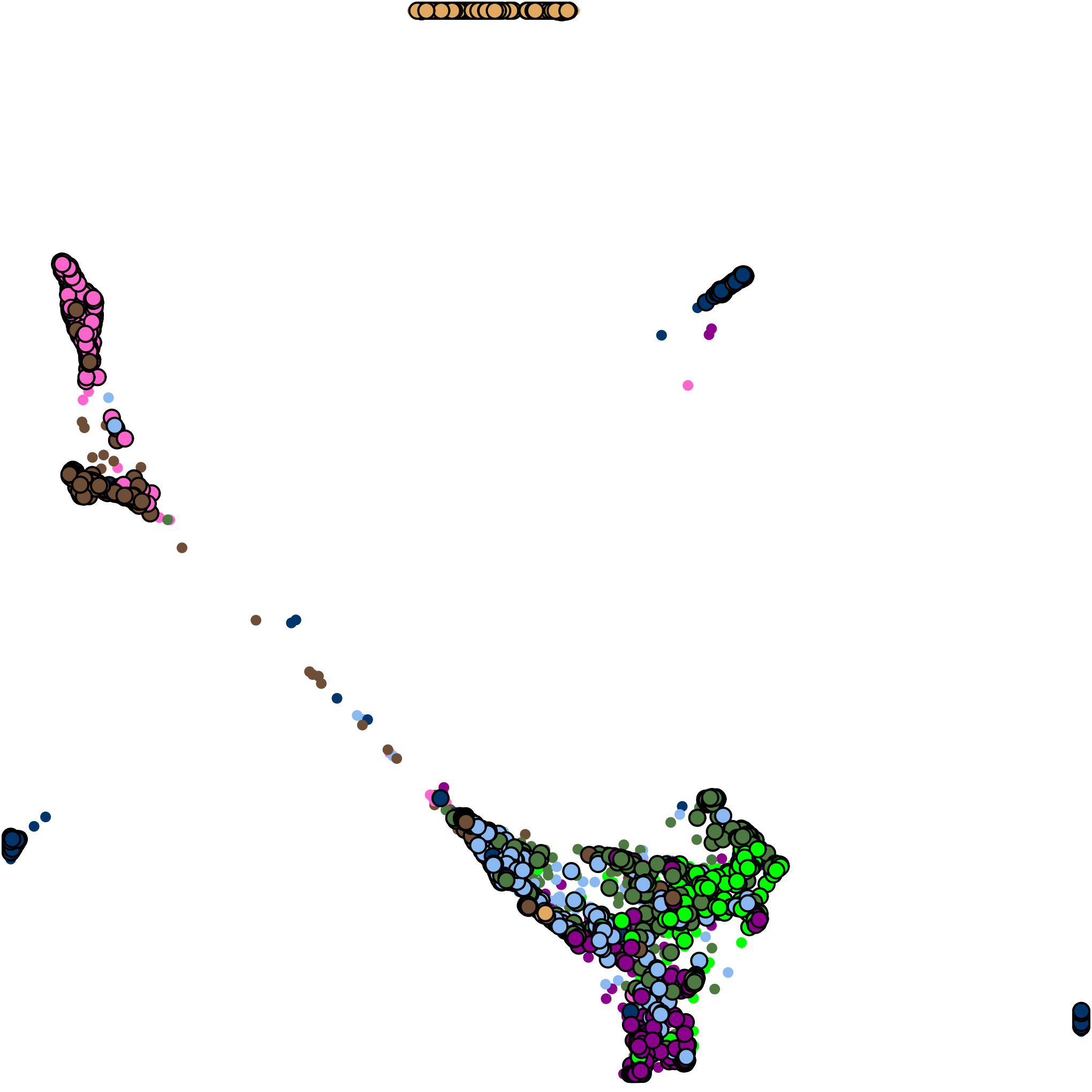}}}
 \subfigure[{\scriptsize LVSDE on a subset of MNIST}]{\fbox{\includegraphics[width=.32\linewidth,height=4.5cm]{overall/images/polished/041.png}}}
 \subfigure[{\scriptsize LVSDE 1 on 1000 genomes project}]{\fbox{\includegraphics[width=.32\linewidth,height=4.5cm]{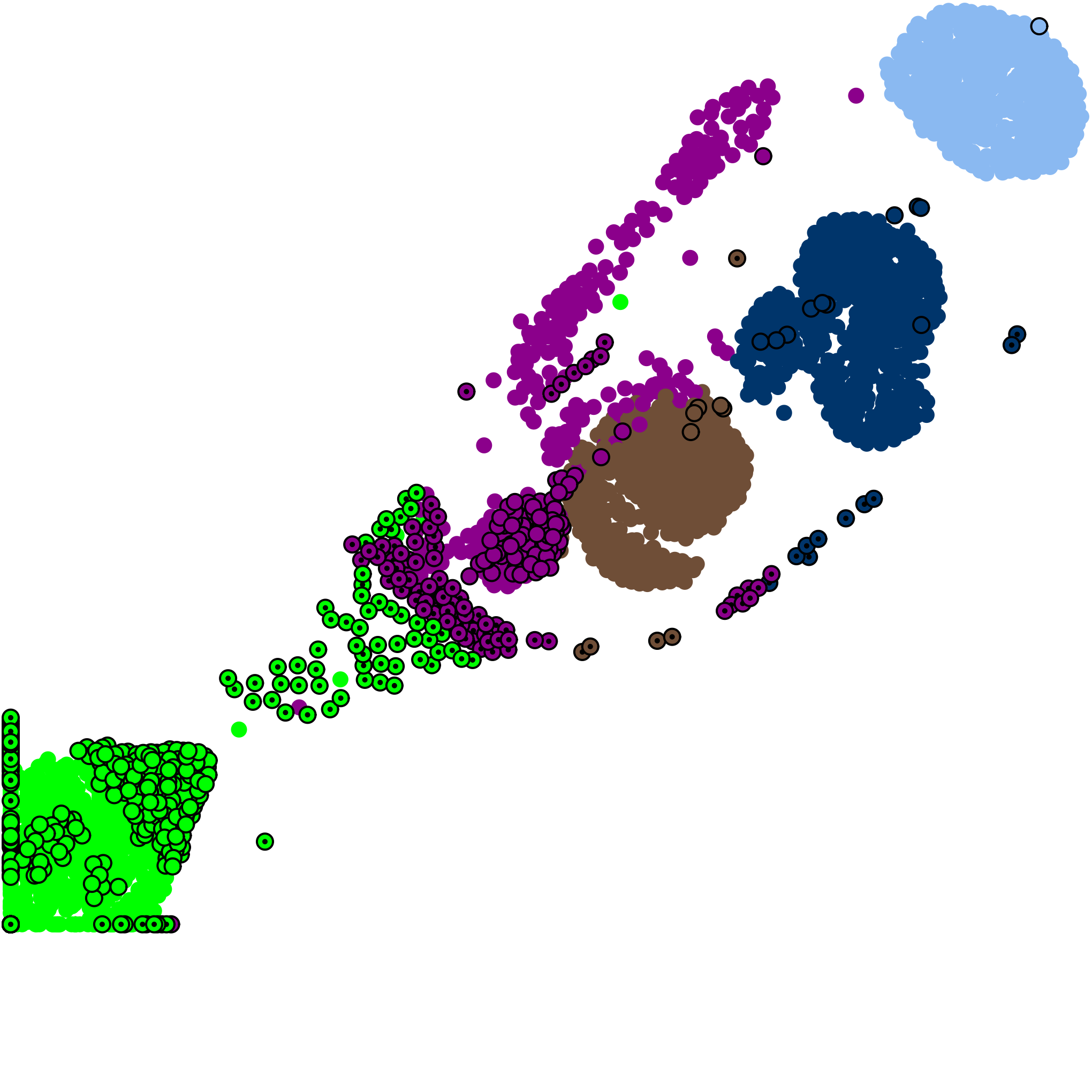}}}\\
 \subfigure[{\scriptsize UMAP on MeeefTCD}]{\fbox{\includegraphics[width=.32\linewidth,height=4.5cm]{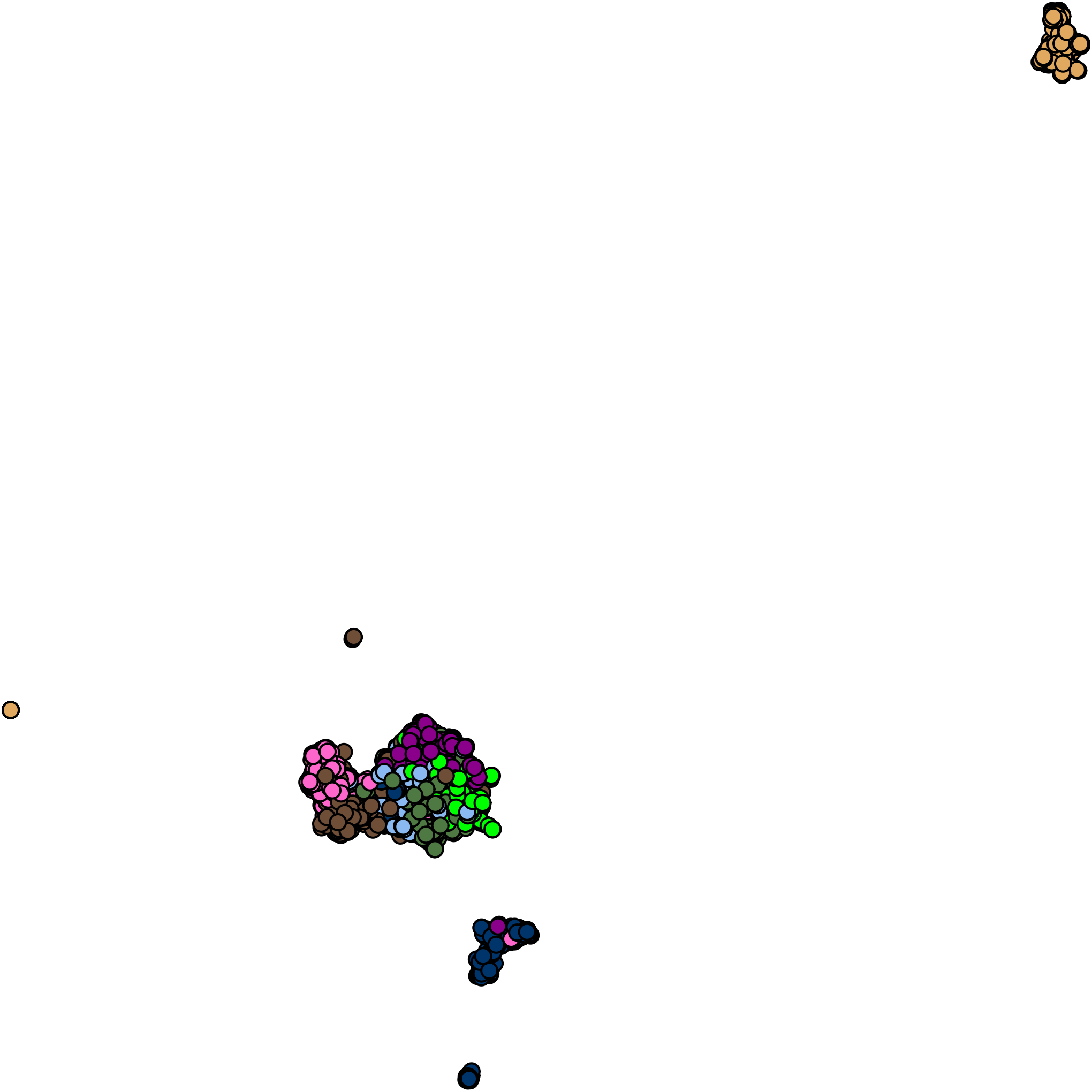}}}
 \subfigure[{\scriptsize UMAP on a subset of MNIST}]{\fbox{\includegraphics[width=.32\linewidth,height=4.5cm]{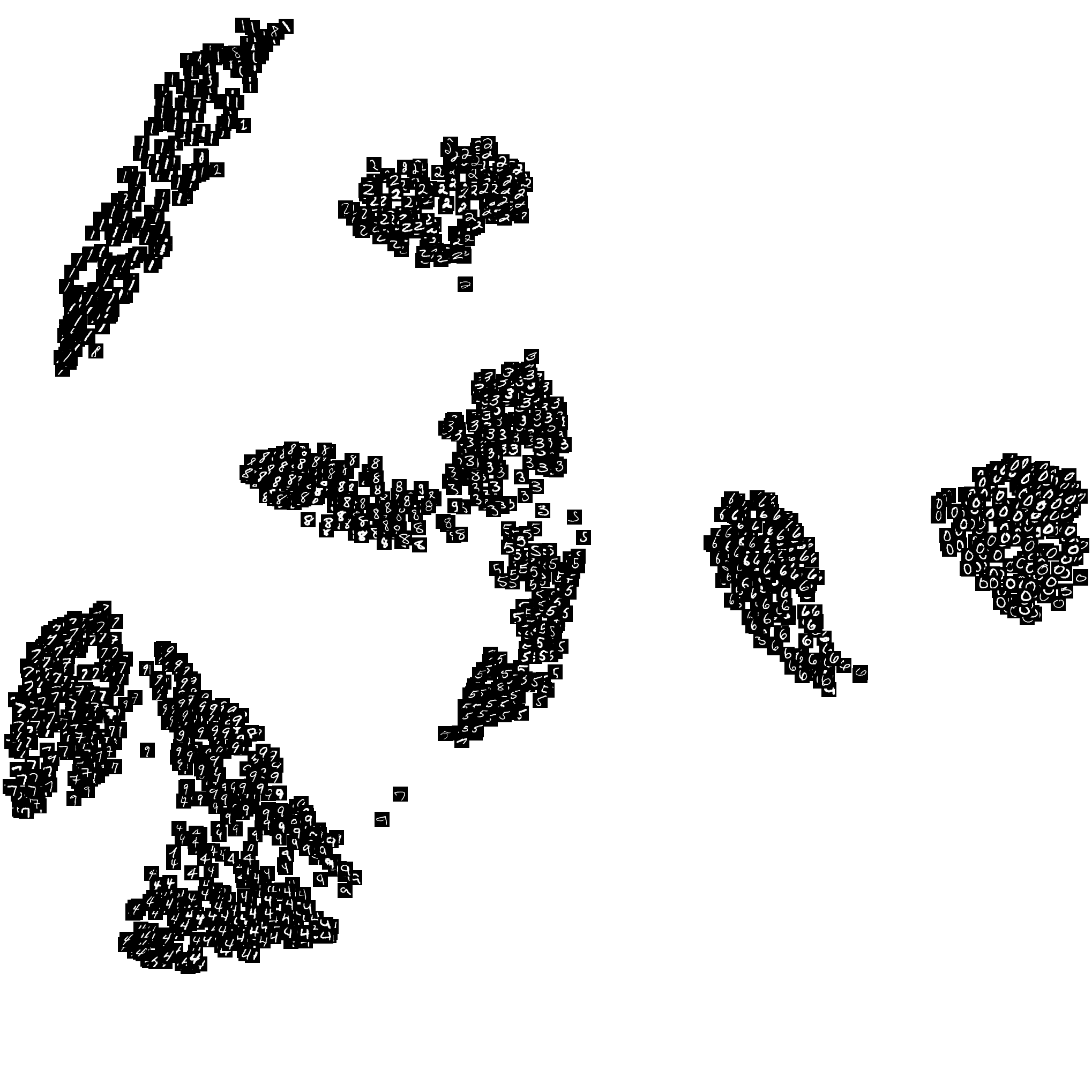}}}
 \subfigure[{\scriptsize UMAP on 1000 genomes project}]{\fbox{\includegraphics[width=.32\linewidth,height=4.5cm]{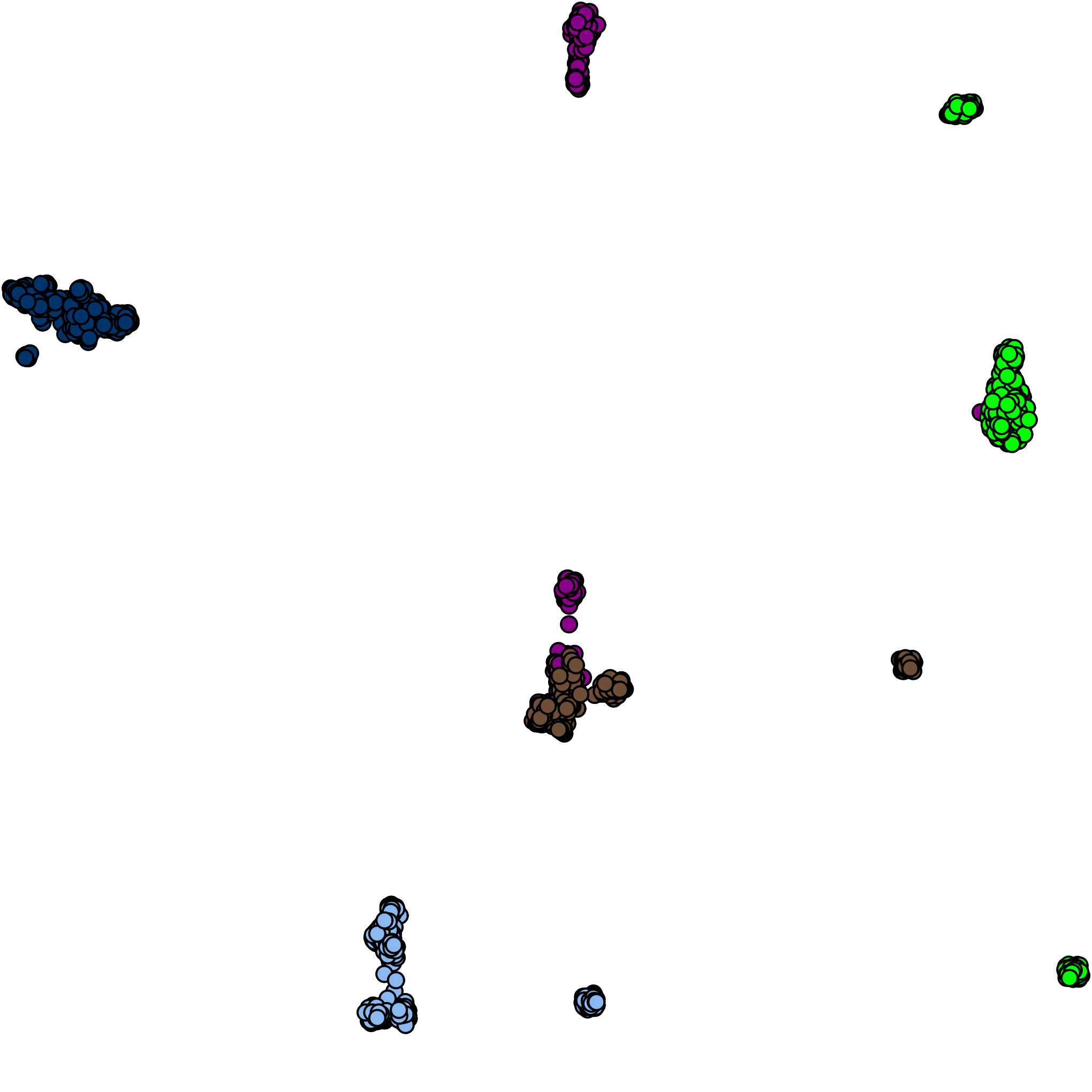}}}\\
 \subfigure[{\scriptsize LVSDE on MeeefTCD}]{\fbox{\includegraphics[width=.32\linewidth,height=4.5cm]{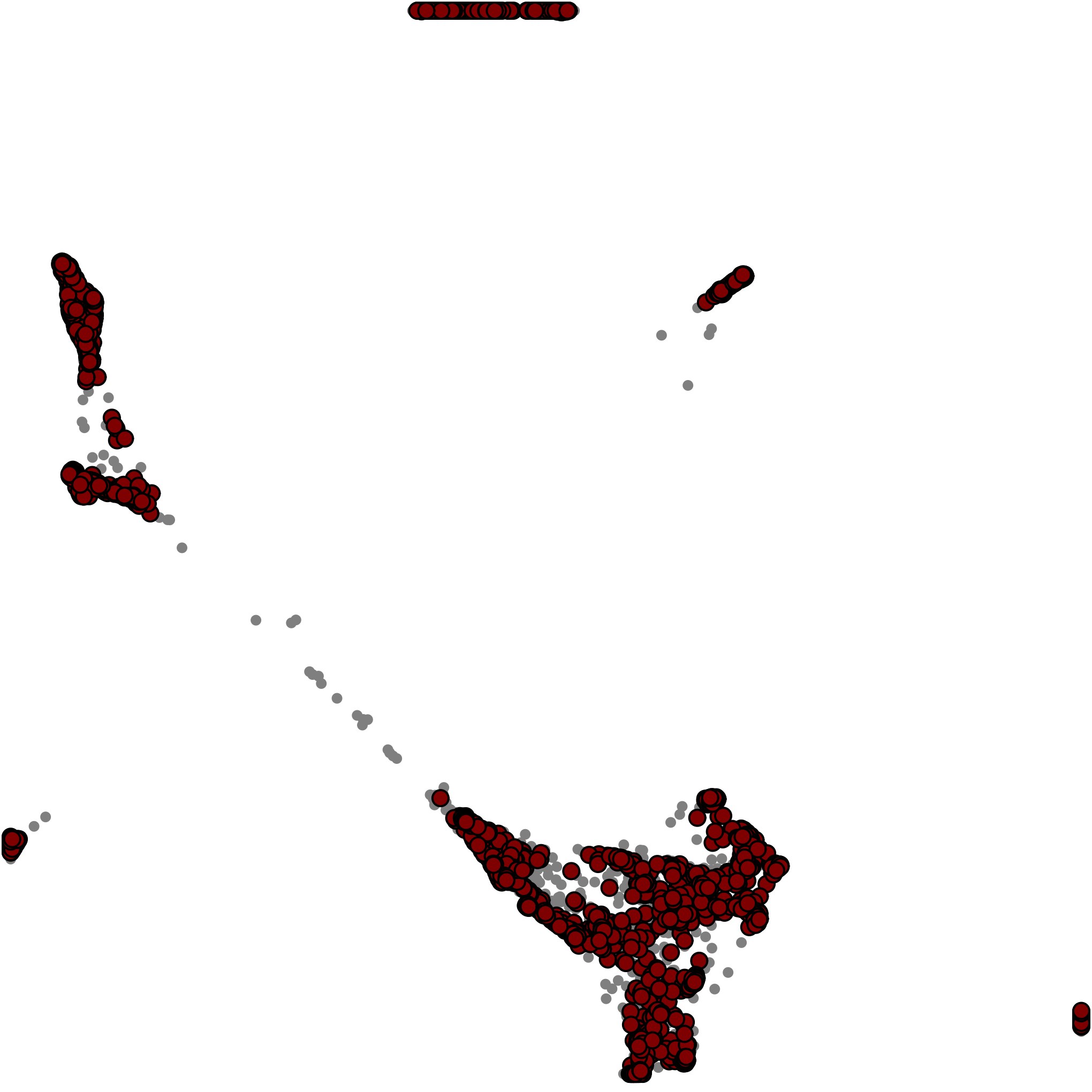}}}
 \subfigure[{\scriptsize LVSDE on a subset of MNIST}]{\fbox{\includegraphics[width=.32\linewidth,height=4.5cm]{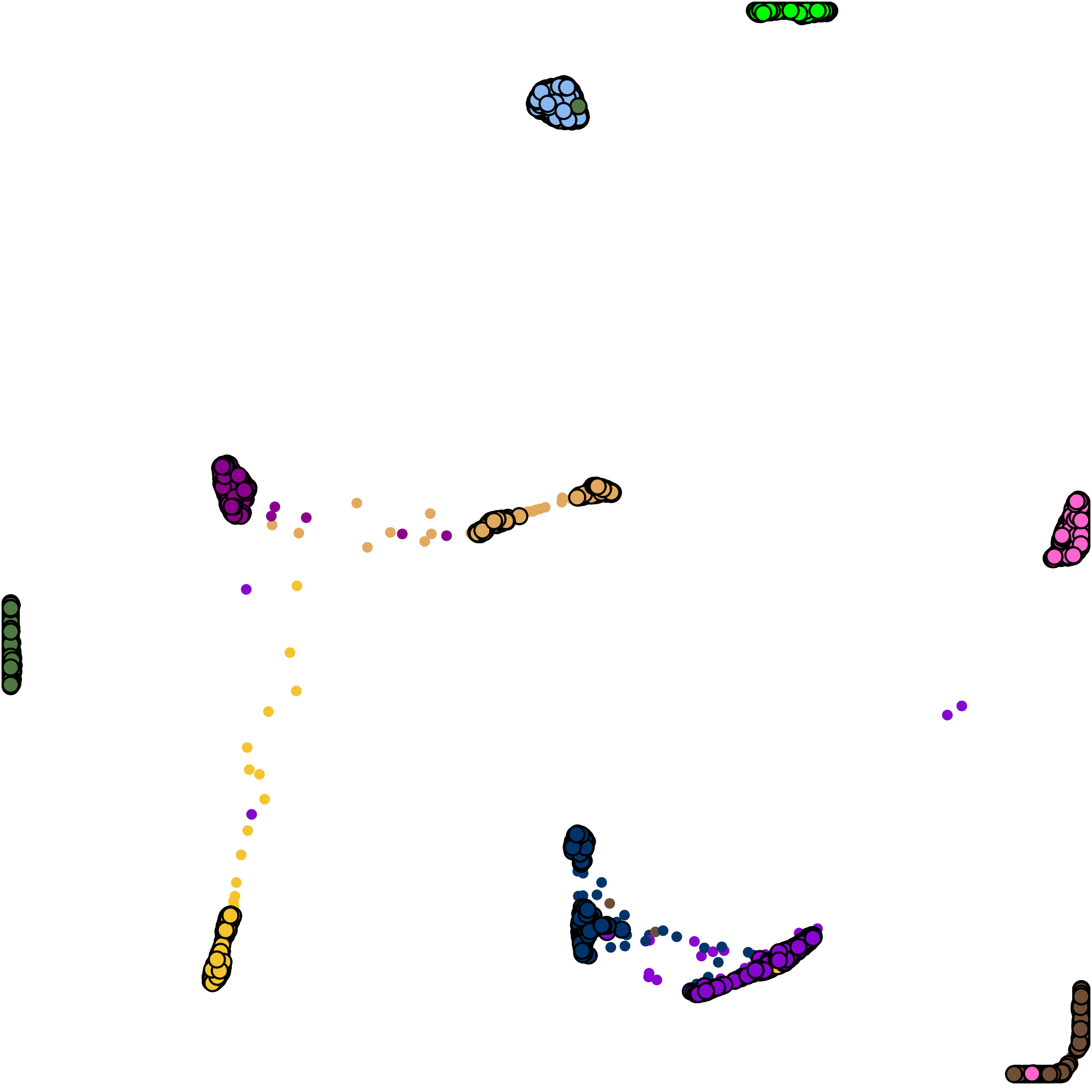}}}
 \subfigure[{\scriptsize LVSDE 2 on 1000 genomes project}]{\fbox{\includegraphics[width=.32\linewidth,height=4.5cm]{overall/images/polished/042.png}}}\\
 \subfigure[{\scriptsize t-SNE on MeefTCD}]{\fbox{\includegraphics[width=.32\linewidth,height=4.5cm]{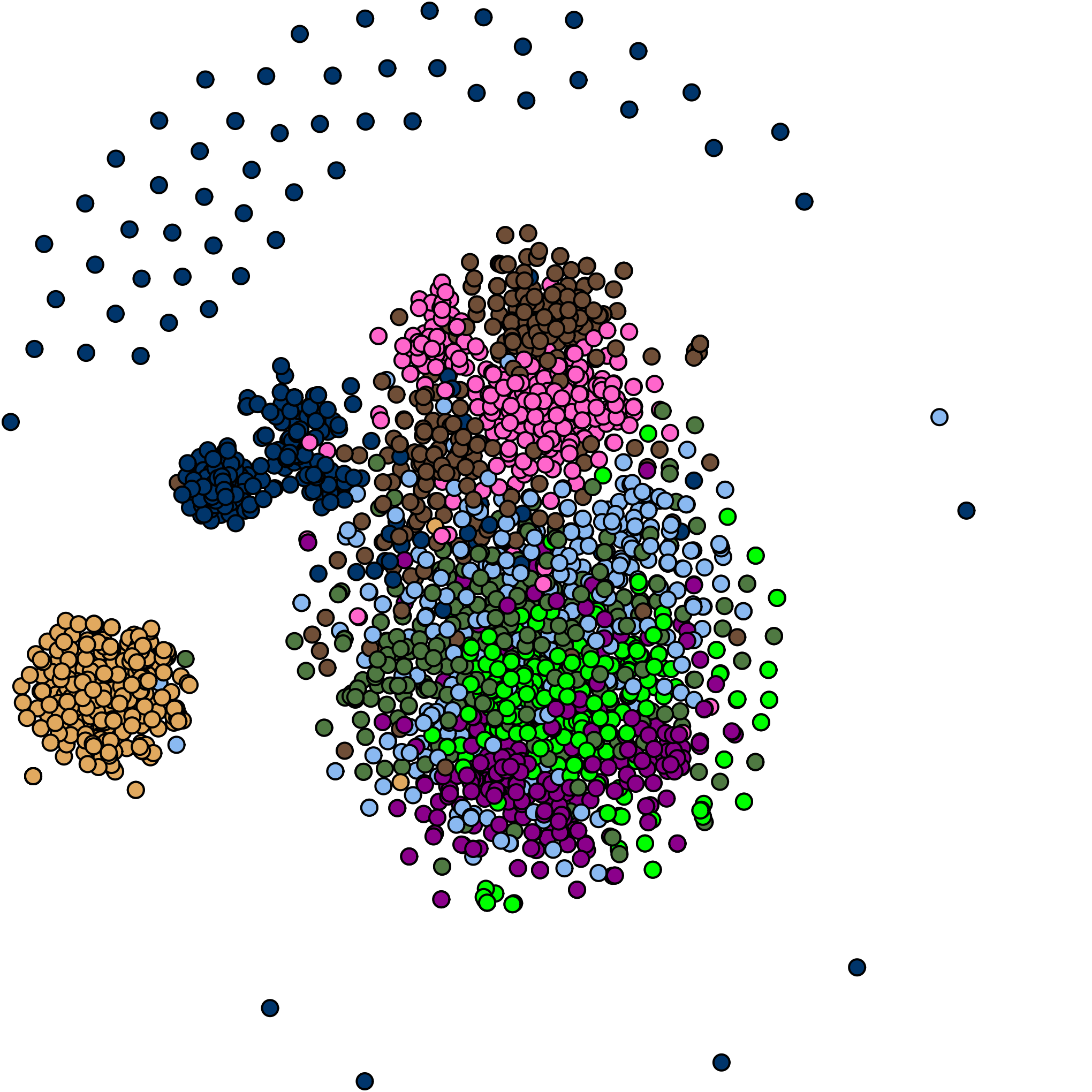}}}
 \subfigure[{\scriptsize t-SNE on a subset of MNIST}]{\fbox{\includegraphics[width=.32\linewidth,height=4.5cm]{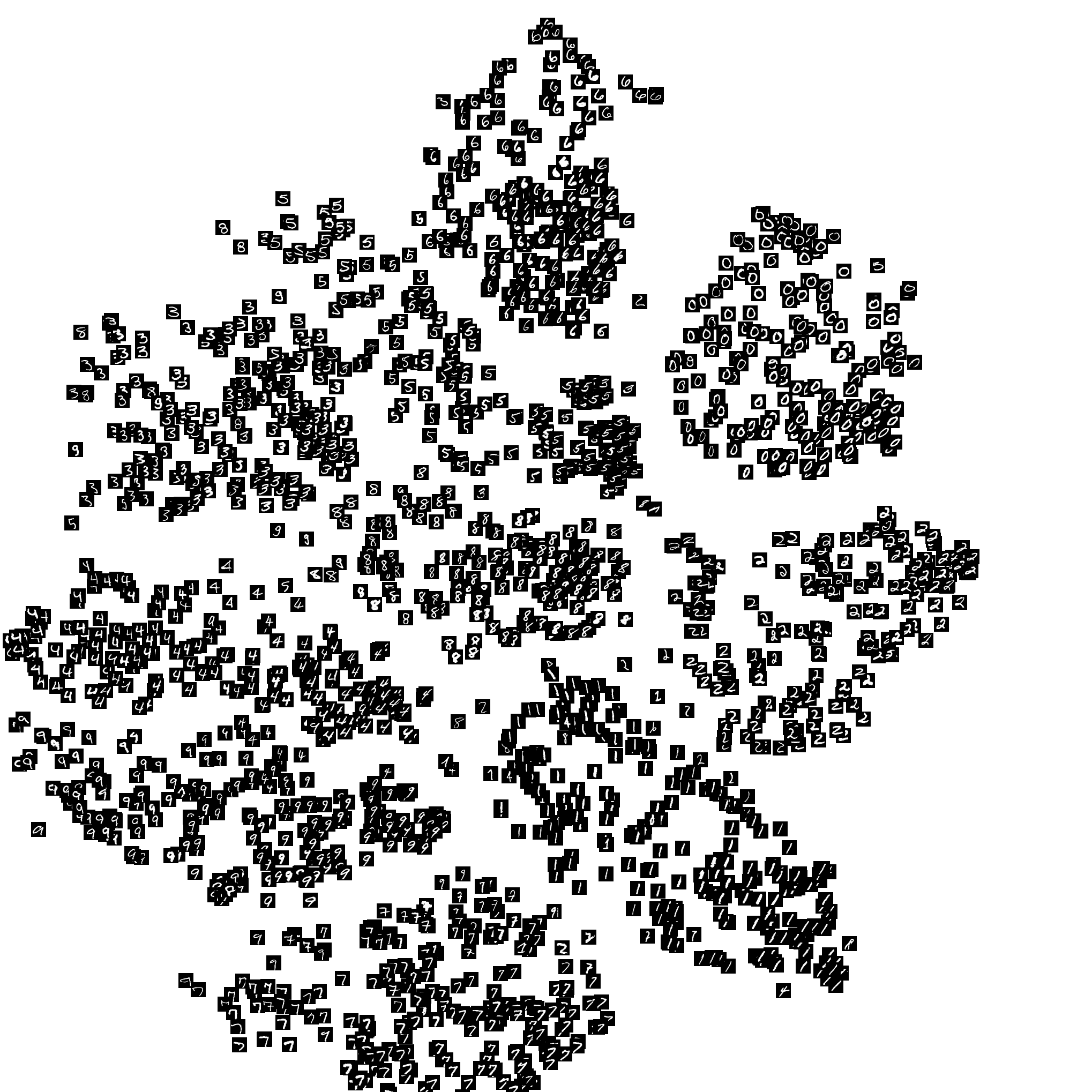}}}
 \subfigure[{\scriptsize t-SNE on 1000 genomes project}]{\fbox{\includegraphics[width=.32\linewidth,height=4.5cm]{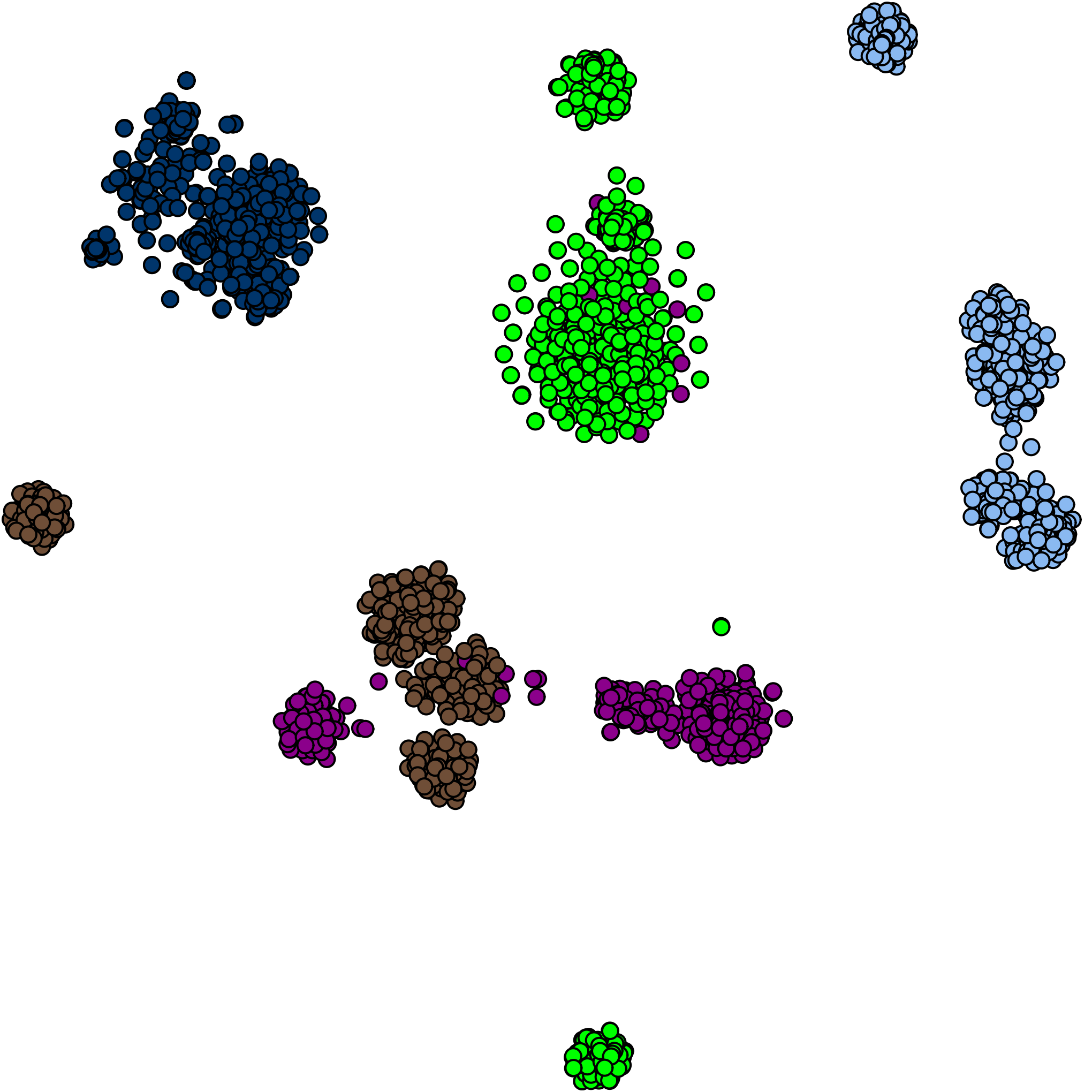}}}
 \caption{Embeddings of 3 data sets using different dimensionality reduction methods. In this figure, LVSDE 1 is abbreviation for LVSDE configuration 1, and LVSDE 2 is abbreviation for LVSDE configuration 2.}
 \label{fig:images}
\end{figure*}

Starting with qualitative results, the most notable result in terms of meaning is the embedding of the 1000 genomes project data set~\cite{a51_1000_genomes} distances by LVSDE displayed in Fig.~\ref{fig:1000_genomes_distances_embedding_LVSDE}. The 1000 genomes project being one of the first genomes data sets for studying human genome variety contains the DNA samples from different locations of the world labelled with geographical location. One important result that Fig.~\ref{fig:1000_genomes_distances_embedding_LVSDE} shows is that not only LVSDE has not produced random locations for duplications but also duplication of some of points of America from some area close to Europe to an area close to duplicates of some of points of Africa in an area of visual space between Europe and Africa, matches immigration history of humans from Europe and Africa to America. In this manner LVSDE has achieved multiple successes. One is identifying points to duplicate and second, duplicating them in a meaningful way to a correct area of visual space showing its capacity to guide the duplication. While Figs.~\ref{fig:images}(l) and ~\ref{fig:images}(f) show embeddings of 1000 genomes projected distances using t-SNE and UMAP, the fact that they do not support duplication, puts them in a weaker position in terms of meaning. In Fig.~\ref{fig:two_configs} two configurations of LVSDE are used for embedding the 1000 genomes project data set~\cite{a51_1000_genomes} distances. While configuration 1 performs better semantically, configuration 2, like UMAP and t-SNE in Figs.~\ref{fig:images}(l) and ~\ref{fig:images}(f) artificially performs better in groups separation, however knowing the phylogenetic connection of human genomes that separation has less semantic harmony with the reality of significant connection between human genomes. Configuration 1 although being better does not show the ideal representation of connection human genomes either and for future work it is an interesting research area to improve that in order to have an embedding method that shows the connection of human genomes better and more inclusive. The shortcoming in representation of that connection is an important flaw in the current state dimensionality reduction methods improved by LVSDE but not to the full extent. 

\begin{figure*}[!ht]
\centering
 \subfigure[LVSDE on a subset of MNIST]{\fbox{\includegraphics[width=.42\linewidth]{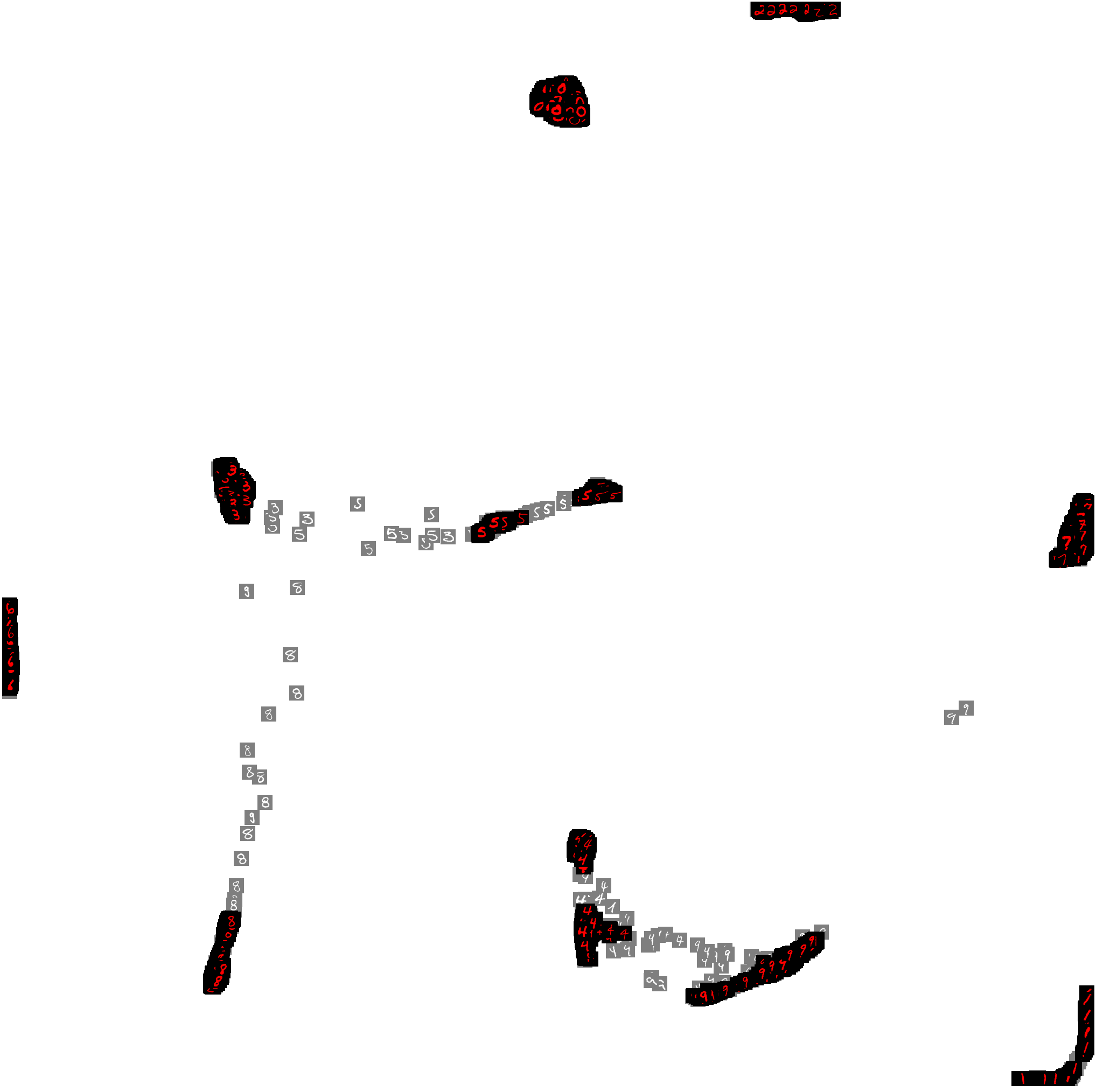}}\quad\quad}\quad
 \subfigure[A cropped section of LVSDE on a subset of MNIST]{\fbox{\includegraphics[width=.45\linewidth]{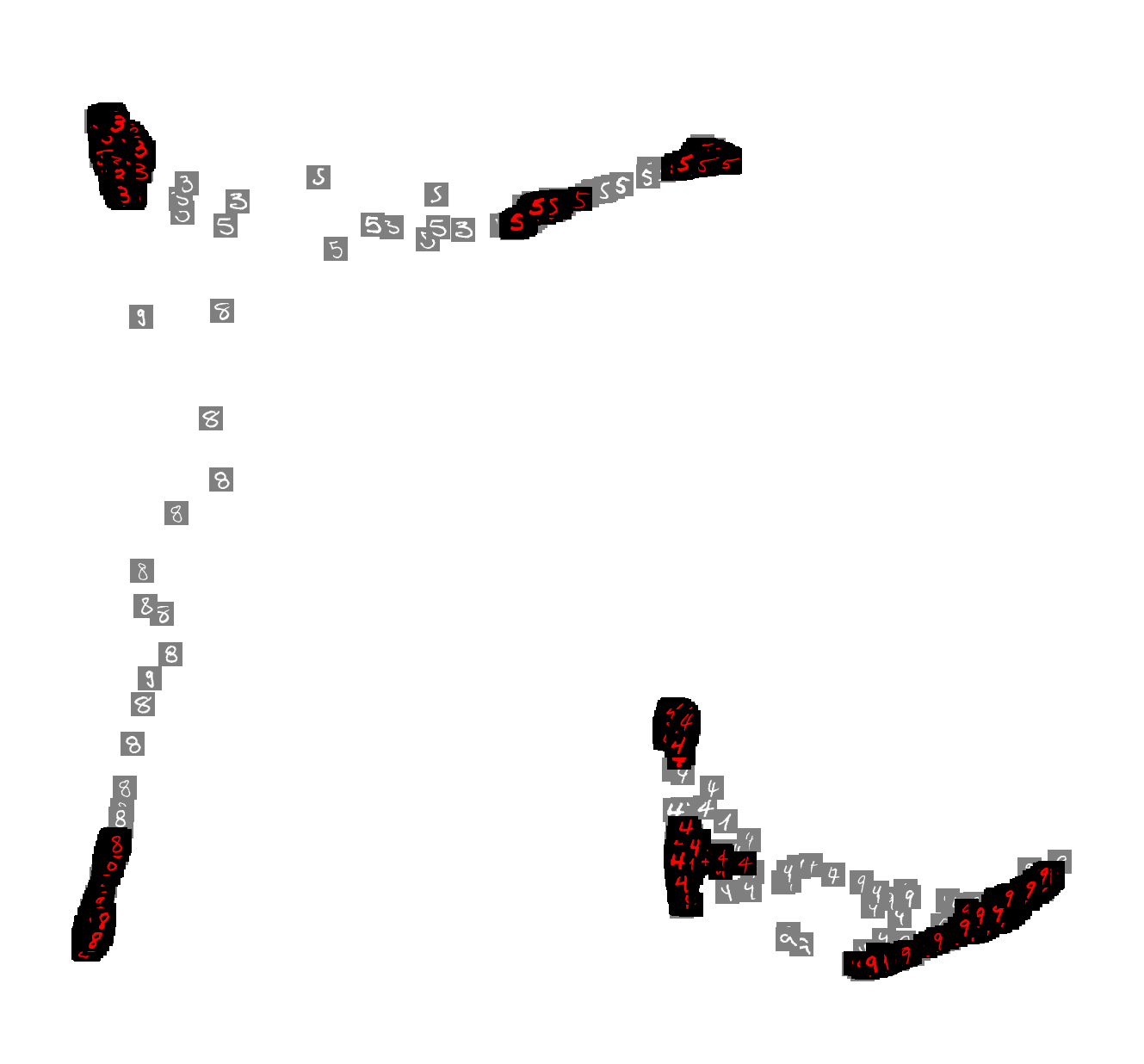}}\quad\quad}\\
 \subfigure[A cropped section of LVSDE on a subset of MNIST]{\fbox{\includegraphics[width=.50\linewidth]{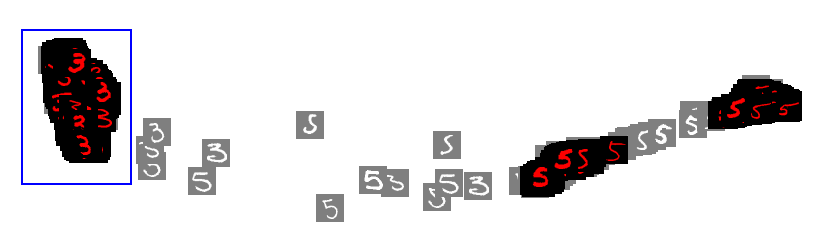}}\quad\quad}\quad
 \subfigure[Overlap reduced for blue rectangle in part c]{\fbox{\includegraphics[width=.16\linewidth]{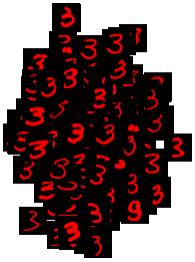}}\quad\quad\quad}\\
 \subfigure[A cropped section of LVSDE on a subset of MNIST]{\fbox{\includegraphics[width=.50\linewidth]{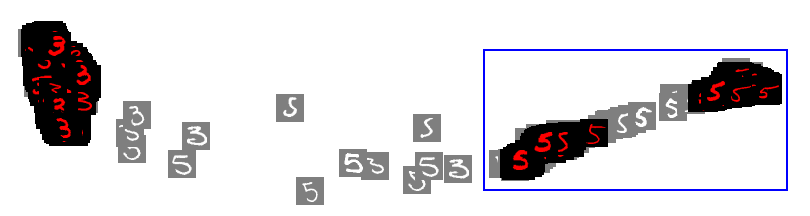}}\quad\quad}\quad
 \subfigure[Overlap reduced for blue rectangle in part e]{\fbox{\includegraphics[width=.2\linewidth]{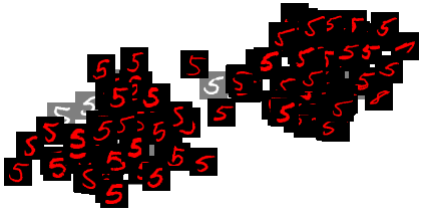}}\quad\quad}\quad
\caption{(a) An LVSDE embedding of a subset of MNIST data set using the information from the whole data set training data.
(b) A cropped section of the LVSDE embedding of a subset of MNIST data set. Separation of digits $3$,$5$ and $8$ on the red layer are better than popular techniques. Separation of digits $4$ and $9$ on the red layer are better than popular techniques. By using layers, the identification of groups is much easier when only considering red layer while gray layer provides some background information.
(c,e) A cropped section of the LVSDE embedding of a subset of MNIST data set.
(d,f) A variant of overlap reduction algorithm of \cite{a62} by Nachmanson et al. is used on the points in blue rectangle in part c or e of this figure.
}
\label{fig:digits_1}
\end{figure*}

\begin{figure*}[!ht]
\centering
 \subfigure[LVSDE on a subset of MNIST]{\fbox{\includegraphics[width=.4\linewidth]{overall/images/polished/041.png}}\quad\quad}\quad
 \subfigure[A cropped section of LVSDE on a subset of MNIST]{\fbox{\includegraphics[width=.47\linewidth]{overall/images/polished/008.png}}\quad\quad}\quad\\
 \subfigure[A cropped section of LVSDE on a subset of MNIST]{\fbox{\includegraphics[width=.40\linewidth]{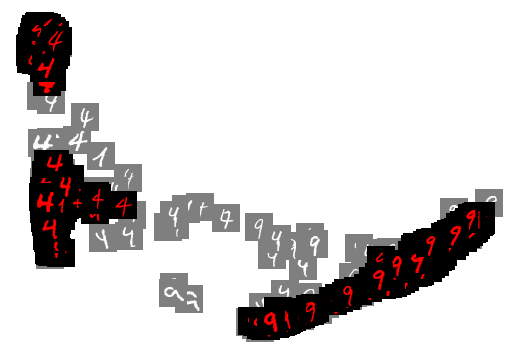}}\quad\quad}\quad
 \subfigure[Duplicates flagged for user selection]{\fbox{\includegraphics[width=.40\linewidth]{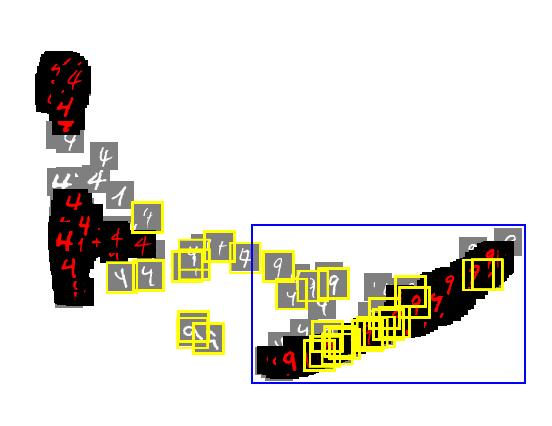}}\quad\quad}\quad
\caption{(a) An LVSDE embedding of a subset of MNIST data set using the information from the whole data set training data.
(b) A cropped section of the LVSDE embedding of a subset of MNIST data set. Separation of digits $3$,$5$ and $8$ on the red layer are better than popular techniques. Separation of digits $4$ and $9$ on the red layer are better than popular techniques. By using layers, the identification of groups is much easier when only considering red layer while gray layer provides some background information.
(c) A cropped section of the LVSDE embedding of a subset of MNIST data set.
(d) Yellow rectangles specify points that have a corresponding duplicate point in the blue rectangle. The blue rectangle is specified by the user.
}
\label{fig:digits_2}
\end{figure*}

Moving on from 1000 genomes project data set, the next important result is LVSDE on a subset of MNIST data set. For MNIST data set, it is known that popular dimensionality reduction methods have some difficulty in separating digits $3$,$5$ and $8$. The handwritten digits do not necessarily correspond to digital digits in a $100$\% confident way. A handwritten digit, especially if not written well may be for example $80$\% $3$, $60$\% $5$ and $30$\% $8$, the so called multivariate fuzzy semantics that was talked about in Section~\ref{section:motivation_and_philosophy}. Digits $3$, $5$ and $8$ are not the only digits that cause trouble for dimensionality reduction techniques on MNIST data set. Digits $4$ and $9$ also confuse dimensionality reduction techniques limiting the ability of dimensionality reduction techniques to separate them. Again the case of multivariate fuzzy semantics exists for some of the handwritten digits of $4$ and $9$. Looking at Figs.~\ref{fig:digits_1} and ~\ref{fig:digits_2}, separation of digits $3$,$5$ and $8$ on the red layer are better than popular techniques and also separation of digits $4$ and $9$ on the red layer are better than popular techniques. By using layers, the identification of groups in the LVSDE embedding of the subset of MNIST data set is much easier when only considering red layer while gray layer provides some background information. For comparison, Figs.~\ref{fig:images}(k) and \ref{fig:images}(e) show t-SNE and UMAP embeddings of the subset of the MNIST data set but the reader may already have a good sense of MNIST embeddings by t-SNE and UMAP from prior publications.

\begin{figure*}[htb]
 \centering
 \subfigure[LVSDE on IRIS]{\fbox{\includegraphics[width=.42\linewidth]{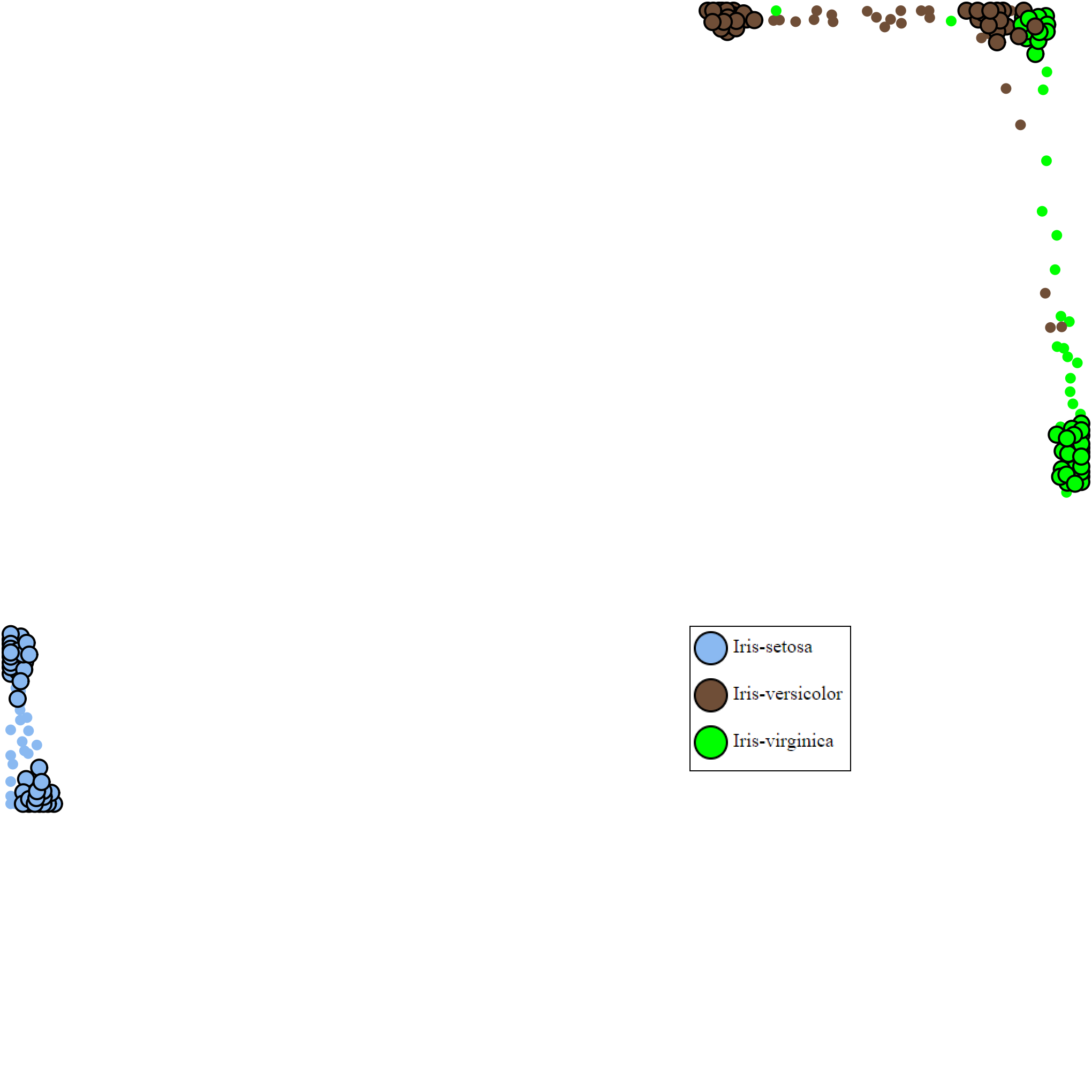}}}\quad
 \subfigure[LVSDE on IRIS]{\fbox{\includegraphics[width=.42\linewidth]{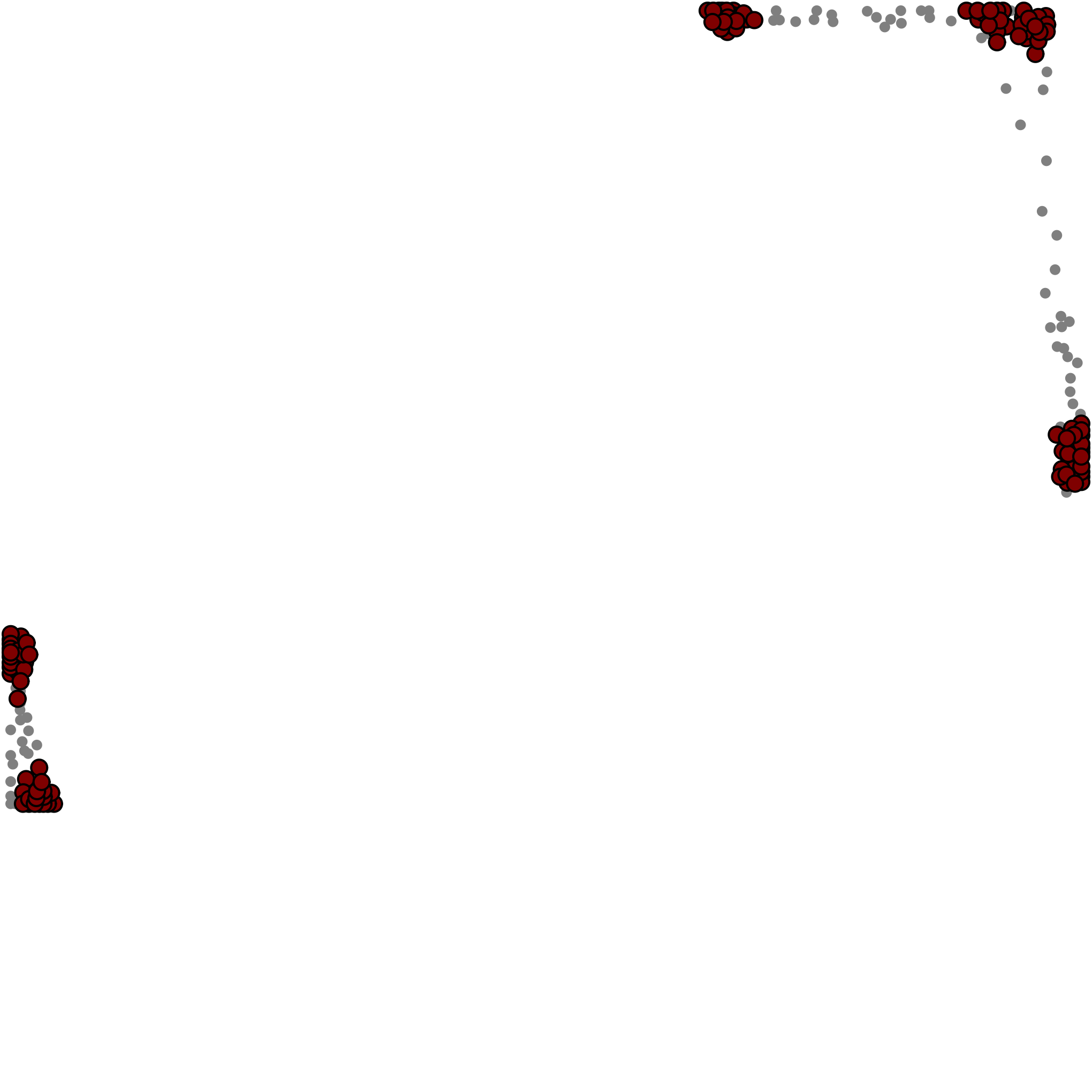}}}\quad\\
 \subfigure[t-SNE on IRIS]{\fbox{\includegraphics[width=.42\linewidth]{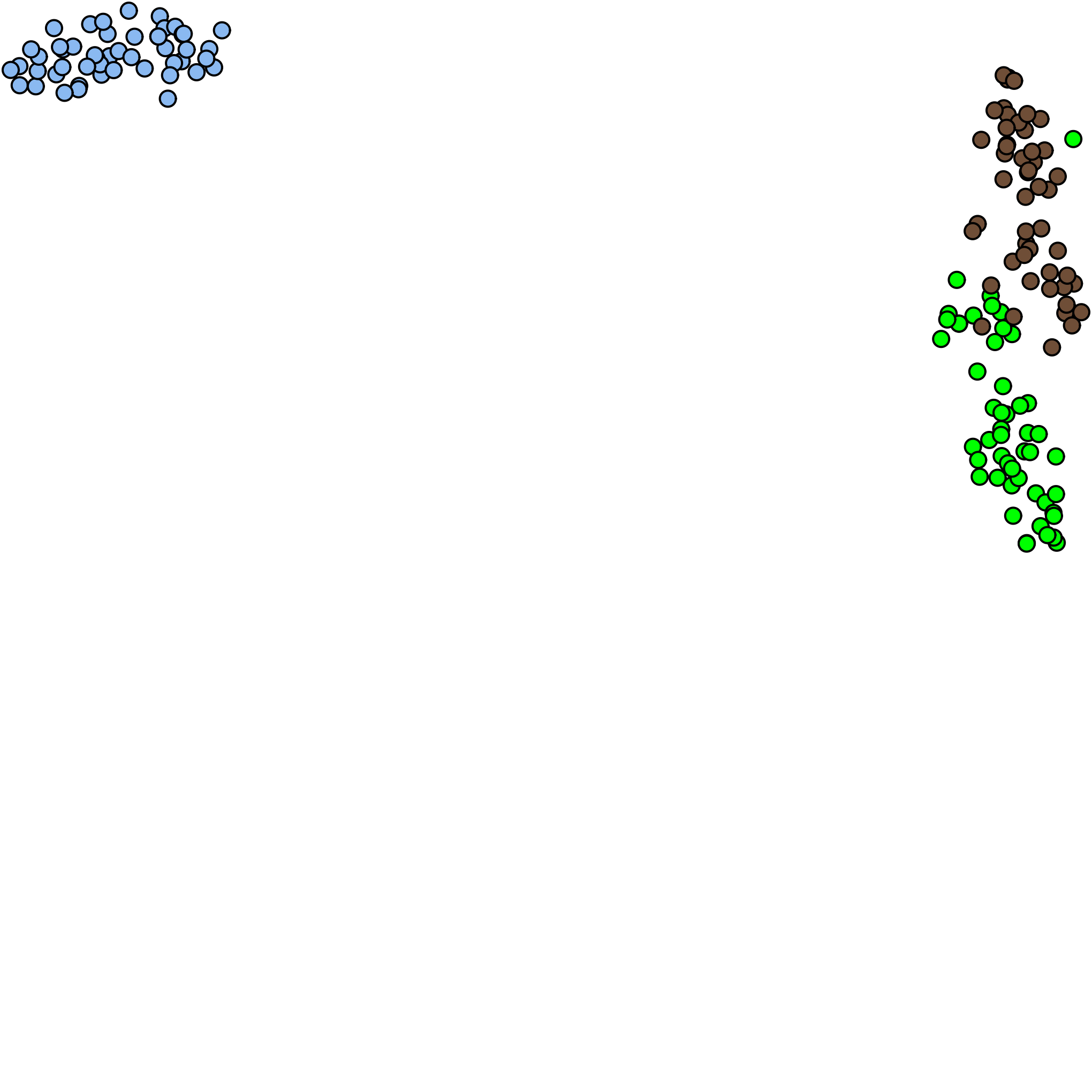}}}\quad
 \subfigure[UMAP on IRIS]{\fbox{\includegraphics[width=.42\linewidth]{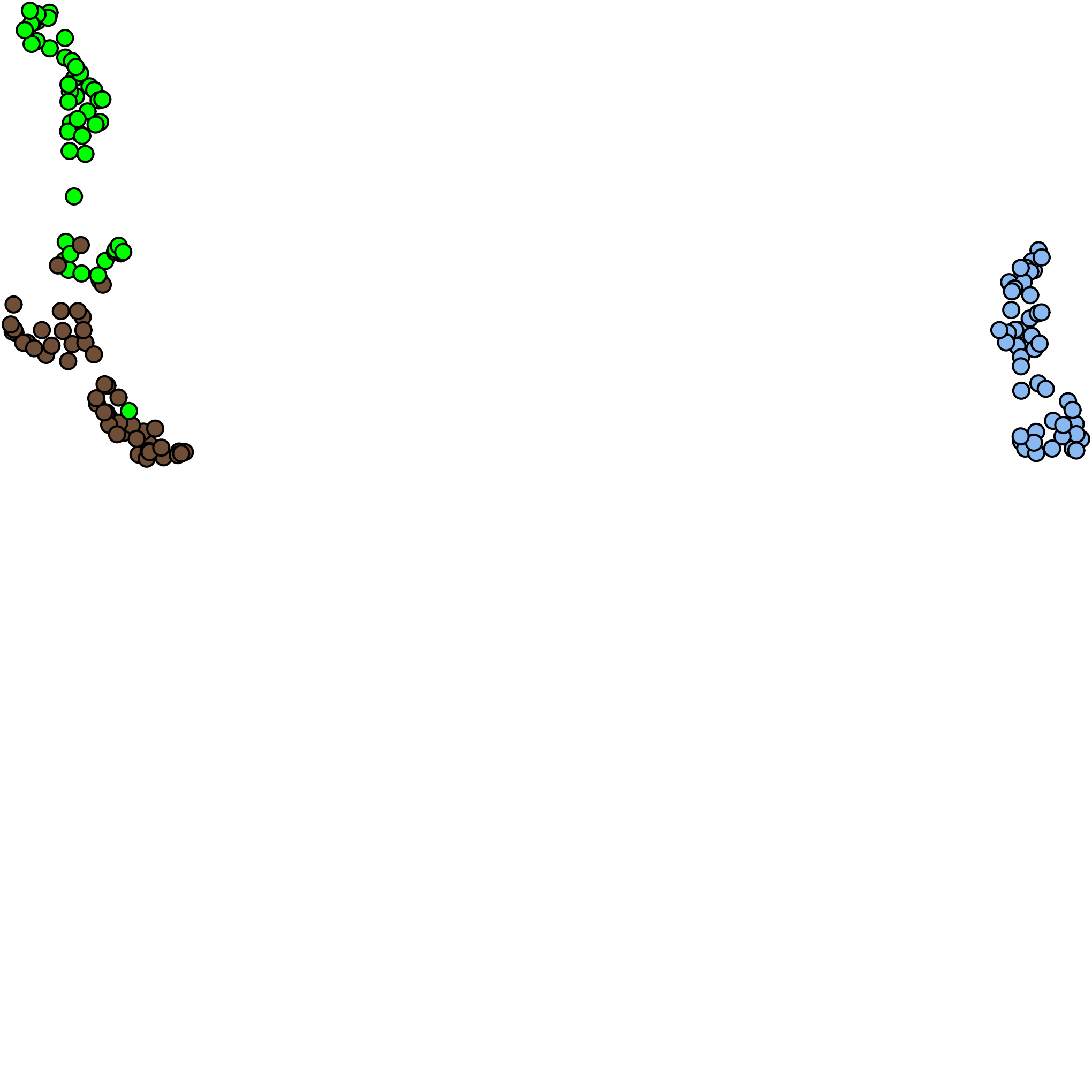}}}\quad
 \caption{Embeddings of the IRIS data set using different dimensionality reduction methods. For LVSDE points with a black circle around them are in the red layer and the points in the gray layer are drawn smaller. For LVSDE in part b, points in the red layer are coloured red and the points in the gray layer are coloured gray. LVSDE has performed better for subgroup detection.}
 \label{fig:IRIS}
\end{figure*}

Next in line is the IRIS data set where detecting subgroups is the center of attention. Identifying a subgroup means identifying a significantly large subset of a group in a confident manner that only contains data instances from that group but not all of them. Being a very simple data set, IRIS has only three classes for groups. By looking at Fig.~\ref{fig:IRIS}(b) which does not have class colours, one would normally immediately identify at least 4 subgroups, out of which only one of them is wrong meaning at most $25$\% chance of failure. In contrast, if class colours are removed from Fig.~\ref{fig:IRIS}(c) for t-SNE on IRIS, one would normally identify only two subgroups, out of which one of them is wrong meaning $50$\% chance of failure. While UMAP on IRIS shown in Fig.~\ref{fig:IRIS}(d), is not as bad as t-SNE, it is not as clear as LVSDE, and interpretation may vary more among different observers comparing to LVSDE.

\begin{figure*}[htb]
 \centering
 \subfigure[LVSDE on MeefTCD]{\fbox{\includegraphics[width=.42\linewidth]{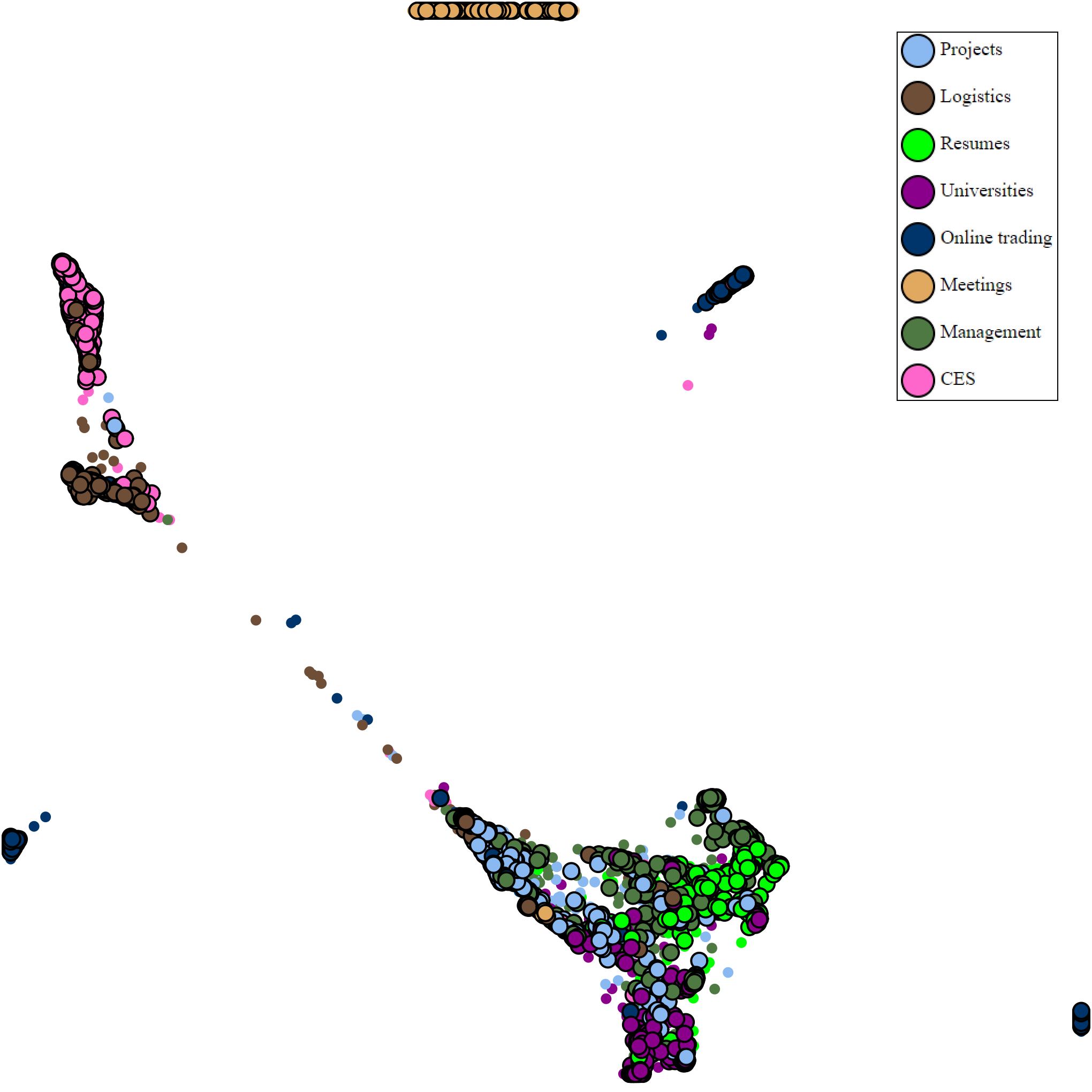}}}\quad
 \subfigure[LVSDE on MeefTCD (two classes combined)]{\fbox{\includegraphics[width=.42\linewidth]{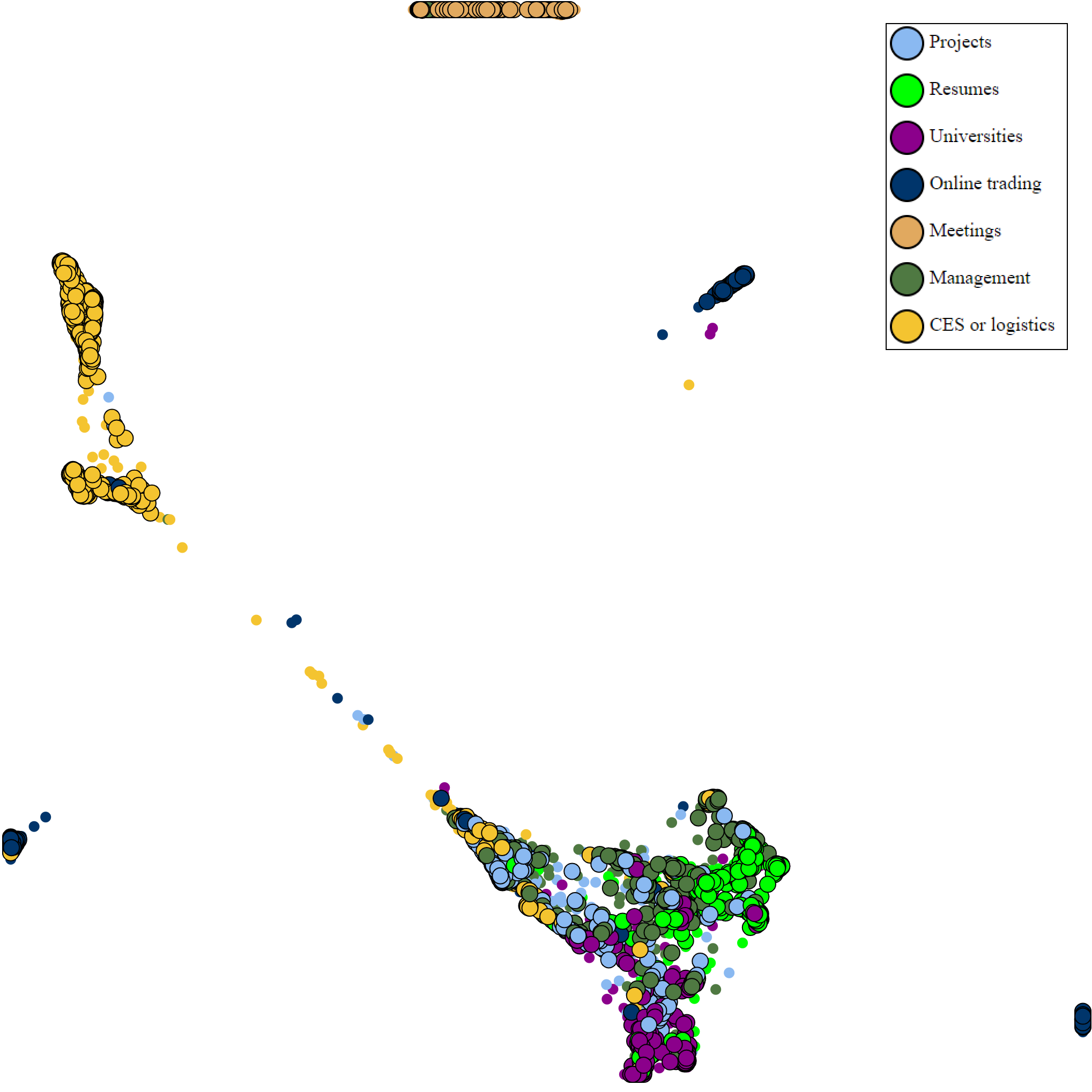}}}\quad\\
 \subfigure[t-SNE on MeefTCD]{\fbox{\includegraphics[width=.42\linewidth]{overall/images/polished/021.png}}}\quad
 \subfigure[UMAP on MeefTCD]{\fbox{\includegraphics[width=.42\linewidth]{overall/images/polished/015.png}}}\quad
 \caption{Embeddings of the MeefTCD data set using different dimensionality reduction methods. For LVSDE points with a black circle around them are in the red layer and the points in the gray layer are drawn smaller. For LVSDE in part b, two of the classes are combined. The separation of the combination of two classes CES and logistics is much better in the LVSDE embedding.}
 \label{fig:MeeefTCD_1}
\end{figure*}

\begin{figure*}[htb]
 \centering
 \subfigure[LVSDE on MeefTCD]{\fbox{\includegraphics[width=.42\linewidth]{overall/images/polished/013.png}}}\quad
 \subfigure[LVSDE on MeefTCD (two classes combined)]{\fbox{\includegraphics[width=.42\linewidth]{overall/images/polished/033.png}}}\quad\\
 \subfigure[LVSDE on MeefTCD (two classes combined)]{\fbox{\includegraphics[width=.22\linewidth]{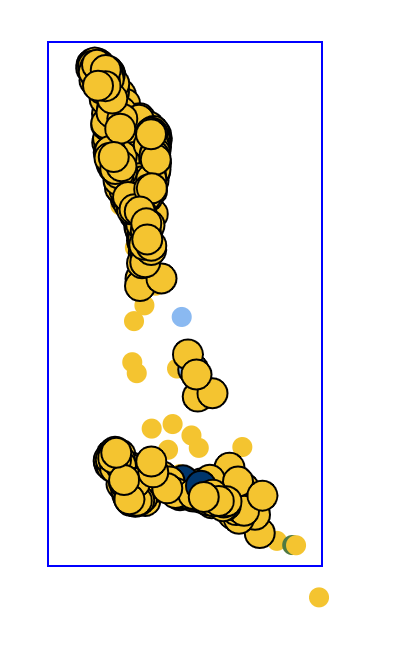}}}\quad
 \subfigure[Overlap reduced for blue rectangle in part c]{\fbox{\includegraphics[width=.22\linewidth]{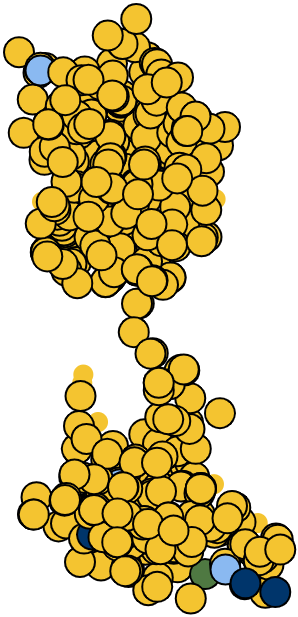}}}\quad
 \caption{Embedding of the MeefTCD data set using LVSDE method. Points with a black circle around them are in the red layer and the points in the gray layer are drawn smaller. For LVSDE in part b, c and d, two of the classes are combined. The separation of the combination of two classes CES and logistics is much better in the LVSDE embedding. (d) A variant of overlap reduction algorithm of \cite{a62} by Nachmanson et al. is used on the points in blue rectangle in part c of this figure}
 \label{fig:MeeefTCD_2}
\end{figure*}

For the MeeefTCD data set, identification of combinational groups is the center of attention. Looking at Figs.~\ref{fig:MeeefTCD_1} and~\ref{fig:MeeefTCD_2} the separation of the combination of two classes CES and logistics is much better in the LVSDE embedding comparing to t-SNE and UMAP.

\FloatBarrier
\cleardoublepage

\FloatBarrier

\onecolumn

\begin{multicols}{2}
While embeddings can be visually evaluated qualitatively, it is good to also have some quantitative measurement of the embeddings. The objectives of qualitative analysis and quantitative are not necessarily the same as they evaluate different aspects of the embeddings and quantitative analysis alone at least in the way that is done in this paper should not be considered as the ultimate indicator of better embeddings and rather it provides a single view of one aspect of embeddings. Since Multi-layered Multi-point Dimensionality Reduction and therefore Strict Red Gray Embeddings are not completely compatible with existing quantitative measurements some discussion is needed on how to adapt current quantitative measurements for them. In particular a measure called $\Lambda$ measure in this paper is used in this paper which is nothing new or different from KNN classification accuracy except adaptation to Multi-layered Multi-point Dimensionality Reduction and Strict Red Gray Embeddings. For a layer $\theta$ (or set of layers $L$) of a Multi-layered Multi-point Dimensionality Reduction embedding, the $\Lambda$ measure is basically the percentage (or ratio) of data instances that have at least one projection in $\theta$ (or $L$) and the class label for that data instance matches the label with maximum occurrence among $k$ nearest neighbours of any of the projections of that data instance in $\theta$ (or $L$) where neighbours are limited to a specific set of layer(s) in visual space called classification layer(s) $\widehat{L}$ and the result of evaluation is denoted by $\Lambda_{(\theta,\widehat{L})}$ or $\Lambda_{(L,\widehat{L})}$ respectively ($k$ is called evaluation neighbourhood size, $\theta$ or $L$ are called evaluation layer(s). The choice of $k$, $\theta$ or $L$ and $\widehat{L}$ are parameters to evaluation. In this paper $k=15$ is used the evaluation neighbourhood size for all experiments of the paper).

Tables~\ref{table:table_lambda} and~\ref{table:table_lambda_2} show the quantitative results and comparison on the four data sets using different dimensionality reduction methods.

\end{multicols}

\begin{table*}[h]
\scriptsize
\begin{tabular}{| l | p{2.5cm} | c | c | c | c | c |}
\hlineB{4}
Data set & Embedding method & Evaluation layer(s) & Classification layer(s) & Evaluation \\
\hlineB{4}
1000 genomes & LVSDE 1 & red and gray & red and gray & 98.562\% \\
1000 genomes & LVSDE 1 & red and gray & red & 98.083\% \\
1000 genomes & LVSDE 1 & red & red & 98.551\% \\
1000 genomes & LVSDE 1 & gray & gray & 97.802\% \\
1000 genomes & LVSDE 1 & gray & red & 95.330\% \\
1000 genomes & LVSDE 1 & gray & red and gray & 98.901\% \\
\hline
1000 genomes & LVSDE 2 & red and gray & red and gray & 99.681\% \\
1000 genomes & LVSDE 2 & red and gray & red & 99.641\% \\
1000 genomes & LVSDE 2 & red & red & 99.708\% \\
1000 genomes & LVSDE 2 & gray & gray & 99.333\% \\
1000 genomes & LVSDE 2 & gray & red & 99.333\% \\
1000 genomes & LVSDE 2 & gray & red and gray & 99.556\% \\
\hline
1000 genomes & UMAP & one layer (all) & one layer (all) & 99.321\% \\
\hline
1000 genomes & t-SNE (Barnes Hut variant) & one layer (all) & one layer (all)  & 99.241\% \\
\hline
\hlineB{4}
MNIST & LVSDE & red and gray & red and gray & 96.350\% \\
MNIST & LVSDE & red and gray & red & 95.950\% \\
MNIST & LVSDE & red & red & 96.587\% \\
MNIST & LVSDE & gray & gray & 91.558\% \\
MNIST & LVSDE & gray & red & 88.312\% \\
MNIST & LVSDE & gray & red and gray & 93.506\% \\
\hline
MNIST & UMAP & one layer (all) & one layer (all) & 96.250\% \\
\hline
MNIST & t-SNE (Barnes Hut variant) & one layer (all) & one layer (all)  & 95.450\% \\
\hlineB{4}
\end{tabular}\vspace{0.1cm}
\caption{Experiments' $\Lambda_{(L,\widehat{L})}$ measures on the embeddings of two of the data sets (1000 genomes project data set~\cite{a51_1000_genomes} distances and a subset of MNIST) using different $L$ (evaluation layer(s)) and $\widehat{L}$ (classification layer(s)). In this table, LVSDE 1 is abbreviation for LVSDE configuration 1, and LVSDE 2 is abbreviation for LVSDE configuration 2. For t-SNE and UMAP,  all projected points are assumed to be in one layer which is used as evaluation layer and classification layer. All the numbers in this table are rounded to 3 decimal places. Evaluation neighbourhood size $k=15$ is used.}
\label{table:table_lambda}
\end{table*}

\begin{table*}[h!]
\scriptsize
\begin{tabular}{| l | p{2.5cm} | c | c | c | c | c |}
\hlineB{4}
Data set & Embedding method & Evaluation layer(s) & Classification layer(s) & Evaluation \\
\hlineB{4}
MeeefTCD & LVSDE & red and gray & red and gray & 76.083\% \\
MeeefTCD & LVSDE & red and gray & red & 75.667\% \\
MeeefTCD & LVSDE & red & red & 76.513\% \\
MeeefTCD & LVSDE & gray & gray & 69.778\% \\
MeeefTCD & LVSDE & gray & red & 72.000\% \\
MeeefTCD & LVSDE & gray & red and gray & 72.889\% \\
\hline
MeeefTCD & UMAP & one layer (all) & one layer (all) & 74.708\% \\
\hline
MeeefTCD & t-SNE (Barnes Hut variant) & one layer (all) & one layer (all)  & 75.583\% \\
\hline
\hlineB{4}
IRIS & LVSDE & red and gray & red and gray & 95.333\% \\
IRIS & LVSDE & red and gray & red & 95.333\% \\
IRIS & LVSDE & red & red & 97.368\% \\
IRIS & LVSDE & gray & gray & 86.111\% \\
IRIS & LVSDE & gray & red & 88.889\% \\
IRIS & LVSDE & gray & red and gray & 91.667\% \\
\hline
IRIS & UMAP & one layer (all) & one layer (all) & 97.333\% \\
\hline
IRIS & t-SNE (Barnes Hut variant) & one layer (all) & one layer (all)  & 97.333\% \\
\hlineB{4}
\end{tabular}\vspace{0.1cm}
\caption{Experiments' $\Lambda_{(L,\widehat{L})}$ measures on the embeddings of two of the data sets (MeeefTCD and IRIS) using different $L$ (evaluation layer(s)) and $\widehat{L}$ (classification layer(s)). For t-SNE and UMAP,  all projected points are assumed to be in one layer which is used as evaluation layer and classification layer. All the numbers in this table are rounded to 3 decimal places. Evaluation neighbourhood size $k=15$ is used.}
\label{table:table_lambda_2}
\end{table*}

\cleardoublepage

\FloatBarrier

\section{Availability}

Please visit \url{https://web.cs.dal.ca/~barahimi/chocolate-lvsde/} for an implementation of the LVSDE (date accessed: May 29, 2022). Chocolate LVSDE is the name of a particular implementation of LVSDE in written in Go programming language (majority of the code volume) and Python programming language, sometimes interoperating through C programming language interface of Go as the intermediate language interface, in addition to some codes written in Javascript, CSS and HTML. Version 1.7 of Chocolate LVSDE is used for the experiments of this paper which is a parallel implementation of LVSDE where the number of parallel goroutines is one less than the number of CPU cores.

\section{Conclusion And Future Work}
\label{sec:conclusions}
A novel dimensionality reduction method named LVSDE was presented and the philosophy of its underlining platform Multi-layered Multi-point Dimensionality Reduction and Multi-point Dimensionality Reduction was discussed and elaborated on. The road ahead is open on many fronts including but not limited to improving speed, application to data sets in different domains and visual metaphors. If the challenge of visually connecting different projections of a data instance is adequately addressed, preservation of well defined finite point set typology structures can become a significant research direction in future. Finding better measures to choose the number of projected points in the red layer is another direction to improve. While LVSDE limits the number of projected points per data instance to 2, going beyond 2 is another direction of interest.

\section{Acknowledgments}
Dr. Fernando Paulovich used to be in the authors list of this preprint (as a collective of all versions) but he has requested that his name be removed from the authors list of this paper (future submissions and the arXiv preprint as the collective of all versions). The request was made at the same time that he was withdrawing from PhD supervision of Farshad Barahimi. Other than the endorsement of Farshad Barahimi, no other endorsement should be implied from this version of this preprint.

\printbibliography[title={References}]

\end{document}